%% file: main.tex
\documentclass[twoside,11pt]{article}

\usepackage{blindtext}

\usepackage{lastpage}
\usepackage{jmlr2e}

\input{math_commands.tex}

\usepackage[utf8]{inputenc} %
\usepackage[T1]{fontenc}    %
\usepackage{hyperref}
\usepackage{url}
\usepackage{booktabs}       %
\usepackage{amsfonts}       %
\usepackage{nicefrac}       %
\usepackage{microtype}      %
\usepackage{xcolor}         %

\usepackage{graphicx}
\usepackage{epsfig}
\usepackage{pifont}%
\usepackage{enumitem}
\usepackage{verbatim}
\usepackage{bbm}
\usepackage{makecell,multirow, tabularx}
\usepackage{algorithm}
\usepackage{algorithmic}
\usepackage{balance}
\usepackage{upquote}
\usepackage{physics}
\usepackage{stmaryrd}
\usepackage{wrapfig}
\usepackage{subcaption}
\usepackage{scalerel}

\usepackage{chngpage}

\newtheorem{app_definition}{Definition}
\newtheorem{app_theorem}{Theorem}

\newtheorem{app_corollary}{Corollary}

\newcommand\tab[1][1cm]{\hspace*{#1}} %
\newcommand\smalltab[1][1mm]{\hspace*{#1}} %

\usepackage{xspace}

\newcommand{\hgg}[1]{\textcolor{black}{#1\xspace}}

\newcommand{\jmlr}[1]{\textcolor{black}{#1\xspace}}
\newcommand{\jmlrr}[1]{\textcolor{black}{#1\xspace}}

\def\mcirc{\mathbin{\scalerel*{\bigcirc}{t}}}

\makeatletter
\renewcommand\@makefntext[1]{\noindent\makebox[1.8em][r]{\@thefnmark\,}#1}
\makeatother

\jmlrheading{26}{2025}{1-\pageref{LastPage}}{12/22; Revised
11/24}{2/25}{22-1428}{Hyungyu Lee, Saehyung Lee, Ho Bae, and Sungroh Yoon}
\ShortHeadings{Regularizing Hard Examples Improves Adversarial Robustness}{Lee, Lee, Bae, and Yoon}
\firstpageno{1}

\begin{document}

\title{\jmlr{Regularizing Hard Examples Improves Adversarial Robustness}}

\author{Hyungyu Lee
\email rucy74@snu.ac.kr \\
\addr Electrical and Computer Engineering \\
Interdisciplinary Program in Artificial Intelligence \\ 
Seoul National University, Seoul 08826, Republic of Korea \\
\name Saehyung Lee 
\email halo8218@snu.ac.kr \\
\addr Electrical and Computer Engineering \\
Interdisciplinary Program in Artificial Intelligence \\ 
Seoul National University, Seoul 08826, Republic of Korea \\
\name Ho Bae*
\email hobae@ewha.ac.kr \\
\addr Department of Cyber Security \\
Ewha Womans University, Seoul 03760, Republic of Korea \\
\name Sungroh Yoon*
\email sryoon@snu.ac.kr \\
\addr Electrical and Computer Engineering \\
Interdisciplinary Program in Artificial Intelligence \\ 
AIIS, ASRI, INMC, and ISRC, Seoul National University \\
Seoul National University, Seoul 08826, Republic of Korea \\
}

\renewcommand*{\thefootnote}{\fnsymbol{footnote}}
\footnotetext[1]{Corresponding authors}

\editor{Quanquan Gu}

\maketitle

\input{01_abstract}

\input{1_introduction}

\input{2_background}
\input{3_analysis}

\input{4_method}

\input{5_experiment}
\input{6_conclusion}

\acks{
This work was supported by the National Research Foundation of Korea (NRF) grants funded by the Korea government (Ministry of Science and ICT, MSIT) (2022R1A3B1077720 and 2022R1A5A708390811), by the Institute of Information \& Communications Technology Planning \& Evaluation (IITP) grants funded by the Korea government (MSIT) (NO.RS-2021-II211343 for the Artificial Intelligence Graduate School Program at Seoul National University and RS-2024-004398199 for AI-Based Automated Vulnerability Detection and Safe Code Generation), by the Korea Evaluation Institute of Industrial Technology (KEIT) grants funded by the Korea government (MOTIE), and by the BK21 FOUR program of the Education and Research Program for Future ICT Pioneers, Seoul National University in 2025.
}

\bibliography{reference}

\newpage

\appendix

\input{B_proofs}
\input{C_implementations}

\input{D_RelatedWork}

\input{E_ablation}

\input{F_comparison}

\end{document}

%% file: math_commands.tex
\usepackage{amsmath,amsfonts,bm}

\def\eqref#1{equation~\ref{#1}}

\def\1{\bm{1}}

\DeclareMathAlphabet{\mathsfit}{\encodingdefault}{\sfdefault}{m}{sl}
\SetMathAlphabet{\mathsfit}{bold}{\encodingdefault}{\sfdefault}{bx}{n}

\DeclareMathOperator*{\argmax}{arg\,max}

%% file: 01_abstract.tex
\begin{abstract}

Recent studies have validated that pruning hard-to-learn examples from training improves the generalization performance of neural networks (NNs). In this study, we investigate this intriguing phenomenon---the negative effect of hard examples on generalization---in adversarial training. Particularly, we theoretically demonstrate that the increase in the difficulty of hard examples in adversarial training is significantly greater than the increase in the difficulty of easy examples. Furthermore, we verify that hard examples are only fitted through memorization of the label in adversarial training. 
\jmlr{
We conduct both theoretical and empirical analyses of this memorization phenomenon, showing that pruning hard examples in adversarial training can enhance the model's robustness. However, the challenge remains in finding the optimal threshold for removing hard examples that degrade robustness performance.
}
Based upon these observations, we propose a new approach, difficulty proportional label smoothing (DPLS), to \jmlr{adaptively} mitigate the negative effect of hard examples, thereby improving the adversarial robustness of NNs. Notably, our experimental result indicates that our method can successfully leverage hard examples while circumventing the negative effect. 

\end{abstract}


%% file: 1_introduction.tex
\section{Introduction}
\label{introduction}

The structural regularities of classification datasets have been investigated in several studies~\citep{importance_cscore,importance_el2n,importance_blockdrop}, and measures have been proposed for identifying the regularities of training samples through training statistics. For example, \citet{importance_cscore} discovered that the learning speed of training examples is strongly correlated with the structural regularities, i.e., regular examples are learned quickly, whereas irregular examples are learned slowly. \citet{importance_el2n} measured the difficulty of a training example by using the loss gradient norm.
Based on the proposed measures, the aforementioned studies clarified the relationship between example difficulty and generalization in the context of deep neural networks (DNNs). In particular, a compelling finding among their observations motivates our study:
 \emph{memorizing hard-to-learn/irregular examples can deteriorate the generalization ability of DNNs.}
In other words, hard-to-learn examples are unrepresentative outliers of a class \hgg{or data with corrupted labels}, and thus, these examples can result in the degradation of generalization. In this study, we explore this intriguing phenomenon---the negative effect of hard examples on generalization---in adversarial settings.

\input{fig/fig_introduction_acc_from_pgd}

Adversarial training differs from standard training: it uses adversarial examples as the training data. Notably, adversarial training requires more data than standard training~\citep{adv_more_data}. Additionally, the improvement in generalization rendered by using more data is observably greater in an adversarial setting than in a standard setting~\citep{defense_oat,gowal_ddpm}. Therefore, recent studies on adversarial training have focused on approaches for using more data effectively to bridge the gap between train and test inferences of DNNs \citep{defense_rst,gowal_ddpm}. Nevertheless, in this study, we demonstrate that the improvement in generalization performance could be achieved by regularizing the use of data. Specifically, we demonstrate that the improvement achieved by regularizing the \emph{training of hard examples} is greater in adversarial training.

Fig.~\ref{fig:introduction_testacc_from_pgd} illustrates the difference between the effect of pruning hard examples in standard training (denoted as STD) and adversarial training by using projected gradient descent~\citep{attack_pgd} (denoted as PGD).
After training models on subsets of CIFAR-10 \citep{dataset_cifar} for 100 epochs, we either pruned hard examples from the training dataset or continued by training on the entire training data.\footnote{
Example difficulty is measured by accumulating 0-1 loss along the training trajectory. For further details, refer to Appendix~\ref{appendix:implementation_experiment_details_figure1}.}
We evaluated the test accuracy for clean examples in the STD models and the test accuracy for adversarial examples in the PGD models. Observably, pruning hard examples during training can improve the generalization performance, which is consistent with the previous study's result that training of hard examples can deteriorate generalization. However, the result indicates that the performance increase is greater in adversarial training. 
It can be inferred that the negative effect of hard examples on training is intensified by adversarial perturbations, and thus, the improvement by pruning hard examples is more substantial in adversarial training.

In this study, we verify that the cause of this result is the negative effect of hard examples in adversarial training.
We theoretically prove that the difficulty of each example increases in an adversarial setting and that the increase is more significant for hard examples than for easy examples.
\jmlr{
In this context, we demonstrate that hard examples are fitted only through memorization in adversarial training and that pruning them from the training process improves robustness. Based on this, we conduct a theoretical analysis of the memorization phenomenon of hard examples, proving that including hard examples in adversarial training reduces robustness performance. To demonstrate this empirically, we perform analytical experiments on the assumptions and outcomes of the theoretical analysis, showing that the theoretical analysis aligns with actual results. Consequently, we show that hard examples negatively impact the model's robustness. However, the challenge remains that finding the optimal difficulty threshold for removing hard examples that deteriorate robustness is an intractable proposition.
}

To address this problem, we propose a regularization method, difficulty proportional label smoothing (DPLS), to mitigate the negative effect of hard examples. Label smoothing (LS)~\citep{arch_inception} regularizes the prediction of each example to prevent the memorization of the one-hot label. In this perspective, DPLS extends this LS approach to adopt a smoothing factor that is proportional to the difficulty of the example for regularizing the training of each example. Using the results of experiments on a variety of datasets and algorithms, we assess our proposed methodology and find that DPLS could successfully leverage hard examples while avoiding the negative effect.

\jmlr{
In Section \ref{related work:related work}, we introduce the background and previous studies related to the topic addressed in this paper. In Section \ref{analysis:analysis}, we conduct a theoretical and empirical analysis of learning hard examples in adversarial training. In Section \ref{method:method}, we analyze approaches for effectively learning hard examples in adversarial training and propose a methodology. In Section \ref{experiment:experiment}, we demonstrate the effectiveness of the proposed methodology through experiments on various algorithms and datasets. Finally, in Section \ref{conclusion}, we present the conclusions of the paper.
}

%% file: fig/fig_introduction_acc_from_pgd.tex
\begin{wrapfigure}[22]{R}{0.52\linewidth} %
\begin{center}
\vskip -2mm
\raisebox{0pt}[\dimexpr\height-1.5\baselineskip\relax]{\includegraphics[width=0.9\linewidth]{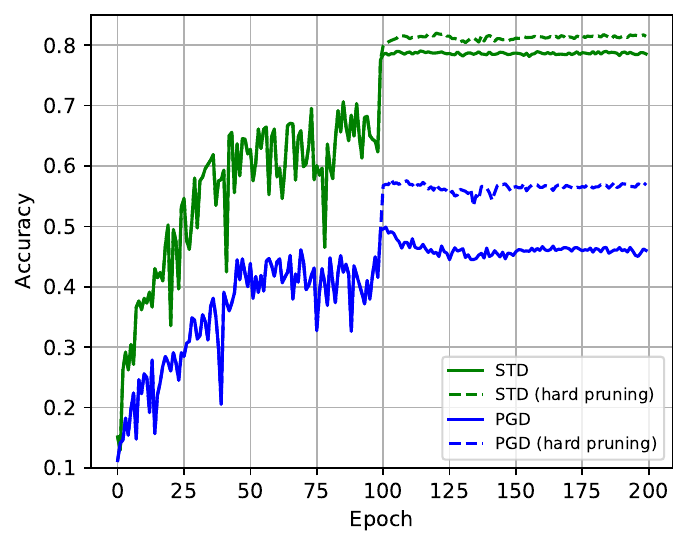}}%
\end{center}
\vskip -8mm
\caption{The test accuracy (for clean examples) of the standard (STD) models and the test accuracy (for adversarial examples) of the adversarial (PGD~\citep{attack_pgd}) models trained on the subsets of CIFAR-10.
On pruning hard examples from training, test accuracy increases both in the STD and PGD models, but the increase in the PGD model is greater than that in the STD model.
}
\label{fig:introduction_testacc_from_pgd}
\end{wrapfigure}

%% file: 2_background.tex
\section{Related work}
\label{related work:related work}

\paragraph{Adversarial robustness}
Adversarial training~\citep{attack_pgd} employs adversarial examples as training data to improve robustness. For a dataset $\mathcal{D}=\{(\bm{\mathnormal{x}}_i, {\mathnormal{y}}_i)\}_{i=1}^n$, where $\bm{\mathnormal{x}}_i \in \mathbb{R}^{d}$ represents a clean data, and $y_i \in \{1, ..., C\}$ denotes its corresponding label of $C$ classes. Adversarial training can be formulated in an empirical risk minimization form featuring the following optimization:

\begin{equation} \label{background_eq:adversarial training}
    \min_{\theta} {\sum_{i=1}^{n}}
    \max_{\mathbf{\delta} \in S} \mathcal{L}(f(\bm{\mathnormal{x}_i + \bm{\delta}}),\mathnormal{y_i};\theta{}).
\end{equation}
Here, $\theta$ denotes the parameter of a model $f$, $\mathcal{L}$ represents the loss function, and $S=\{\bm{\delta} \in \mathbb{R}^{d} : ||\bm{\delta}|| \leq \epsilon\}$ where $\epsilon$ is an adversarial budget, which depicts an upper bound of adversarial perturbations. Another form of adversarial training is TRADES~\citep{defense_trades}, which controls the trade-off between robustness and standard accuracy by dividing training loss into the loss of clean examples and regularization for robustness. For adversarial attacks, fast gradient sign method (FGSM) is a general method for generating adversarial examples~\citep{attack_fgsm}. PGD is a multi-step version of FGSM~\citep{attack_pgd}. Adaptive auto attack ($\mathrm{A^3}$) is an efficient and reliable attack which utilizes adaptive random starts and statistics-based discarding strategy~\citep{attack_aaa}.

\paragraph{\jmlr{Robust and non-robust feature}}

\jmlr{
In \citet{adv_odds_with_accuracy}, certain data features were defined as moderately or slightly correlated with the data labels. These features are useful for improving performance when training a standard classifier, but they have the characteristic of reducing the classifier's performance because they can be manipulated by adversarial attacks. Such features are referred to as `non-robust features.' In contrast, features that are not manipulated by adversarial attacks and maintain their correlation with the data labels are referred to as `robust features.' \citet{adv_not_bugs_are_features} formulated a mathematical definition of robust and non-robust features based on their correlation with the labels. Specifically, for a data pair $(x,y)$ from a dataset $\mathcal{D}$, a feature $g$ is defined as a $\rho$-useful feature if it satisfies $\mathbb{E}[y \cdot g(x)] \geq \rho$, where $\rho > 0$. Here, for an adversarial perturbation $\delta$, a useful feature $g$ is defined as a robust feature if it satisfies $\mathbb{E}\left[\inf_{\delta} y \cdot g(x + \delta)\right] \geq \rho$, and as a non-robust feature if it does not satisfy this condition. We conduct a theoretical analysis utilizing these concepts of robust and non-robust features as defined in previous studies.
}

\paragraph{Overfitting and memorization in adversarial training}
\label{background:robust overfitting}
\citet{overfitting_adv} conducted analysis on overfitting in adversarial training, termed robust overfitting.  \citet{overfitting_adv_kd} utilized a smoothing method that combines knowledge distillation~\citep{smoothing_kd} and stochastic weight average~\citep{smoothing_swa}. \citet{overfitting_label} also used a knowledge distillation method to mitigate overfitting caused by distribution shift and label noise.
\citet{overfitting_mlcat} proposed a method to prevent overfitting of small-loss data combined with adversarial weight perturbation~\citep{defense_awp}.
In~\citet{overfitting_sat,overfitting_reweight,overfitting_adv_memorization}, they used the exponential moving average of predictions as a label for training to prevent robust ovefitting. In particular, \citet{overfitting_adv_memorization} demonstrated that DNNs are sufficient to memorize training adversarial examples with random labels; reportedly the causes of robust overfitting may be the memorization of one-hot labels.

\paragraph{Measurement on difficulty of examples}

\citet{importance_el2n} demonstrated that the norm of the error vector (EL2N score) can be used to identify important and difficult training data. They discovered that examples with high score tend to be hard to learn but important for improving the performance. However, pruning certain examples with the highest score can improve generalization performance, indicating that the highest score examples tend to be unrepresentative.
\citet{importance_cscore} proposed consistency score (C-score), which is the expected accuracy of each instance for the models trained with various data subsets excluding the given instance. While these scores can be utilized in several applications~\citep{cscore_application2}, we use these scores as the example difficulty in our study.

%% file: 3_analysis.tex
\section{Analysis}
\label{analysis:analysis}

In this section, we conduct a theoretical and empirical analysis of the effect of hard examples on training. Through this analysis, we reveal that in adversarial training, learning from hard examples primarily occurs through memorization. Next, we perform a theoretical analysis of the memorization of hard examples that occurs in adversarial training. Finally, we demonstrate the theoretical analysis through various empirical verifications of the memorization of hard examples.

\subsection{Analysis on the effect of hard examples}
\label{method:effect of hard examples}

We first demonstrate through a theoretical analysis of hard examples that training hard examples becomes more difficult as the adversarial budget in adversarial training increases, and we illustrate this theoretical analysis with experimental evidence. Next, we compare the learning patterns of hard examples in standard and adversarial training, confirming that in adversarial training, the learning of hard examples primarily occurs through memorization.

\subsubsection{Theoretical analysis of the effect of hard examples}

\input{fig/fig_method_theory_complement}

We theoretically analyze the effect of hard examples in both standard and adversarial training. A simple model is designed to demonstrate the manner in which adversarial perturbations affect hard examples in adversarial training.
We aim to design examples featuring different difficulties in learning. Specifically, we define a binary classification model $f(\bm{\mathnormal{x}})=\sigma(\bm{w}^\top \bm{\mathnormal{x}}+\jmlr{b})$, which is a mapping function $f:\mathcal{X}\rightarrow\mathcal{Y}$ from input space $\mathcal{X}$ to output space $\mathcal{Y}$, where $(\bm{\mathnormal{x}},y) \in \mathbb{R}^d \times \{\pm1\}$ and $\sigma(\cdot)$ represents a sigmoid function. To ensure simplicity, we assume that $\norm{\bm{w}}=1$.
Thereafter, we set the loss function as cross-entropy loss and define easy and hard examples by controlling the distance between each example and the decision boundary. The hard example is set as $(\bm{\mathnormal{x}}_1, y_1=+1)$ where $\norm{\bm{w}^\top \bm{\mathnormal{x}}_1+\jmlr{b}}=d_1$, and the easy example is set as $(\bm{\mathnormal{x}}_2, y_2=+1)$ where $\norm{\bm{w}^\top \bm{\mathnormal{x}}_2+\jmlr{b}}=d_2$ and $d_1 < d_2$. The differences between the gradient norm of the examples in standard and adversarial settings are as follows:

\begin{definition}[Gradient difference]
The differences between the loss gradient norm on $\bm{x}_1$ and $\bm{x}_2$ in standard and adversarial training are defined as follows:
\begin{equation}
\begin{gathered}
    \mathcal{D}_{\textup{std}}(\bm{\mathnormal{x}}_1, \bm{\mathnormal{x}}_2)=\norm{(1-\sigma(\bm{w}^\top \bm{\mathnormal{x}}_1+\jmlr{b}))\bm{\mathnormal{x}}_1} -
    \norm{(1-\sigma(\bm{w}^\top \bm{\mathnormal{x}}_2+\jmlr{b}))\bm{\mathnormal{x}}_2}\\
    \mathcal{D}_{\textup{adv}}(\bm{\mathnormal{x}}_1, \bm{\mathnormal{x}}_2, \epsilon)=\norm{(1-\sigma(\bm{w}^\top (\bm{\mathnormal{x}}_1+\bm{\delta})+\jmlr{b}))(\bm{\mathnormal{x}}_1+\bm{\delta})} - 
    \norm{(1-\sigma(\bm{w}^\top (\bm{\mathnormal{x}}_2+\bm{\delta})+\jmlr{b}))(\bm{\mathnormal{x}}_2+\bm{\delta})}.
\end{gathered}
\end{equation}

\end{definition}
\label{eq:method_theorem_overfitting}

Here, $\bm{\delta}$ represents an adversarial perturbation, where $\norm{\bm{w}^\top \bm{\delta}} = \epsilon$. $\mathcal{D}_{\textup{std}}$ denotes the difference between the loss gradient norm on inputs in standard training and $\mathcal{D}_{\textup{adv}}$ denotes that in adversarial training.
The following theorem and corollary demonstrate that the difference between the loss gradient norm of hard and easy examples in adversarial training is more significant than that in standard training:

\begin{theorem}
\label{eq:method_corollary_overfitting}
Let the gradient norm of easy examples is smaller than the gradient norm of hard examples in both standard and adversarial training. Then, if $\bm{w}^\top \bm{\mathnormal{x}}_1 \geq 0$ and  $0<\epsilon < \frac{(d_1 +d_2)}{2}$, it satisfies
\begin{equation}
\mathcal{D}_{\textup{std}}(\bm{\mathnormal{x}}_1, \bm{\mathnormal{x}}_2) <  \mathcal{D}_{\textup{adv}}(\bm{\mathnormal{x}}_1, \bm{\mathnormal{x}}_2, \epsilon).    
\end{equation}
\end{theorem}

\begin{corollary}
\label{eq:method_corollary}
The increase in gradient difference by an adversarial attack is proportional to the difficulty of the example and the size of the upper bound of adversarial perturbations. Thus, if $\norm{\bm{w}^\top \bm{\mathnormal{x}}_3}=d_3$ and $d_2 < d_3$, it satisfies
\begin{equation}
\mathcal{D}_{\textup{adv}}(\bm{\mathnormal{x}}_1, \bm{\mathnormal{x}}_2, \epsilon) <  \mathcal{D}_{\textup{adv}}(\bm{\mathnormal{x}}_1, \bm{\mathnormal{x}}_3, \epsilon),
\end{equation}
and if $\epsilon_1 < \epsilon_2$, it satisfies
\begin{equation}
\mathcal{D}_{\textup{adv}}(\bm{\mathnormal{x}}_1, \bm{\mathnormal{x}}_2, \epsilon_1) <  \mathcal{D}_{\textup{adv}}(\bm{\mathnormal{x}}_1, \bm{\mathnormal{x}}_2, \epsilon_2).    
\end{equation}
\end{corollary}

We have reported the proof in Appendix~\ref{appendix:proof3.1}. In the model whose training loss is sufficiently converged, the outputs of easy examples are saturated, where $\sigma(\bm{w}^\top \bm{\mathnormal{x}}_2) \approx 1$.
Thus, the loss gradient norm of easy examples in standard training is assumed to be similar to that in adversarial training. The assumption is validated through empirical analysis.
The change in the gradient norm of hard examples through the addition of adversarial perturbation is larger than that of easy examples. 
\hgg{According to~\citet{importance_el2n}, the gradient norm can be used as the difficulty (importance) score. In other words, the difficulty increment of hard examples is more significant than that of easy examples.
}

\subsubsection{Empirical analysis of the effect of hard examples} 

The preliminary analysis verifies that the increase in the difficulty of hard examples is greater than the increase in that of easy examples in adversarial training, and we demonstrated that hard examples exerts a negative effect on generalization (Fig.~\ref{fig:introduction_testacc_from_pgd}).
We accordingly hypothesize that a major factor deteriorating generalization in adversarial training is the large increase in the difficulty of hard examples.
We empirically verify this hypothesis.

We select hard examples according to the average loss of the models along the training trajectory in accordance with the strategy reported by \citet{importance_cscore}.
The loss can be of any kind, which is a measure of the error of each example. We employ 0-1 loss as the measurement of difficulty. 
For a clear representation, the model trained with standard training is denoted as STD, and the model trained via adversarial training as PGD. We represent the accuracy for clean examples as standard accuracy and the accuracy for adversarial examples as robust accuracy.

\input{fig/fig_method_norm_ratio}

\paragraph{Comparing the difficulty of hard examples} We initially conduct an empirical study on the difficulty of hard examples by comparing the gradient norm in standard and adversarial training.
\citet{importance_el2n} reported that the average error vector norm (EL2N score) can approximate the expected training loss gradient norm of each example.
Thus, we employ this to compare the gradient norm in both training.
We select the top 10k hard and top 10k easy example subsets from the training dataset of CIFAR-10 and subsequently calculate the average error vector norm of examples in each subset during training, where the error vector norm of an input pair $(\bm{\mathnormal{x}},y)$ is defined as $\mathbb{E}\norm{f(\bm{\mathnormal{x}}) - \bm{\mathnormal{y}}}_2$ for the one-hot label vector $\bm{\mathnormal{y}}$.
We train several models by varying the adversarial budget $\epsilon=8$ (PGD), $4, 2$, and $0$ (STD), and the result is obtained after the learning rate decay epoch, which is the training point indicating that the model is sufficiently trained.

As illustrated in Fig.~\ref{fig:method_norm_ratio} (left), the average norm values of easy examples are small in all models, and the difference in the average norm of easy examples between models is also comparatively small. However, for the results of hard examples, the average norm of hard examples increases significantly as the adversarial budget $\epsilon$ increases. In the PGD model (right), greater difference is observed between clean and adversarial examples of hard examples than that of easy examples.
This result indicates that the loss gradient norm of hard examples is rendered greater in adversarial training while that of easy examples does not, which demonstrates our theoretical analysis presented in Section~\ref{method:effect of hard examples}.
\paragraph{Comparing the training of hard examples}

The difficulty of hard examples is significantly amplified in adversarial training. We therefore investigate whether hard examples are fitted normally in adversarial training.
Figs.~\ref{fig:method_acc_from_std} and \ref{fig:method_acc_from_pgd} show the test accuracy and training accuracy of the top 10k hard, top 10k easy, and entire training examples in the STD and PGD models. Observably, test accuracy decreases marginally following the learning rate decay (epoch 100) in the STD model. In contrast, the result of the PGD model indicates that when the training robust accuracy of hard examples starts to increase (at the decay epoch 100), the test robust accuracy starts to decrease rapidly.
Moreover, the training accuracy of the easy examples remains unchanged after the decay, indicating that the decrease in the test performance is mostly attributed to the training of hard examples.

\input{fig/fig_method_demo}

Fig.~\ref{fig:method_train_testacc_from_all} shows the training accuracy of hard examples and the test accuracy of the models with varying adversarial budget $\epsilon$. The decrease in test accuracy and the increase in training accuracy of hard examples after the learning rate decay are observed to be severe as the $\epsilon$ increases. 
\jmlr{
Therefore, given these characteristics in the learning patterns, we can regard this phenomenon as the memorization of hard examples that occurs in adversarial training, in contrast to standard training. Specifically, as the adversarial budget $\epsilon$ increases in adversarial training, it suggests that an increasing number of hard examples are being learned through memorization.
}

\paragraph{\jmlr{Pruning hard examples from training}}

\input{table/tab_method_tradeoff}

\jmlr{
We observed the phenomenon of hard examples being memorized after the learning rate decay, as shown in Fig. \ref{fig:method_acc_from_pgd}. Next, we conducted experiments to verify whether the hard examples fitted through memorization actually contribute to the model's performance. To do this, as in Fig. \ref{fig:introduction_testacc_from_pgd}, we trained the model after excluding the hard examples from the training dataset and then compared the performance. Specifically, we pruned the top 10k easy and top 10k hard examples from the training dataset of CIFAR-10, respectively.
}

\jmlr{
Table \ref{tab:method_tradeoff} shows the standard and robust accuracies of the PGD (None) and PGD pruning models (Easy and Hard pruning). From the robust accuracy results, we can see that excluding easy examples from the training dataset results in the expected decrease in robust accuracy. In contrast, when hard examples are excluded, robust accuracy increases. This suggests that learning hard examples may reduce the model's robustness.
}
\jmlr{
However, when focusing on standard accuracy, we observe that excluding hard examples from the training process leads to a decrease in standard performance. Taking this into consideration, learning hard examples appears to increase standard accuracy but decrease robust accuracy. This characteristic resembles the robustness-accuracy trade-off observed by \citet{adv_odds_with_accuracy}. Therefore, it implies that the model may be learning non-robust features through the memorization of hard examples.
} 

\subsection{\jmlr{Theoretical analysis on the training of hard examples}}
\label{method:theoretical training of hard examples}

\jmlr{
In Section \ref{method:effect of hard examples}, we demonstrated that learning hard examples is challenging in adversarial training and observed that these hard examples are primarily learned through memorization. Subsequently, through hard example pruning experiments, we observed that the memorization of hard examples exhibits patterns similar to the learning of non-robust features. In this section, we conduct a theoretical analysis of this phenomenon in adversarial training. Specifically, we perform a theoretical analysis of the effects of learning through the memorization of hard examples in a binary classification task. We investigate the impact of hard examples on the model's performance in adversarial training, and through this, we theoretically demonstrate that including hard examples in the training process leads to a decrease in robustness performance.
}

\paragraph{\jmlr{Definition of data distributions in binary classification task}}

\jmlr{
We first define the task for the theoretical analysis of learning through the memorization of hard examples. Following \citet{adv_odds_with_accuracy} and \citet{adv_not_bugs_are_features}, we define a binary classification task consisting of the following input-output pair $(\bm{\mathnormal{x}}, y)$.
}

\jmlr{
\begin{definition}[Normal example data distribution]
\begin{equation} 
\label{eq:method_gaussian_distribution_default}
\begin{gathered}
    \mathnormal{y}\stackrel{u.a.r.}{\sim} \{-1, +1\}, 
    \tab
    \bm{\mathnormal{x}}  \in \mathbb{R}^{d+2} =
    \begin{cases}
    x_{0} {\sim} \mathcal{N}(\eta y, 1),
    \\
    x_{i \in \{1,...,d\}}\stackrel{i.i.d.}{\sim} \mathcal{N}(\beta{\eta y}, 1), 
    \\
    x_{d+1}= 0,
    \end{cases}
\end{gathered}
\end{equation}
\end{definition}
}
\jmlr{
where $\beta$ is a constant that satisfies $0 < \beta < 1$. Here, assuming $\beta{\eta} < \epsilon < \eta$, $x_{0}$ corresponds to a robust feature, while $x_{i \in \{1,...,d\}}$ correspond to non-robust features. Additionally, we introduce a `memorization feature' corresponding to $x_{d+1}$. The memorization feature represents a feature that exists only in specific data, specifically in hard examples, but does not exist in normal examples. If a model learns a memorization feature, it can be considered that the model has memorized it. We define that data sampled from the normal example data distribution does not exhibit memorization features. Therefore, we refer to examples without memorization features sampled from the distribution of Equation \ref{eq:method_gaussian_distribution_default} as a `normal example.' Here, the set of datasets that can be generated from this distribution is defined as $\mathcal{X}_{normal}$.
}

\jmlr{
Next, we define the distribution of hard examples in the above binary classification task. In Section \ref{method:effect of hard examples}, we observed that hard examples are learned through memorization. Additionally, we observed that this memorization seems to involve the learning of non-robust features. Based on this, we make the following assumptions about hard examples.
}

\jmlr{
\newtheorem{assumption}{Assumption} 
\begin{assumption} 
\label{eq:method_assumption_hard_example}
\textnormal{\textbf{(Properties of hard examples from the perspective of adversarial robustness)}}
    A hard example, difficult for the model to learn, consists only of non-robust and memorization features. The memorization feature functions as a robust feature during the training phase but acts as a non-robust feature for unseen data during the test phase.
\end{assumption}
}

\jmlr{
These assumptions describe hard examples that become more difficult to learn as the adversarial budget increases, aligning them with the properties of non-robust features. Their learning through memorization corresponds to \emph{memorization features}, which act as robust features during training by avoiding zero assignment in adversarial training \citep{adv_odds_with_accuracy}. However, during testing on unseen data, these features behave as non-robust features, increasing standard accuracy but decreasing robust accuracy. Based on these assumptions, we define the distribution of hard examples.
}
\jmlr{
\begin{definition}[Hard example data distribution]
\begin{equation} 
\label{eq:method_hard_example}
\begin{gathered}
    \mathnormal{y}\stackrel{u.a.r.}{\sim} \{-1, +1\}, 
    \smalltab \smalltab \smalltab \smalltab
    \bm{\mathnormal{x}}_{hard}  \in \mathbb{R}^{d+2} =
    \begin{cases}
    x_{0} =0,
    \\
    x_{i \in \{1,...,d\}}\stackrel{i.i.d.}{\sim} \mathcal{N}(\beta{\eta y}, 1), 
    \\
    x_{d+1}=
        \begin{cases}
        \mathcal{N}(\eta y, 1), & \textit{training phase} \\
        \mathcal{N}(\beta \eta y, 1), & \textit{test phase, unseen data} \\
        \end{cases}
    \\
    \end{cases}
\end{gathered}
\end{equation}
\end{definition}
}

\jmlr{
The hard example data distribution is characterized by the robust feature $x_0$ taking a value of 0. The memorization feature, while being a non-robust feature, is defined to behave like a robust feature during training. We define the set of datasets that can be generated from this distribution as $\mathcal{X}_{hard}$.
}

\paragraph{\jmlr{Analysis on the effect of training hard examples}}

\jmlr{
Next, we compare the performance of a binary classification model trained on datasets sampled from the data distributions introduced above. To train the model, we sample $m$ normal examples to form the training dataset $\mathcal{X}_{normal}^{train} \subset \mathcal{X}_{normal}$ and sample $n$ hard examples to construct a hard example training set $\mathcal{X}_{hard}^{train} \subset \mathcal{X}_{hard}$. We then examine the case where the model is trained using the combined dataset $\mathcal{X}_{normal}^{train} \cup \mathcal{X}_{hard}^{train}$, consisting of $m+n$ data points. For simplicity, we redefine the model $f$ as $f = \text{sign}(\bm{w}^\top \bm{\mathnormal{x}} + b)$ and assume $\norm{\bm{x}} \leq 1$. Here, the weight $\bm{w}_{adv}$ of the adversarially trained model $f_{adv}$ (with $\beta{\eta} < \epsilon < \eta$) is learned, and the model's standard and robust accuracies are expressed as follows.
}
\jmlr{
\begin{equation} 
\label{eq:method_std_adv_weight_hard}
\begin{gathered}
    \begin{cases}
    w_{0} = \frac{m}{{Z}_{adv}(m+n)},
    \\
    w_{i \in \{1,...,d\}} = 0, 
    \\
    w_{d+1}=\frac{n}{{Z}_{adv}(m+n)},
    \\
    \end{cases}    
    \begin{cases}
    {Pr}\big[{f}_{adv}(\bm{x})=y\big]=Pr\bigg[\frac{1}{{Z}_{adv}}\bigg(\frac{m}{m+n} \cdot \mathcal{N}(\eta, 1)\bigg)+by > 0\bigg], \\
    {Pr}\big[{f}_{adv}(\bm{x}+\bm{\delta})=y\big]=Pr\bigg[\frac{1}{{Z}_{adv}}\bigg(\frac{m}{m+n} \cdot \mathcal{N}(\eta - \epsilon, 1)\bigg)+by > 0\bigg],
    \end{cases}
\end{gathered}
\end{equation}
}
\jmlr{
Here, $Z_{adv}$ is a constant that satisfies $\norm{\bm{w}} = 1$. The model's performance is measured on a subset of the normal example data distribution, $\mathcal{X}_{normal}^{test} \subset (\mathcal{X}_{normal} \setminus \mathcal{X}_{normal}^{train})$, which does not include the training data. Since normal examples do not have memorization features, the results are as above. Then, based on Equation \ref{eq:method_std_adv_weight_hard}, the following theorem holds when $m > 0$.
}

\jmlr{
\begin{theorem} 
\label{eq:method_theorem_acc_comparison}
\textnormal{\textbf{(Degradation of normal test accuracy with increasing hard examples in training)}}
For a model $f_{adv}$ adversarially trained on the combined dataset $\mathcal{X}_{normal}^{train} \cup \mathcal{X}_{hard}^{train}$, where $\mathcal{X}_{normal}^{train}$ consists of $m$ normal examples and $\mathcal{X}_{hard}^{train}$ consists of $n$ hard examples, both the standard accuracy $\text{Pr}\big[{f}_{adv}(\bm{x})=y\big]$ and the robust accuracy $\text{Pr}\big[{f}_{adv}(\bm{x}+\bm{\delta})=y\big]$ on the normal example test dataset $\mathcal{X}_{normal}^{test}$ are monotonically decreasing with respect to $n$.
\end{theorem}
}
\jmlr{
The above theorem demonstrates that including additional hard examples during training does not contribute to improving the performance on normal examples in adversarial training. Now, we similarly evaluate the models' performance on hard examples. The performance is measured on a subset of the hard example data distribution, $\mathcal{X}_{hard}^{test} \subset (\mathcal{X}_{hard} \setminus \mathcal{X}_{hard}^{train})$.
}
\jmlr{
\begin{equation} 
\label{eq:method_std_adv_acc_hard_hard}
\begin{gathered}    
    \begin{cases}
    {Pr}\big[{f}_{adv}(\bm{x}_{hard})=y\big]=Pr\bigg[\frac{1}{{Z}_{adv}}\bigg(\frac{n}{m+n} \cdot \mathcal{N}(\beta\eta, 1)\bigg) +by > 0\bigg],\\
    {Pr}\big[{f}_{adv}(\bm{x}_{hard}+\bm{\delta})=y\big]=Pr\bigg[\frac{1}{{Z}_{adv}}\bigg(\frac{n}{m+n} \cdot \mathcal{N}(\beta\eta - \epsilon, 1)\bigg) +by > 0\bigg],
    \end{cases}
\end{gathered}
\end{equation}
}

\jmlr{
Hard examples do not contain robust features, and their performance is determined by the model weight corresponding to the memorization feature. Then, based on Equation \ref{eq:method_std_adv_acc_hard_hard}, the following theorem holds when $n > 0$.
}
\jmlr{
\begin{theorem} 
\label{eq:method_theorem_acc_comparison_hard}
\textnormal{\textbf{(Trade-off in hard test accuracy with increasing hard examples in training)}}
For a model $f_{adv}$ adversarially trained on the combined dataset $\mathcal{X}_{normal}^{train} \cup \mathcal{X}_{hard}^{train}$, where $\mathcal{X}_{normal}^{train}$ consists of $m$ normal examples and $\mathcal{X}_{hard}^{train}$ consists of $n$ hard examples, the standard accuracy $\text{Pr}\big[{f}_{adv}(\bm{x})=y\big]$ on the hard example test dataset $\mathcal{X}_{hard}^{test}$ is monotonically increasing with respect to $n$, but the robust accuracy $\text{Pr}\big[{f}_{adv}(\bm{x}+\bm{\delta})=y\big]$ is monotonically decreasing with respect to $n$.
\end{theorem}
}

\jmlr{
To summarize Theorem \ref{eq:method_theorem_acc_comparison} and Theorem \ref{eq:method_theorem_acc_comparison_hard}, including hard examples in adversarial training can lead to either an increase or decrease in standard accuracy. This is because the performance on normal examples decreases, while the performance on hard examples increases. However, robust accuracy always decreases because performance declines across all types of examples. The derivation process of the equations introduced in this section, as well as the proofs of the theorems and corollaries, can be found in Appendix \ref{appendix:proof3.2}.
}

\subsection{\jmlr{Empirical verification on the training of hard examples}}
\label{method:empirical verification of hard examples}
\jmlr{
In this section, we conduct an empirical analysis to validate the assumptions and theorems discussed in Section \ref{method:theoretical training of hard examples}. We first verify the assumptions regarding hard examples, and then we empirically examine the theorems.
}
\subsubsection{\jmlr{Empirical verification of the properties of hard examples}}
\label{method:empirical verification of assumption}

\jmlr{
We proceed with the validation of Assumption \ref{eq:method_assumption_hard_example}. If the assumption holds, the properties of a model learning hard examples are characterized as follows.
}
\jmlr{
\begin{enumerate}    
    \item \textbf{Memorization}: For the hard example dataset $\mathcal{X}_{hard}^{train}$, the model $f_{adv, hard}$, which was adversarially trained with $\mathcal{X}_{hard}^{train}$ included, exhibits high robust accuracy on $\mathcal{X}_{hard}^{train}$. However, the model $f_{adv, \setminus hard}$, which was adversarially trained without $\mathcal{X}_{hard}^{train}$, shows nearly 0\% robust accuracy on unseen $\mathcal{X}_{hard}^{train}$.
    \item \textbf{Infeasibility of adversarial training}: Standard training is possible with only hard examples, but adversarial training is not. Specifically, let $\bar{f}$ be the model trained solely on $\mathcal{X}_{hard}^{train}$. Then, $Pr\big[\bar{f}_{std}(\bm{x})\big] > Pr\big[\bar{f}_{adv}(\bm{x})\big] > \frac{1}{C}$, while $Pr\big[\bar{f}_{adv}(\bm{x}+\bm{\delta})\big] < \frac{1}{C}$, where $C$ is the number of class.
    \item \textbf{Trade-off}: As the proportion of hard examples in the training dataset increases, the test robust accuracy decreases, while the test standard accuracy increases.
\end{enumerate}
}
\input{fig/fig_method_memorization}

\paragraph{\jmlr{Property 1: Memorization}}

\jmlr{
Based on Assumption \ref{eq:method_assumption_hard_example}, when a hard example dataset $\mathcal{X}_{hard}^{train}$ is included in adversarial training, the memorization feature acts as a robust feature for the hard examples included in the training. As a result, even when adversarial attacks are applied to these hard examples, the memorization feature enables high robust accuracy on these examples. In contrast, if the same dataset $\mathcal{X}_{hard}^{train}$ is not included in adversarial training, the memorization feature acts as a non-robust feature for these hard examples or may not have been learned at all. In this case, when adversarial attacks are applied to these unseen hard examples, the memorization feature cannot be utilized, leaving no available features for prediction, resulting in 0\% accuracy. To illustrate this, we conduct the following experiment.
}

\jmlr{
For the CIFAR-10 dataset, the hard example data subset $\mathcal{X}_{hard}^{train}$ is composed of the top 5k hard examples. Two models are trained: one including the top 5k hard examples (${f}_{adv,hard}$) and the other excluding them (${f}_{adv,\setminus hard}$). The dataset including the hard examples was trained using all 50k CIFAR-10 data points, while the dataset excluding the hard examples was trained using the remaining 45k data points after removing the top 5k hard examples. After training, the performance of both models was evaluated on the top 5k hard examples $\mathcal{X}_{hard}^{train}$. Additionally, for comparison with normal examples, a normal example data subset was composed of the top 5k easy examples, and the same experiments were conducted.
}

\jmlr{
Fig.~\ref{fig:method_memorization} compares the memorization property in adversarial training. Observably, normal examples show significant accuracy on data not included in the training, even if they were not part of the training process. In contrast, when hard examples are included in the training, the model can achieve up to 100\% accuracy on these hard examples. However, when they are not included in the training, the model shows 0\% accuracy. This demonstrates that, unlike normal examples, hard examples exhibit the memorization property. Additional details regarding Fig.~\ref{fig:method_memorization} can be found in Appendix \ref{appendix:implementation_experiment_memorization_and_pruning}.
}

\paragraph{\jmlr{Property 2: Infeasibility of adversarial training}}

\jmlr{
From the Assumption \ref{eq:method_assumption_hard_example}, when adversarial training is conducted using only hard examples, the model only learns the memorization feature; it leads to the robust accuracy that is lower than random prediction due to the memorization feature acting as a non-robust feature. Conversely, standard accuracy is higher than random prediction because of the same memorization feature. In contrast, standard training, which learns both non-robust and memorization features, will show higher standard accuracy compared to the adversarially trained model. To illustrate this, we conduct the following experiment.
}

\input{table/tab_method_subset}

\jmlr{
We construct a hard example data subset using the top 10k hard examples and a normal example data subset using the top 10k easy examples. Then, instead of using the entire dataset, we train models using only these respective subsets through standard training and adversarial training to compare their performance. We measure the standard (Std.) and robust (Rob.) accuracy for the standard (STD) and adversarial (PGD) models. 
}

\jmlr{
In Table~\ref{tab:method_subset}, the STD model trained with the hard example data subset exhibits performance comparable to that of the STD model trained with the easy example data subset. On the other hand, the PGD model trained with the hard example data subset shows lower standard accuracy than the STD model. As expected, in terms of robust accuracy, the model performs worse than random prediction (10\%). Therefore, it can be concluded that standard training is possible with only hard examples, but adversarial training is not. Additional details can be found in Appendix \ref{appendix:implementation_experiment_memorization_and_pruning}.
}

\paragraph{\jmlr{Property 3: Trade-off}}

\input{fig/fig_method_noise_only}

\jmlr{
We have already demonstrated in Table \ref{tab:method_tradeoff} through the hard examples pruning experiment that learning hard examples exhibits the robustness-accuracy trade-off. To further support this, we additionally conduct a label corruption (shuffling) experiment, where the labels of hard examples are randomly changed. When label corruption is applied to hard examples, the corresponding weight of the memorization feature is learned incorrectly but does not become zero. Consequently, while this does not increase standard accuracy, the vulnerability of the non-robust feature remains, so unlike with pruning, robust accuracy is expected to remain unchanged rather than increase.
}

\jmlr{
We construct a hard example data subset using the top 5k hard examples and a normal example data subset using the top 5k easy examples. We replace only the labels of data in each subset with random labels. We then conduct adversarial training on each of these datasets.
}

\jmlr{
The models trained using the dataset with corrupted labels for normal and hard examples are shown in Fig.~\ref{fig:method_corruption_plot}, along with their standard (left) and robust (right) accuracy curves. Examining the results for standard accuracy (left), it is evident that performance decreases for both normal and hard examples when label corruption is applied. In the results for robust accuracy (right), it is observed that while the performance of the models with corrupted normal examples consistently declines during training, the performance of the model with corrupted hard examples behaves similarly to that of the PGD model (None). This indicates that even when the labels of hard examples are corrupted, they do not significantly affect the robust accuracy of the adversarial training model. Additional details regarding Fig.~\ref{fig:method_corruption_plot} can be found in Appendix \ref{appendix:implementation_experiment_memorization_and_pruning}.
}

\jmlr{
Consequently, we empirically verify that the various characteristics that can arise in the learning of hard examples, as suggested by Assumption \ref{eq:method_assumption_hard_example}, appear as expected. Therefore, we confirm that Assumption \ref{eq:method_assumption_hard_example}—which posits that hard examples consist only of non-robust features and memorization features, with the memorization feature acting as a robust feature in the training data and as a non-robust feature in the test data—is supported to a certain degree.
}
\subsubsection{\jmlr{Empirical verification of the training of hard examples}}

\input{fig/fig_method_excluding}

\jmlr{
In this section, we conduct an empirical analysis of Theorem \ref{eq:method_theorem_acc_comparison} and Theorem \ref{eq:method_theorem_acc_comparison_hard}, which provide a theoretical analysis of the effect of increasing the number of hard examples in the training data during adversarial training. We empirically verify the theory through further experiments related to Fig. \ref{fig:introduction_testacc_from_pgd} and Table \ref{tab:method_tradeoff}. We prune the top $n\%$ easy examples ($\subset \mathcal{X}_{normal}$) and the top $n\%$ hard examples ($\subset \mathcal{X}_{hard}$) from the training dataset of CIFAR-10, respectively ($0 \leq n \leq 35$). Figs.~\ref{fig:method_std_excluding} and \ref{fig:method_pgd_excluding} show the standard and robust accuracies of the PGD pruning models.
}

\jmlr{
In Fig.~\ref{fig:method_pgd_excluding}, pruning hard examples improves the robust accuracy. This aligns with the theoretical analysis presented in Section \ref{method:theoretical training of hard examples}, which suggests that the robustness of adversarial training models decreases as the number of hard examples in the training dataset increases. In contrast to hard examples, pruning normal examples leads to a decrease in performance. This outcome is likely due to the increased proportion of hard examples in the overall dataset.
}

\jmlr{
Fig.~\ref{fig:method_all_excluding} demonstrates that the improvement from pruning hard examples decreases as the adversarial budget \(\epsilon\) decreases. Therefore, the performance improvement observed by pruning hard examples from the training dataset can be considered a characteristic of adversarial training.
}

\jmlr{
Through Theorem \ref{eq:method_theorem_acc_comparison} and Theorem \ref{eq:method_theorem_acc_comparison_hard}, we observed that increasing the number of hard examples in the training dataset decreases the standard accuracy on normal examples but increases the standard accuracy on hard examples. Fig.~\ref{fig:method_std_excluding} further confirms that the increase in standard accuracy is indeed predominant. To verify this, we conduct a theoretical analysis comparing the increase and decrease in standard accuracy as presented in Theorem \ref{eq:method_theorem_acc_comparison} and Theorem \ref{eq:method_theorem_acc_comparison_hard}. As noted in Corollary \ref{eq:app_corollary_acc_comparison_hard} in Appendix \ref{appendix:proof3.3}, when the number of normal examples is sufficiently greater than the number of hard examples, increasing the number of hard examples in the training dataset leads to an increase in overall standard accuracy, as observed in Fig.~\ref{fig:method_std_excluding}. Since typical datasets are assumed to have more normal examples than hard examples, this condition appears to be probable.
}

%% file: fig/fig_method_theory_complement.tex
\begin{wrapfigure}[12]{R}{0.3\linewidth} %
\begin{center}
\vskip -2mm
\raisebox{0pt}[\dimexpr\height-1.5\baselineskip\relax]{\includegraphics[width=1.0\linewidth]{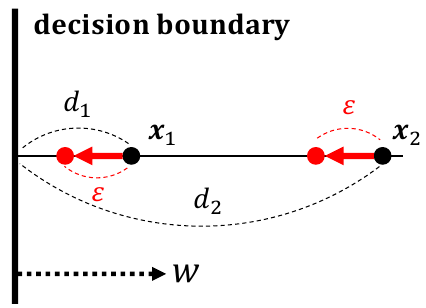}}%
\end{center}
\vskip -2mm
\caption{The schematic illustration of easy ($x_2$) and hard examples ($x_1$).}
\label{fig:method_theory_complement}
\end{wrapfigure}

%% file: fig/fig_method_norm_ratio.tex
\begin{wrapfigure}[11]{R}{0.6\linewidth} %
\begin{center}
\vskip 2mm
\raisebox{0pt}[\dimexpr\height-1.5\baselineskip\relax]{\includegraphics[width=1.0\linewidth]{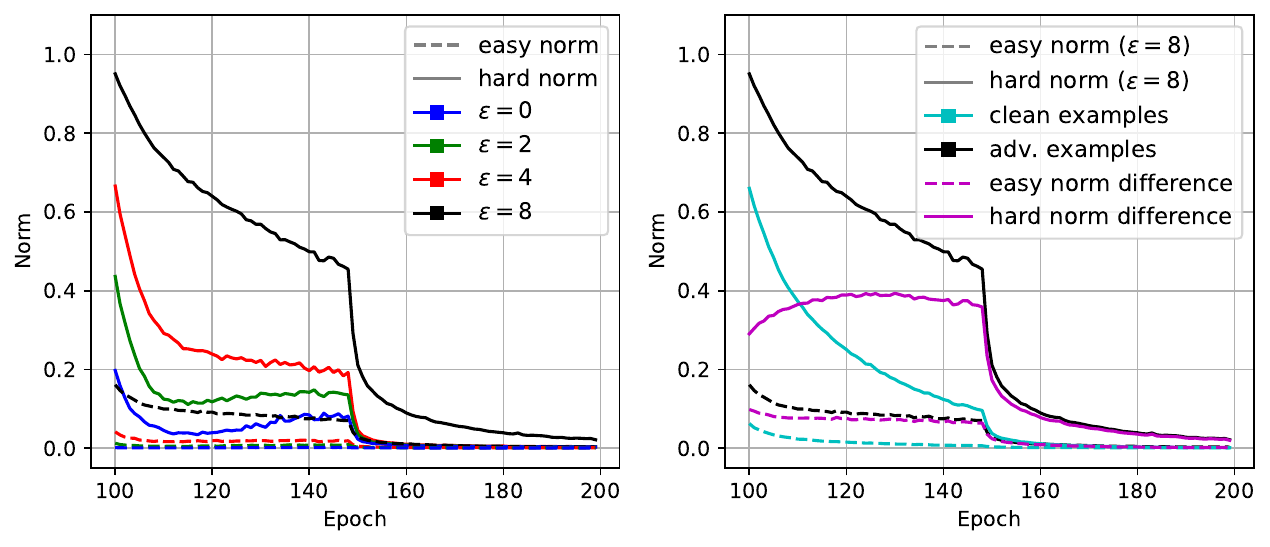}}%
\end{center}
\vskip -5mm
\caption{\hgg{Average error vector norm (EL2N score~\citep{importance_el2n}) of training examples.}
}
\label{fig:method_norm_ratio}
\end{wrapfigure}

%% file: fig/fig_method_demo.tex
\begin{figure}[t]
  \centering
  \begin{subfigure}[b]{0.32\linewidth}
    \includegraphics[width=\linewidth]{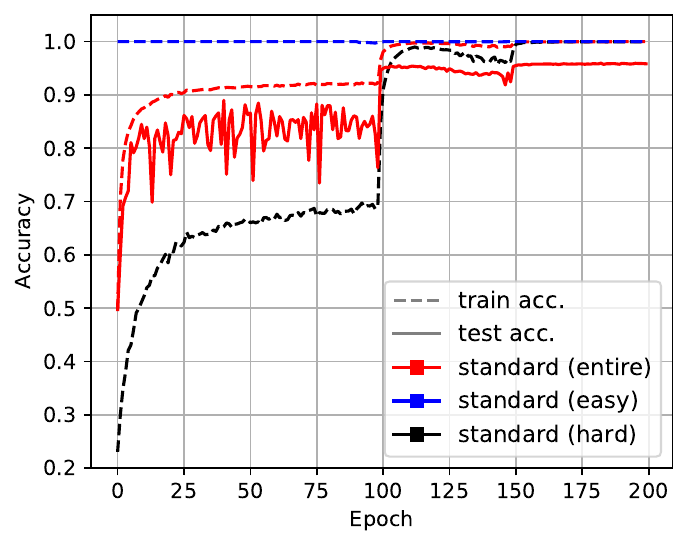}
    \caption{}
    \label{fig:method_acc_from_std}
  \end{subfigure}
  \hfill
  \begin{subfigure}[b]{0.32\linewidth}
    \includegraphics[width=\linewidth]{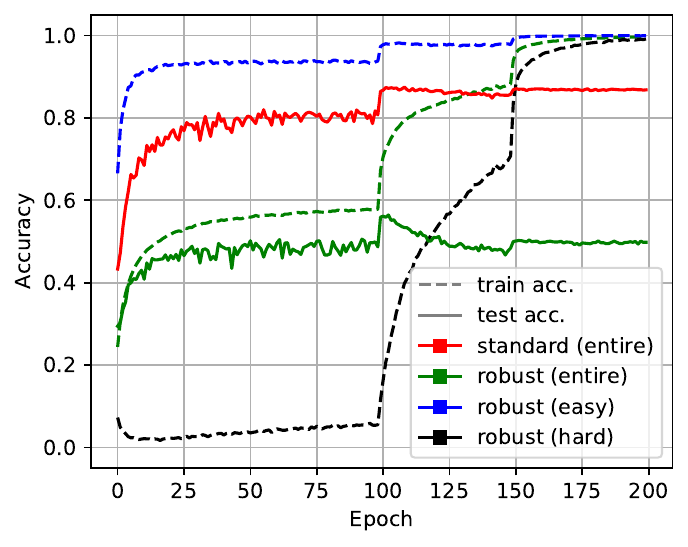}
    \caption{}
    \label{fig:method_acc_from_pgd}
  \end{subfigure}
  \hfill
  \begin{subfigure}[b]{0.32\linewidth}
    \includegraphics[width=\linewidth]{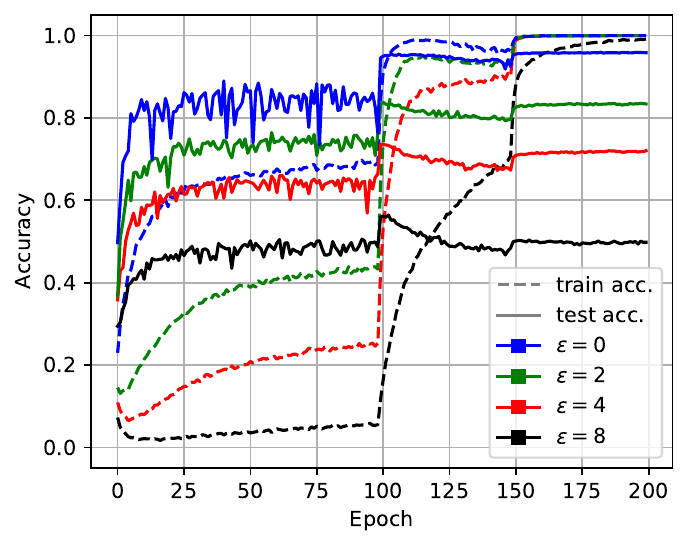}
    \caption{}
    \label{fig:method_train_testacc_from_all}
  \end{subfigure}
  \caption{\hgg{(a) and (b) show the training and test accuracy curves in the STD and PGD models, respectively. (c) shows the training accuracy of hard examples and test accuracy of the models with different adversarial budgets $\epsilon$.}}
  \label{fig:method_demo}
\end{figure}

%% file: table/tab_method_tradeoff.tex
\begin{wraptable}[9]{r}{0.3\linewidth}
\centering
\footnotesize
\addtolength{\tabcolsep}{-2pt}
 \vskip -0mm
\caption{\jmlr{Performance of models with and without pruning 10k hard or easy examples.}}
\vskip -2mm
\begin{tabular}{c|ccc}
    \toprule
     \multicolumn{1}{c|}{\multirow{1}{*}{Model}} &
     \multicolumn{1}{c}{\multirow{1}{*}{Std.}} & \multicolumn{1}{c}{\multirow{1}{*}{Rob.}} \\
    \midrule[.05em]
    \jmlr{None} & \jmlr{87.19} & \jmlr{51.80} \\
     \jmlr{Easy pruning} & \jmlr{87.27} & \jmlr{49.71} \\
     \jmlr{Hard pruning} & \jmlr{84.70} & \jmlr{53.36} \\
    \midrule[.1em]
\end{tabular}
\label{tab:method_tradeoff}
\end{wraptable}

%% file: fig/fig_method_memorization.tex
\begin{wrapfigure}[15]{R}{0.42\linewidth} %
\begin{center}
\vskip -4mm
\raisebox{0pt}[\dimexpr\height-1.5\baselineskip\relax]{\includegraphics[width=1.0\linewidth]{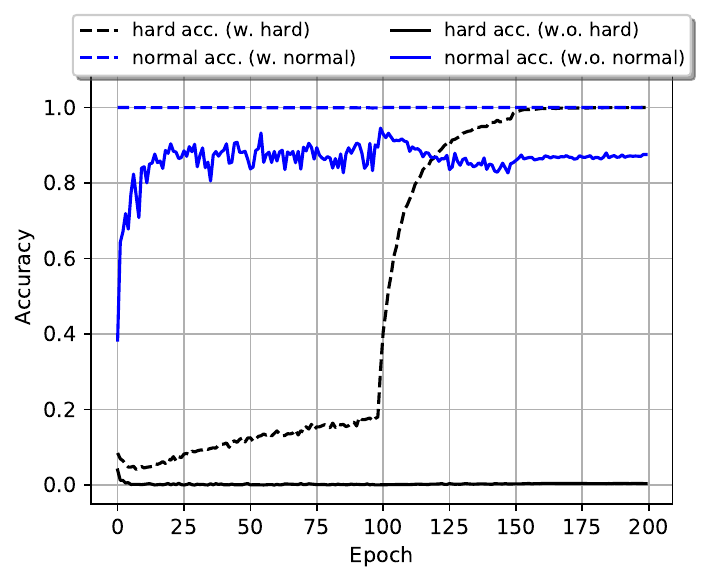}}%
\end{center}
\vskip -6mm
\caption{\jmlr{Comparison of the memorization property in adversarial training. Only hard examples exhibit the memorization property.}
}
\label{fig:method_memorization}
\end{wrapfigure}

%% file: table/tab_method_subset.tex
\begin{wraptable}[7]{r}{0.3\linewidth}
\centering
\footnotesize
\addtolength{\tabcolsep}{-2pt}
 \vskip -6mm
\caption{\jmlr{Performance of models trained with 10k examples.}}
\vskip -2mm
\begin{tabular}{c|ccc}
    \toprule
     \multicolumn{1}{c|}{\multirow{2}{*}{Method}} &
     \multicolumn{1}{c}{\multirow{2}{*}{STD}} & \multicolumn{2}{c}{\multirow{1}{*}{PGD}} \\
     & & Std. & Rob. \\
    \midrule[.05em]
     \jmlr{Normal} & \jmlr{78.46} & \jmlr{61.63} & \jmlr{39.56} \\
     \jmlr{Hard} & \jmlr{77.60} & \jmlr{19.92} & \jmlr{8.68} \\
    \midrule[.1em]
\end{tabular}
\label{tab:method_subset}
\end{wraptable}

%% file: fig/fig_method_noise_only.tex
\begin{wrapfigure}[12]{R}{0.64\linewidth} %
\begin{center}
\vskip -2.5mm
\raisebox{0pt}[\dimexpr\height-1.5\baselineskip\relax]{\includegraphics[width=1.0\linewidth]{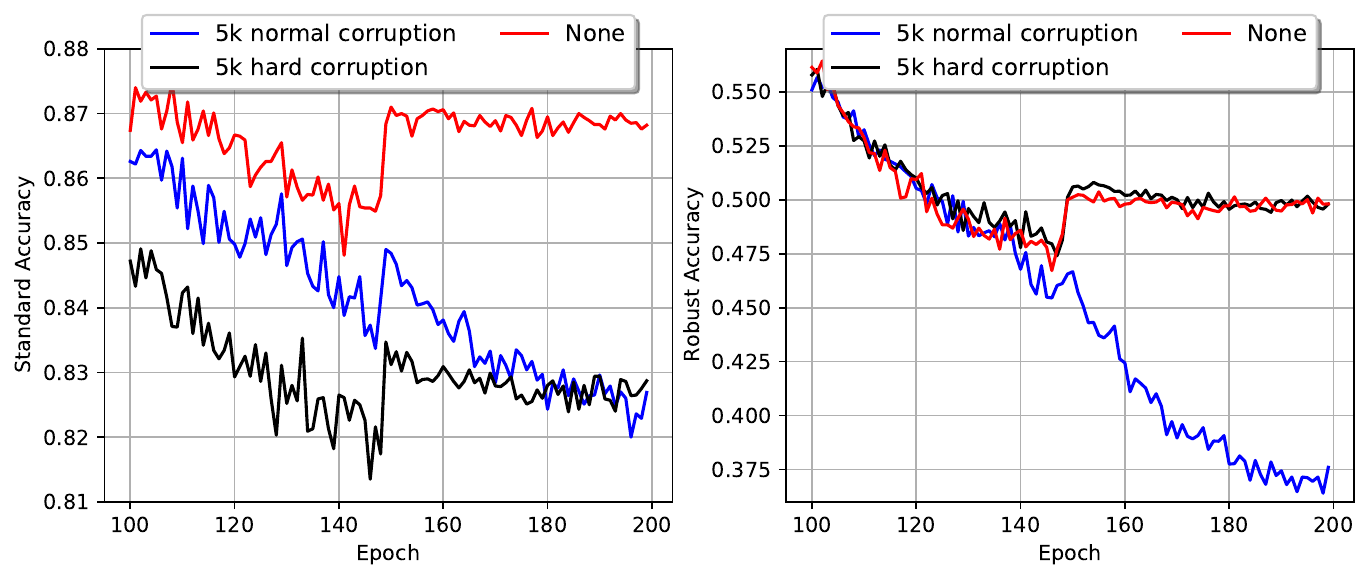}}%
\end{center}
\vskip -6mm
\caption{\jmlr{The standard and robust accuracies of the models whose labels of subsets are corrupted.}
}
\label{fig:method_corruption_plot}
\end{wrapfigure}

%% file: fig/fig_method_excluding.tex
\begin{figure}[t]
  \centering
  \begin{subfigure}[b]{0.32\linewidth}
    \includegraphics[width=\linewidth]{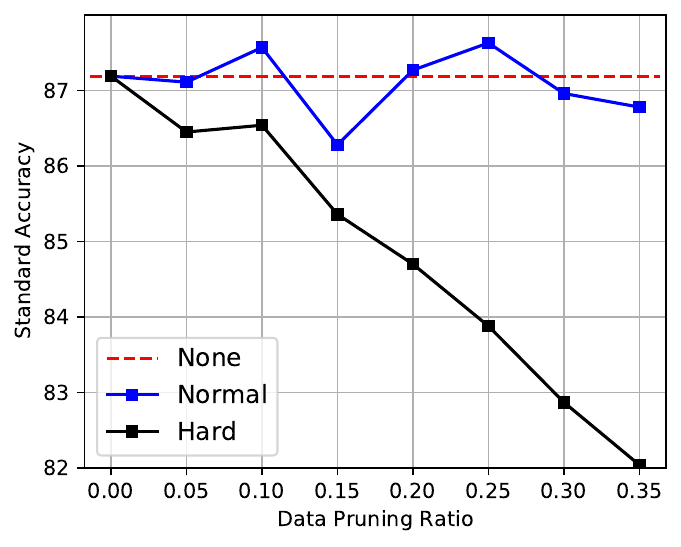}
    \caption{}
    \label{fig:method_std_excluding}
  \end{subfigure}
  \hfill
  \begin{subfigure}[b]{0.32\linewidth}
    \includegraphics[width=\linewidth]{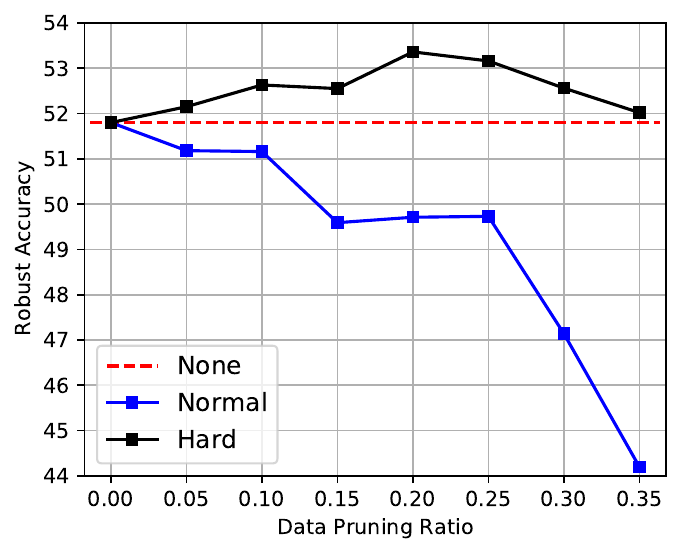}
    \caption{}
    \label{fig:method_pgd_excluding}
  \end{subfigure}
  \hfill
  \begin{subfigure}[b]{0.32\linewidth}
    \includegraphics[width=\linewidth]{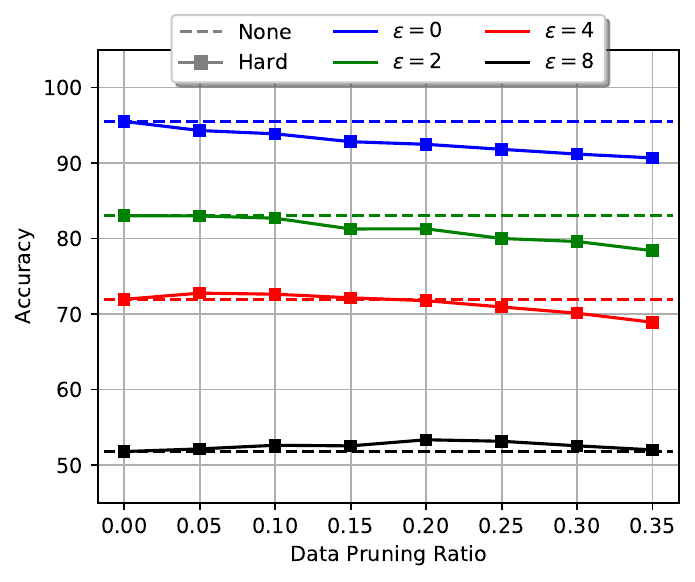}
    \caption{}
    \label{fig:method_all_excluding}
  \end{subfigure}
  \caption{\hgg{(a) and (b) show the standard and robust accuracies of STD and PGD models where subset of training data is pruned. (c) shows the results that repeat the experiment (a) or (b) for different adversarial budgets.}}
  \label{fig:method_excluding}
\end{figure}

%% file: 4_method.tex
\section{Method}
\label{method:method}

\jmlr{
In this section, we explore approaches to mitigate the negative effects of learning hard examples. We evaluate the effectiveness of methodologies that regularize the learning of hard examples based on label smoothing (LS). Through this evaluation, we find that basic regularization techniques like label smoothing are inadequate for regularizing the learning of hard examples. To address this, we theoretically demonstrate that the appropriate direction for label smoothing regularization in the context of hard examples is to allocate the label smoothing factor in proportion to the difficulty of the hard examples. Accordingly, we propose Difficulty Proportional Label Smoothing (DPLS), a method that properly regularizes hard examples and enables effective adversarial training.
}

\subsection{Mitigation of the negative effect of hard examples} 
\label{method:mitigation of the effect of hard examples}

Although training hard examples impairs the performance in adversarial training, simply pruning hard examples from training is not the optimal solution. 
\jmlr{
In Section \ref{method:theoretical training of hard examples}, we defined normal and hard examples as disjoint distributions, but in real data, this may not be the case. As shown in Fig. \ref{fig:method_pgd_excluding}, pruning the top 20\% of hard examples improves performance, but beyond that point, performance starts to decline. Moreover, according to Fig. \ref{fig:method_all_excluding}, the point at which performance degradation begins varies depending on the adversarial budget $\epsilon$. These results suggest that finding the appropriate hard example pruning point depends on the type of dataset and the adversarial budget, making the task of identifying the optimal difficulty threshold for removing hard examples that deteriorate robustness an intractable proposition.
}

Thus, instead of pruning hard examples, we applied label smoothing (LS)~\citep{arch_inception} to subsets of the dataset to mitigate the negative effect of hard examples.
\jmlr{
Label smoothing (LS) is a technique aimed at reducing model overconfidence by assigning a value of $\lambda$ to the correct label while distributing the value $1-\lambda$ uniformly across all labels. Here, $\lambda$ represents the label smoothing factor and lies within the range $[\frac{1}{C}, 1]$, where $C$ is the number of class. Following the above pruning experiment, we applied LS ($\lambda=0.9$) to the top $n\%$ of hard examples during training and then evaluated the model's performance.
}

Figures.~\ref{fig:method_pgd_ls} depict the test robust accuracy of the PGD LS models. As a result of pruning models, applying LS to the hard example subset increases robust accuracy, but applying LS to the normal example subset decreases performance. 
\jmlr{
Fig. \ref{fig:_method_exclude_and_compare_exclude_ls} shows that the increase in robust accuracy is similar to that observed with pruning. According to Fig.~\ref{fig:method_pgd_ls_clean}, only marginal decreases are observed in standard accuracy. This implies that LS produces effects similar to pruning; however, the decrease in standard accuracy is less pronounced compared to pruning.
Nevertheless, considering the performance drop observed when LS is applied to normal examples, it is expected that continuing to increase the proportion of hard examples to which LS is applied will eventually lead to a decrease in performance. Consequently, similar to pruning, determining the optimal difficulty threshold for applying LS to hard examples that negatively impact robustness is a challenging and complex task.
}

\input{fig/fig_method_ls}

\subsection{Our method to mitigate the negative effect of hard examples} 
\label{method:our method to mitigate the negative effect of hard examples}

\jmlr{
Since it is difficult to accurately measure and define the difficulty of data, it is challenging to precisely determine whether a given data point is a hard example that does not contain robust features. We can expect that there will be data points on the boundary between hard and normal examples, depending on the difficulty of learning the data. Therefore, considering this difficulty, we define an input-label pair distribution that includes both hard and normal examples as follows.
}
\jmlr{
\begin{definition}[Data distribution with difficulty]
\begin{equation} 
\label{eq:method_hard_example_difficulty}
\begin{gathered}
    \mathnormal{y}\stackrel{u.a.r.}{\sim} \{-1, +1\}, 
    \tab
    \bm{\mathnormal{x}}  \in \mathbb{R}^{d+2} =
    \begin{cases}
    x_{0} =\mathcal{N}(\alpha \eta y, 1),
    \\
    x_{i \in \{1,...,d\}}\stackrel{i.i.d.}{\sim} \mathcal{N}(\beta{\eta y}, 1), 
    \\
    x_{d+1}=
        \begin{cases}
        \mathcal{N}(\gamma\eta y, 1), & \textit{training phase} \\
        \mathcal{N}(\beta \eta y, 1), & \textit{test phase, unseen data} \\
        \end{cases}
    \\
    \end{cases}
\end{gathered}
\end{equation}
\end{definition}
}
\jmlr{
Here, $0 \leq \alpha \leq 1$ represents the difficulty of the data, with each data point having a different value. This reflects the difficulty through the coefficient of the robust feature, with lower difficulty corresponding to a higher value of $\alpha$. When $\alpha \eta \leq \epsilon$, the data point can be considered a hard example. $0 \leq \gamma \leq 1$ represents the coefficient of the memorization feature, which is inversely related to the robust feature coefficient $\alpha$; when $\gamma \eta > \epsilon$, the data point is classified as a hard example.
}

\jmlr{
Next, we explore approaches to mitigate the negative effects of hard examples (data with low $\alpha$) when training a model using a dataset composed of data points with varying levels of difficulty. Unlike pruning, LS sacrifices standard accuracy less and also allows for adjusting the intensity of the method through the LS factor $\lambda$ ($\lambda \in [\frac{1}{C}, 1]$). Therefore, we explore an approach that adaptively mitigates the negative effects of hard examples by leveraging these characteristics of LS.
}

\jmlr{
It is established that during adversarial training, the optimal weight configuration for achieving high robust accuracy is one where the weight values corresponding to features other than the robust feature are zero, i.e., $\bm{w}^{*} = (1, 0, 0, \dots)$. Therefore, when training a model using a dataset composed of data with varying levels of difficulty, we explore an LS approach that aims to train the model's weight $\bm{w}$ to be as close as possible to $\bm{w}^{*}$. Let $\mathcal{X}_{difficulty}$ denote the set of datasets that can be generated by sampling from Equation \ref{eq:method_hard_example_difficulty}, and let the training dataset be $\mathcal{X}_{difficulty}^{train} \subset \mathcal{X}_{difficulty}$. Then, the optimal LS factor for each data point that trains the model's weight $\bm{w}$ to be close to $\bm{w}^{*}$ during adversarial training with $\mathcal{X}_{difficulty}^{train}$ follows the theoretical principles outlined below.
}
\jmlr{
\begin{theorem}
\label{eq:method_theorem_ls_factor}
Let the robust feature coefficients of each data point in the dataset $\mathcal{X}_{difficulty}^{train}$, consisting of $k$ data points $\bm{x}_1, \bm{x}_2, \dots, \bm{x}_k$, be denoted as $\alpha_1, \alpha_2, \dots, \alpha_k$. Then, when performing adversarial training using $\mathcal{X}_{difficulty}^{train}$, the optimal LS factors $\lambda_1, \lambda_2, \dots, \lambda_k$ for each training example are proportional to their respective robust feature coefficients $\alpha_1, \alpha_2, \dots, \alpha_k$.
\end{theorem}
}
\jmlr{
The proof of Theorem \ref{eq:method_theorem_ls_factor} can be found in Appendix \ref{appendix:proof4.2}. Theorem \ref{eq:method_theorem_ls_factor} demonstrates that applying LS in proportion to the robust feature coefficient of each data point during adversarial training is the optimal learning approach for achieving robustness. In other words, this suggests that the optimal LS approach for adversarial training is to decrease the LS factor as the difficulty of the data increases, making the intensity of smoothing regularization proportional to the difficulty. Therefore, based on Theorem \ref{eq:method_theorem_ls_factor}, we propose a regularization approach that adjusts the LS factor for each example during training. 
}

We modify the LS method to ensure that the training of hard examples is regularized, but the training of other examples is less affected.
We use the LS factor that is proportional to the difficulty of each example and control the intensity of regularization, which we term as difficulty proportional label smoothing (DPLS).
The difficulty is calculated by using the average loss of the training models along the training trajectory. We train the model and accumulate the difficulty until the difficulty calculation epoch $T$, and the training progresses with the application of DPLS using the calculated difficulty.
We establish the LS factor of the easiest example as $\lambda=1$ (no smoothing) and the factor of the most difficult example as $\lambda \in [\frac{1}{C}, 1)$, where $C$ represents the number of the class. The procedure of DPLS is described in Algorithms~\ref{algorithm:DPLS} and \ref{algorithm:DPLS_pre}.

Our method maintains the training effect of each training example, but it avoids the negative effect of hard examples. We could utilize several loss types for difficulty measurement including cross-entropy loss, the norm of the error vector (EL2N score)~\citep{importance_el2n}, consistency score (C-score)~\citep{importance_cscore}, and 0-1 loss.
Applying DPLS via any error measure is an effective strategy for mitigating the negative effect of hard examples (refer to Table~\ref{tab:experiment_difficulty_loss}). However, in our experiments, as the 0-1 loss exhibits a higher performance than the other loss types, we mainly use the 0-1 loss as the difficulty measurement loss.
Notably, our method calculates the difficulty using clean examples because the adversarial examples are different from training algorithms and training epochs. 

\input{fig/fig_method_algorithm_dpls_pre}

%% file: fig/fig_method_ls.tex
\begin{figure}[t]
  \centering
  \begin{subfigure}[b]{0.32\linewidth}
    \includegraphics[width=\linewidth]{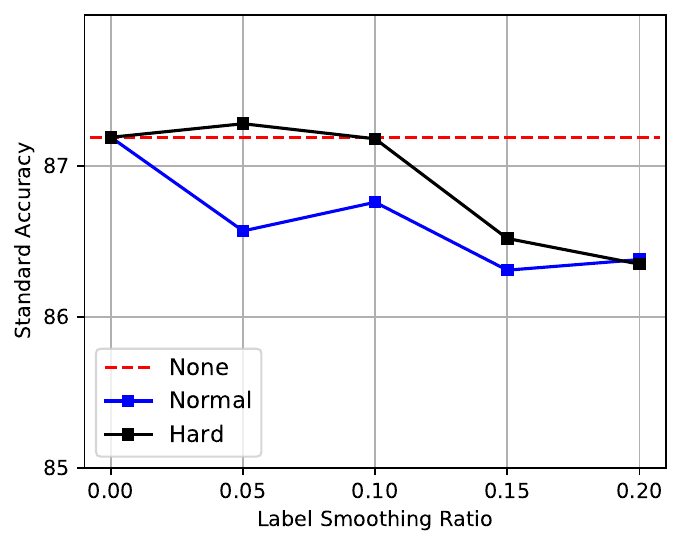}
    \caption{}
    \label{fig:method_pgd_ls_clean}
  \end{subfigure}
  \hfill
  \begin{subfigure}[b]{0.32\linewidth}
    \includegraphics[width=\linewidth]{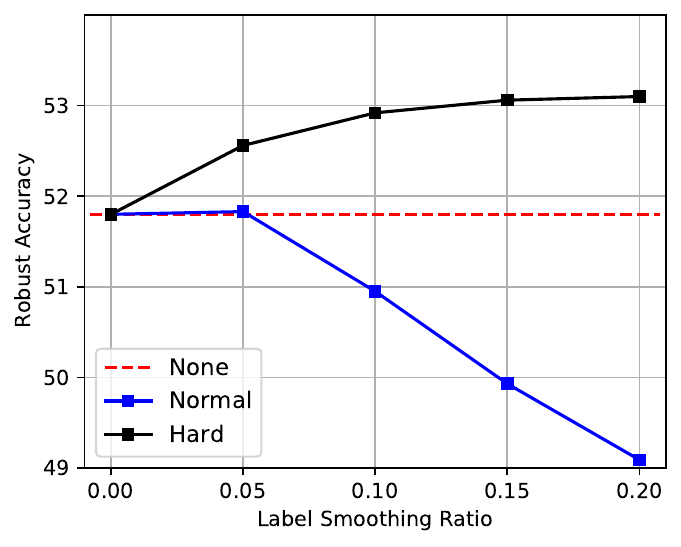}
    \caption{}
    \label{fig:method_pgd_ls}
  \end{subfigure}
  \hfill
  \begin{subfigure}[b]{0.32\linewidth}
    \includegraphics[width=\linewidth]{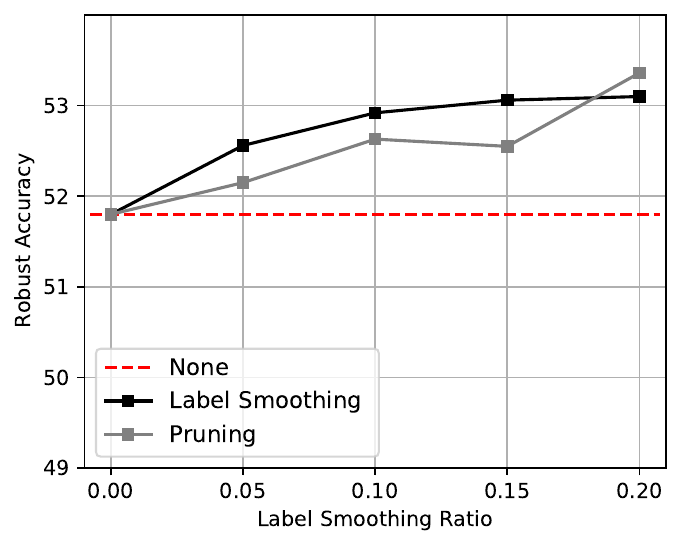}
    \caption{}
    \label{fig:_method_exclude_and_compare_exclude_ls}
  \end{subfigure}
  \caption{\jmlr{(a) the test standard accuracy and (b) robust accuracy of the PGD LS models. (c) the performance comparisons of the PGD LS and pruning models.}}
  \label{fig:method_ls}
\end{figure}

%% file: fig/fig_method_algorithm_dpls_pre.tex
\begin{minipage}{0.5\textwidth}
\begin{algorithm}[H]
\caption{DPLS adversarial training }
\begin{algorithmic}[1]
\label{algorithm:DPLS}
\REQUIRE Dataset $D$, model parameter $\bm{\theta}$, batch size $n$, difficulty calculation epoch $T$, total training epoch $K$, learning rate $\alpha$, difficulty loss storage $\mathcal{S}$
\FOR{$t=1$ \textbf{to} $K$}
\FOR{batch \textbf{in} $D$}
\STATE batch and indices:
\STATE $\{\bm{\mathnormal{x}}_i,\bm{\mathnormal{y}}_i\}^{n}_{i=1}, \mathcal{I} \sim D$
\STATE compute or apply DPLS:
\IF{$t < T$}
\STATE $\mathcal{S} \left[ \mathcal{I} \right] \leftarrow  \{\mathcal{L}_{\textup{difficulty}}(\bm{\mathnormal{x}}_i,\bm{\mathnormal{y}}_i)\}^{n}_{i=1}$
\ELSE
\STATE $\{ \bm{\mathnormal{y}}_i \}^{n}_{i=1} \leftarrow \text{DPLS}(\mathcal{S} \left[ \mathcal{I} \right])$
\ENDIF
\STATE adversarial training:
\STATE \{$\bm{\delta}_i\}^{n}_{i=1} \leftarrow \argmax_{\bm{\delta}_i} \mathcal{L}_{\textup{adv}}(\bm{\mathnormal{x}}_i+\bm{\delta}_i,\bm{\mathnormal{y}}_i)$ 
\STATE $\mathcal{L} \leftarrow \frac{1}{n}\sum_{i=i}^{n}\mathcal{L}_{\textup{adv}}(\bm{\mathnormal{x}}_i+\bm{\delta}_i,\bm{\mathnormal{y}}_i)$
\STATE $\bm{\theta} \leftarrow \bm{\theta} - \alpha\cdot\nabla_{\theta}\mathcal{L}$
\ENDFOR
\ENDFOR
\STATE \textbf{Output:} robust classifier $f$
\end{algorithmic}
\end{algorithm}
\end{minipage}
\hfill
\begin{minipage}{0.41\textwidth}
\begin{algorithm}[H]
\caption{Calculation of DPLS}
\begin{algorithmic}[1]
\label{algorithm:DPLS_pre}
\REQUIRE Dataset $D$, Dataset size $N$, number of class $C$, loss storage $\mathcal{S}$, DPLS factor $\lambda$
\STATE \textbf{Initialization:}
\STATE \quad apply min-max normalization:
\STATE \quad $\mathcal{S} \leftarrow \frac{\mathcal{S} - \textup{min}(\mathcal{S})}{\textup{max}(\mathcal{S})- \textup{min}(\mathcal{S})}$
\FOR{$i=1$ \textbf{to} $N$}
\STATE $(\bm{\mathnormal{x}}_i,\bm{\mathnormal{y}}_i) \leftarrow D[i]$
\STATE apply LS using $\mathcal{S}$:
\STATE $\bm{\mathnormal{y}}_i \leftarrow LS(\bm{\mathnormal{y}}_i, 1 - (1-\lambda)\mathcal{S}[i], C)$
\STATE $ D[i] \leftarrow (\bm{\mathnormal{x}}_i,\bm{\mathnormal{y}}_i)$
\ENDFOR
\STATE \textbf{function} $LS(\bm{y}, \lambda, C)$
\STATE \quad \textbf{for} $i=1$ \textbf{to} $C$ \textbf{do}
\STATE \quad \quad \textbf{if} $\bm{y}[i]=1$ \textbf{then}
\STATE \quad \quad \quad $\bm{y}[i] \leftarrow \lambda$
\STATE \quad \quad \textbf{else}
\STATE \quad \quad \quad $\bm{y}[i] \leftarrow \frac{1 - \lambda}{C-1}$
\STATE \quad \quad \textbf{end if}
\STATE \quad \textbf{end for}
\STATE \quad \textbf{return} $\bm{y}$
\end{algorithmic}
\end{algorithm}
\end{minipage}

%% file: 5_experiment.tex
\input{table/tab_experiment_all110_main_2}

\section{Experiment}
\label{experiment:experiment}

\paragraph{Implementation details}\label{exp:paragrph implementation details}
We conducted experiments on CIFAR-10, CIFAR-100~\citep{dataset_cifar}, SVHN~\citep{dataset_svhn}, \jmlr{and STL-10 \citep{dataset_stl10}} by employing PGD~\citep{attack_pgd} and TRADES~\citep{defense_trades} as the baseline adversarial training algorithms. We utilized WideResNet28-10~\citep{arch_wideresnet} as the architecture of our models. 
We set the LS factor of the most difficult example in DPLS considering the average difficulty score of each dataset and algorithm. We selected the difficulty calculation epoch as $T=90$ for stability, which is 10 epochs before the first learning rate decay.
To evaluate the robustness of the models, we employed the PGD and adaptive auto attack ($A^3$)~\citep{attack_aaa}. Further details are presented in Appendix~\ref{appendix:implementation_details}.

\input{fig/fig_experiment_train_test_acc_from_pgd}

\paragraph{Improvement in robustness}
We compared four mitigation methods: baseline (-), pruning hard examples (pruning), label smoothing (LS), and our proposed method (DPLS).
We pruned the top 10\% of examples with high difficulty for pruning models. For LS models, we applied the same LS factor with the average factor of DPLS. Thus, with the same LS budget, our method assigns a large portion of the budget to the hard examples and LS assigns it uniformly to every example. We trained the baseline model until epoch 90 and continued this training with the application each method.
Table~\ref{tab:experiment_all110_main_2} shows that pruning hard examples and DPLS are effective for mitigating the effect of hard examples and improving the robustness generalization.
The performance improvement of the pruning models demonstrates that the negative effect of hard examples appears in various datasets and algorithms. 
The results of LS models show no improvements in the robustness performance. Contrarily, the DPLS increases the robustness of the models across different training algorithms and datasets for all adversarial budgets $\epsilon$.
The robustness performance difference between LS and DPLS indicates that regularization without considering the effect of each example on training is difficult to improve robust generalization. 

Fig.~\ref{fig:experiment_acc_from_pgd_all} illustrates the training and test robust accuracy results (left) of the PGD and PGD with DPLS (denoted as DPLS) models and training accuracy of top N\% (right) hard examples of the 50,000 training examples of CIFAR-10 (e.g. (0-20\%) denotes the top 10k hard examples).
Observably, comparing DPLS with PGD, the decrease in the training accuracy of the DPLS model coincides with the increase in the test accuracy. DPLS (right) prevents overfitting of hard examples and causes the fast convergence of comparatively easy examples.
This result indicates that our method prevents the memorization of hard examples and mitigate the negative effect of hard examples, thereby leading to decrease in the gap between training and test accuracy by improving the generalization performance.

\paragraph{Comparison with other mitigation methods}

We compared DPLS with the previous robust overfitting mitigation methods: self-adaptive training (SAT)~\citep{overfitting_sat}, knowledge distillation with stochastic weight averaging (KD+SWA)~\citep{overfitting_adv_kd}, and temporal ensemble (TE)~\citep{overfitting_adv_memorization}. We trained the PGD model combined with each method and with total training epoch 200. The details on the settings of each method are described in Appendix~\ref{appendix:implementation_experiment_dpls}.
The results listed in Table~\ref{tab:experiment_overfitting_methods} indicate that all the methods are effective at mitigating robust overfitting.
Among the methods, DPLS exhibits the highest improvement in the test robust accuracy against $A^3$.
Additionally, because our method is orthogonal to SWA~\citep{smoothing_swa}, we trained the model with the method combining DPLS and SWA. Further improvement is observed when combined with SWA.
These results indicate that our method successfully mitigates the memorization of hard examples and prevents overfitting.

\input{table/tab_experiment_collection}

\paragraph{Applying DPLS to other training algorithms}

We applied our method to other adversarial training algorithms to verify that DPLS can consistently increase the robustness generalization performance independent of algorithms. We combined DPLS with three algorithms: adversarial weight perturbation (AWP)~\citep{defense_awp}, misclassification aware adversarial training (MART)~\citep{defense_mart}, and robust self training (RST)~\citep{defense_rst}. We trained each algorithm model combined with TRADES+DPLS and evaluated the trained models.
Table~\ref{tab:experiment_training_algorithms} shows that all training algorithms combined with DPLS yield improvements in robustness. 
For the RST models, we calculated the difficulty of the unlabeled dataset as the ratio of number of training iterations to the number of right predictions.
The result shows that DPLS successfully mitigates the negative effect of hard examples for the unlabeled dataset and improves the robust accuracy for the $A^3$ and PGD attacks.

\paragraph{Various difficulty loss on DPLS}

We compared the loss types of the difficulty for DPLS. Table~\ref{tab:experiment_difficulty_loss} shows the performance of the TRADES models that apply DPLS with several loss functions: C-score, cross-entropy loss (CE loss), EL2N score, and 0-1 loss.
All models in which DPLS was applied exhibited a better performance than the baseline models.
The difficulty calculated from another model (e.g. C-score from STD) was also effective on mitigation.
Any difficulty loss was effective, and DPLS with the 0-1 loss offered the best performance. The results indicate that the 0-1 loss leverages comparatively objective values because the 0-1 loss is less affected by scale or variance at each checkpoint of the model.
As the calculation process can be detached from the training, further improvement is achievable by using the better difficulty measurement loss.
Further experiments and ablation studies on DPLS can be found in Appendix~\ref{appendix:implementation_experiment_dpls} and Appendix~\ref{appendix:ablation}.

%% file: table/tab_experiment_all110_main_2.tex
\begin{table}[t]
\caption{\hgg{Performance of the models for mitigation methods. The best results are indicated in bold. }}
\begin{subtable}[t]{\textwidth}
\centering
\footnotesize %
\addtolength{\tabcolsep}{-1pt} %
\begin{tabular}{cl|ccc||l|ccc}
\toprule
\multicolumn{1}{c}{\multirow{2}{*}{Dataset}} &
\multicolumn{1}{c|}{\multirow{2}{*}{Method}} &
\multirow{2}{*}{Std.} & Robust &  \multicolumn{1}{c||}{Robust} &
\multicolumn{1}{c|}{\multirow{2}{*}{Method}} &
\multirow{2}{*}{Std.} & Robust &  \multicolumn{1}{c}{Robust}
\\
& \multicolumn{1}{c|}{} & & (PGD) & \multicolumn{1}{c||}{($A^3$)} &
\multicolumn{1}{c|}{} & & (PGD) & \multicolumn{1}{c}{($A^3$)}
\\
\midrule[.05em]
\multirow{4}{*}{CIFAR-10}
    & PGD & \textbf{87.19} & 56.44 & 51.80 &
    TRADES & 85.66 & 58.46 & 54.08 
    \\
    & +Pruning & 86.44 & 56.18 & 52.56 &
    +Pruning & 84.07 & 58.04 & 54.64 \\
    & +LS & 86.96 & 57.20 & 51.99 &
    +LS & \textbf{86.32} & 58.72 & 54.02 \\
    & +DPLS & 87.13 &  \textbf{58.43} & \textbf{54.03} &
    +DPLS & 85.36 &  \textbf{59.33} & \textbf{55.19}\\
\midrule[.1em]
\multirow{4}{*}{CIFAR-100}
    & PGD & 62.88 & 31.82 & 27.36 &
    TRADES & 60.78 & 33.19 & 28.05 
    \\
    & +Pruning & 62.28 & 31.64 & 28.05 &
    +Pruning & 60.61 & 33.53 & 29.01 \\
    & +LS & \textbf{63.02} & 32.26 & 27.67 &
    +LS & 62.43 & 33.29 & 27.83 \\
    & +DPLS & 62.95 &  \textbf{35.10} & \textbf{29.33} &
    +DPLS & \textbf{63.32} &  \textbf{34.53} & \textbf{29.68}\\
\midrule[.1em]
\multirow{4}{*}{SVHN}
    & PGD & \textbf{92.96} & 52.88 & 35.20 &
    TRADES & 91.84 & 59.14 & 47.46 
    \\
    & +Pruning & 92.68 & 55.52 & 40.87 &
    +Pruning & 91.54 & \textbf{60.27} & 47.79 \\
    & +LS & 92.65 & 66.20 & 38.44 &
    +LS & 92.27 & 59.65 & 46.37 \\
    & +DPLS & 92.94 &  \textbf{76.53} & \textbf{42.01} &
    +DPLS & \textbf{92.33} &  59.40 & \textbf{49.05}\\
\midrule[.1em]
\multirow{4}{*}{\jmlr{STL-10}}
    & \jmlr{PGD} & \jmlr{\textbf{66.95}} & \jmlr{36.49} & \jmlr{33.12} &
    \jmlr{TRADES} & \jmlr{66.40} & \jmlr{36.27} & 
    \jmlr{33.16}
    \\
    & \jmlr{+Pruning} & \jmlr{66.75} & 
    \jmlr{35.27} & \jmlr{33.15} &
    \jmlr{+Pruning} & \jmlr{65.39} & \jmlr{37.39} & \jmlr{34.91} \\
    & \jmlr{+LS} & \jmlr{66.90} & \jmlr{35.66} & \jmlr{32.05} &
    \jmlr{+LS} & \jmlr{\textbf{66.49}} & \jmlr{37.89} & \jmlr{33.99} \\
    & \jmlr{+DPLS} & \jmlr{66.16} &  \jmlr{\textbf{38.27}} & \jmlr{\textbf{34.75}} &
    \jmlr{+DPLS} & \jmlr{65.63} & \jmlr{\textbf{39.94}} & \jmlr{\textbf{36.06}} \\
\midrule[.1em]

\end{tabular}
\caption{$\epsilon=8$}
\label{tab:experiment_all110_main_2}
\end{subtable}

\begin{subtable}[t]{\textwidth}
\centering
\footnotesize %
\addtolength{\tabcolsep}{-1pt} %
\begin{tabular}{cl|ccc||l|ccc}
\toprule
\multicolumn{1}{c}{\multirow{2}{*}{Dataset}} &
\multicolumn{1}{c|}{\multirow{2}{*}{Method}} &
\multirow{2}{*}{Std.} & Robust &  \multicolumn{1}{c||}{Robust} &
\multicolumn{1}{c|}{\multirow{2}{*}{Method}} &
\multirow{2}{*}{Std.} & Robust &  \multicolumn{1}{c}{Robust}
\\
& \multicolumn{1}{c|}{} & & (PGD) & \multicolumn{1}{c||}{($A^3$)} &
\multicolumn{1}{c|}{} & & (PGD) & \multicolumn{1}{c}{($A^3$)}
\\
\midrule[.05em]
\multirow{4}{*}{CIFAR-10}
    & PGD & 91.78 & 72.60 & 71.96 &
    TRADES & 90.36 & 74.37 & 73.50 
    \\
    & +Pruning & 91.24 & 73.03 & 72.64 &
    +Pruning & 89.12 & 73.76 & 73.32 \\
    & +LS & \textbf{92.21} & 72.39 & 71.55 &
    +LS & \textbf{90.92} & 74.78 & 73.67 \\
    & +DPLS & 92.07 &  \textbf{74.90} & \textbf{74.01} &
    +DPLS & 90.43 &  \textbf{75.38} & \textbf{74.47}\\
\midrule[.1em]
\end{tabular}
\caption{$\epsilon=4$}
\label{tab:experiment_all110_main_2_eps4}
\end{subtable}

\hfill

\begin{subtable}[t]{\textwidth}
\centering
\footnotesize %
\addtolength{\tabcolsep}{-1pt} %
\begin{tabular}{cl|ccc||l|ccc}
\toprule
\multicolumn{1}{c}{\multirow{2}{*}{Dataset}} &
\multicolumn{1}{c|}{\multirow{2}{*}{Method}} &
\multirow{2}{*}{Std.} & Robust &  \multicolumn{1}{c||}{Robust} &
\multicolumn{1}{c|}{\multirow{2}{*}{Method}} &
\multirow{2}{*}{Std.} & Robust &  \multicolumn{1}{c}{Robust}
\\
& \multicolumn{1}{c|}{} & & (PGD) & \multicolumn{1}{c||}{($A^3$)} &
\multicolumn{1}{c|}{} & & (PGD) & \multicolumn{1}{c}{($A^3$)}
\\
\midrule[.05em]
\multirow{4}{*}{CIFAR-10}
    & PGD & 94.17 & 83.20 & 83.04 &
    TRADES & \textbf{93.32} & 84.42 & 84.29 
    \\
    & +Pruning & 92.95 & 82.73 & 82.70 &
    +Pruning & 92.08 & 83.84 & 83.69 \\
    & +LS & \textbf{94.56} & 83.46 & 83.13 &
    +LS & 93.14 & 84.33 & 83.96 \\
    & +DPLS & 94.06 &  \textbf{84.87} & \textbf{84.51} &
    +DPLS & 92.94 &  \textbf{85.18} & \textbf{84.79}\\
\midrule[.1em]
\end{tabular}
\caption{$\epsilon=2$}
\label{tab:experiment_all110_main_2_eps2}
\end{subtable}
\label{tab:experiment_all110_main_2}
\vskip -10mm
\end{table}

%% file: fig/fig_experiment_train_test_acc_from_pgd.tex
\begin{wrapfigure}[11]{R}{0.64\linewidth} %
\begin{center}
\vskip -2mm
\raisebox{0pt}[\dimexpr\height-1.5\baselineskip\relax]{\includegraphics[width=1.0\linewidth]{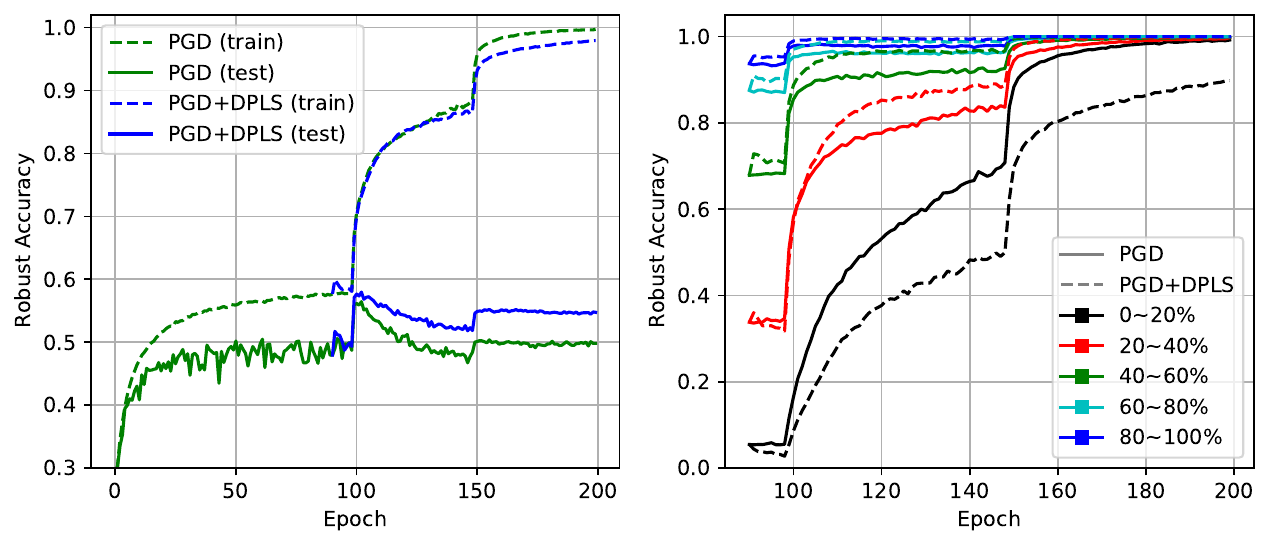}}%
\end{center}
\vskip -8mm
\caption{\hgg{The training and test accuracy curves of the PGD and the PGD with DPLS model on CIFAR-10.}
}
\label{fig:experiment_acc_from_pgd_all}
\end{wrapfigure}

%% file: table/tab_experiment_collection.tex
\begin{table}[t]
\begin{minipage}{.35\linewidth}
\caption{Performance comparison with other robust overfitting mitigation methods.}
\vspace{0pt}
\centering
\footnotesize
\addtolength{\tabcolsep}{-4pt}
\begin{tabular}{l|ccc}
    \toprule
     \multicolumn{1}{c|}{\multirow{1}{*}{Method}} & \multicolumn{1}{c}{Std.} & 
    \multicolumn{1}{c}{\begin{tabular}[c]{@{}c@{}}Robust\\ (PGD)\end{tabular}}& 
    \multicolumn{1}{c}{\begin{tabular}[c]{@{}c@{}}Robust\\ ($A^3$)\end{tabular}} \\
    
    \midrule[.05em]
     PGD & 87.19 & 56.44 & 51.8 \\
    +SAT & 86.94 & 57.52 & 52.94 \\
    +KD+SWA & \textbf{88.28} & 57.93 & 53.49 \\
    +TE & 85.83 & 58.31 & 52.84 \\

    \midrule[.05em]
    +DPLS & 86.17 & 58.14 & 53.98 \\
    +DPLS+SWA & 87.14 & \textbf{60.07} & \textbf{55.91} \\
    
    \midrule[.1em]
\end{tabular}
\label{tab:experiment_overfitting_methods}
\end{minipage}%
\hfill
\begin{minipage}{.3\linewidth}
\centering
\footnotesize
\addtolength{\tabcolsep}{-2pt}
\caption{Robust accuracy of DPLS according to various difficulty measurement.}
\vspace{0pt}
\begin{tabular}{c|cc}
    \toprule
     \multicolumn{1}{c|}{\multirow{1}{*}{Method}} & PGD & TRADES \\
    \midrule[.05em]
    - & 51.8 & 54.08 \\
    C-score & 52.9 & 54.71 \\
    CE loss & 52.73 & 54.45 \\
    EL2N & 53.38 & 54.32 \\
    0-1 loss & \textbf{54.03} & \textbf{55.19} \\
    
    \midrule[.1em]
\end{tabular}
\label{tab:experiment_difficulty_loss}
\end{minipage} 
\hfill
\begin{minipage}{.3\linewidth}
\centering
\footnotesize
\addtolength{\tabcolsep}{-4pt}
\caption{Performance improvements in other adversarial training algorithms.}
\vspace{0pt}
\begin{tabular}{l|ccc}
    \toprule
     \multicolumn{1}{c|}{\multirow{1}{*}{Method}} & \multicolumn{1}{c}{Std.} & 
    \multicolumn{1}{c}{\begin{tabular}[c]{@{}c@{}}Robust\\ (PGD)\end{tabular}}& 
    \multicolumn{1}{c}{\begin{tabular}[c]{@{}c@{}}Robust\\ ($A^3$)\end{tabular}} \\
    
    \midrule[.05em]
    AWP & \textbf{84.82} & 60.19 & 55.54 \\
    +DPLS & 84.4 & \textbf{60.26} & \textbf{56.09} \\
    \cmidrule{1-4}
    
    MART & \textbf{83.9} & 59.05 & 52.19 \\
    +DPLS & 83.81 & \textbf{59.54} & \textbf{53.18} \\
    \cmidrule{1-4}
    
    RST & \textbf{84.82} & 61.53 & 57.26 \\
    +DPLS & 84.43 & \textbf{62.03} & \textbf{57.81} \\
    
    \midrule[.1em]
\end{tabular}
\label{tab:experiment_training_algorithms}
\end{minipage}
\vskip -5mm
\end{table}

%% file: 6_conclusion.tex
\section{Conclusion}
\label{conclusion}

In this study, we analyzed the effect of hard examples in adversarial training. We demonstrated that training hard examples can deteriorate the performance in adversarial training and verified that the cause is the memorization of hard examples.
\jmlr{
We accordingly conducted an analysis of the memorization phenomenon and observed that increasing the number of hard examples during adversarial training leads to a decrease in the model's robust accuracy. While we found that pruning hard examples from the training process enhances the model's robustness, the difficulty lies in identifying the optimal threshold for removing hard examples that weaken robustness performance.
}
Thus, we proposed a method, DPLS, which \jmlr{adaptively} mitigates the memorization of hard examples.
Through experiments on various datasets and algorithms, we verified that our method could successfully leverage hard examples, thereby improving robustness.

%% file: B_proofs.tex
\section{Proofs}
\label{appendix:proof}

\subsection{Section \ref{method:effect of hard examples} proofs}
\label{appendix:proof3.1}

\begin{app_definition}[Gradient difference]
The differences between the loss gradient norm on $\bm{x}_1$ and $\bm{x}_2$ in standard and adversarial training are defined as follows:
\begin{equation}
\begin{gathered}
    \mathcal{D}_{\textup{std}}(\bm{\mathnormal{x}}_1, \bm{\mathnormal{x}}_2)=\norm{(1-\sigma(\bm{w}^\top \bm{\mathnormal{x}}_1 + \jmlr{b}))\bm{\mathnormal{x}}_1} -
    \norm{(1-\sigma(\bm{w}^\top \bm{\mathnormal{x}}_2+ \jmlr{b}))\bm{\mathnormal{x}}_2}\\
    \mathcal{D}_{\textup{adv}}(\bm{\mathnormal{x}}_1, \bm{\mathnormal{x}}_2, \epsilon)=\norm{(1-\sigma(\bm{w}^\top (\bm{\mathnormal{x}}_1+\bm{\delta})+ \jmlr{b}))(\bm{\mathnormal{x}}_1+\bm{\delta})} - 
    \norm{(1-\sigma(\bm{w}^\top (\bm{\mathnormal{x}}_2+\bm{\delta})+ \jmlr{b}))(\bm{\mathnormal{x}}_2+\bm{\delta})}.
\end{gathered}
\end{equation}
\end{app_definition}
\begin{proof}
\begin{align*}
\frac{\partial \mathcal{L}(f(\bm{x}), y)}{\partial \bm{w}} 
&=-\frac{\partial}{\partial \bm{w}}\big(y\log f(\bm{x}) + (1-y)\log(1-f(\bm{x}))\big) \\
&= -\frac{\partial}{\partial \bm{w}}\big(y\log \sigma(\bm{w}^\top \bm{\mathnormal{x}}+ \jmlr{b}) + (1-y)\log(1-\sigma(\bm{w}^\top \bm{\mathnormal{x}}+ \jmlr{b}))\big) \\
&= -\big(y(1-\sigma(\bm{w}^\top \bm{\mathnormal{x}}+ \jmlr{b})) + (1-y)(-\sigma(\bm{w}^\top \bm{\mathnormal{x}}+ \jmlr{b}))\big)\bm{x}.
\end{align*}
Here, $y=+1$ for $\bm{x}_1$ and $\bm{x}_2$
\begin{equation}
\frac{\partial \mathcal{L}(f(\bm{x}), y=+1)}{\partial \bm{w}} = -(1-\sigma(\bm{w}^\top \bm{\mathnormal{x}}+ \jmlr{b}))\bm{x}.
\end{equation}
The gradient norm difference between $\bm{x}_1$ and $\bm{x}_2$ in standard training is represented as:
\begin{equation}
    \norm{\frac{\partial \mathcal{L}(f(\bm{x}_1), y_1)}{\partial \bm{w}}} - \norm{\frac{\partial \mathcal{L}(f(\bm{x}_2), y_2)}{\partial \bm{w}}}
= \norm{(1-\sigma(\bm{w}^\top \bm{\mathnormal{x}}_1+ \jmlr{b}))\bm{x}_1} - \norm{(1-\sigma(\bm{w}^\top \bm{\mathnormal{x}}_2+ \jmlr{b}))\bm{x}_2}.
\end{equation}
Thus, the gradient norm difference between the adversarial examples is represented as:
\begin{multline}
    \norm{\frac{\partial \mathcal{L}(f(\bm{x}_1+\bm{\delta}), y_1)}{\partial \bm{w}}} - \norm{\frac{\partial \mathcal{L}(f(\bm{x}_2+\bm{\delta}), y_2)}{\partial \bm{w}}}\\
= \norm{(1-\sigma(\bm{w}^\top (\bm{\mathnormal{x}}_1+\bm{\delta)}+ \jmlr{b}))(\bm{x}_1+\bm{\delta})} - \norm{(1-\sigma(\bm{w}^\top (\bm{\mathnormal{x}}_2+\bm{\delta})+ \jmlr{b}))(\bm{x}_2+\bm{\delta})}.
\end{multline}
\end{proof}

\begin{app_theorem}
Let the gradient norm of easy examples is smaller than the gradient norm of hard examples in both standard and adversarial training. Then, if $\bm{w}^\top \bm{\mathnormal{x}}_1 + \jmlr{b} \geq 0$ and  $0<\epsilon < \frac{(d_1 +d_2)}{2}$, it satisfies
\begin{equation}
\mathcal{D}_{\textup{std}}(\bm{\mathnormal{x}}_1, \bm{\mathnormal{x}}_2) <  \mathcal{D}_{\textup{adv}}(\bm{\mathnormal{x}}_1, \bm{\mathnormal{x}}_2, \epsilon).    
\end{equation}

\end{app_theorem}
\label{eq:method_corollary_overfitting}

\begin{proof}

Because the gradient norm of easy examples is smaller than the gradient norm of hard examples in both standard and adversarial training, $\mathcal{D}_{\textup{std}}(\bm{\mathnormal{x}}_1, \bm{\mathnormal{x}}_2) > 0$ and $\mathcal{D}_{\textup{adv}}(\bm{\mathnormal{x}}_1, \bm{\mathnormal{x}}_2, \epsilon) > 0$.
Then, we can rewrite Equation~\ref{eq:grad_difference_ineq} as
\begin{equation}
\label{eq:grad_difference_ineq}
\frac{\mathcal{D}_{\textup{std}}(\bm{\mathnormal{x}}_1, \bm{\mathnormal{x}}_2)}{\mathcal{D}_{\textup{adv}}(\bm{\mathnormal{x}}_1, \bm{\mathnormal{x}}_2, \epsilon)} < 1.
\end{equation}
Let the prediction for $x_1$ is not incorrect, where $\bm{w}^\top \bm{x}_1 + \jmlr{b} \geq 0 $. Here, we assume the norm of every input $x$ that is defined in the input space $\mathcal{X}$ is almost the same. Then, the above inequality is represented as
\begin{align}
\frac{\mathcal{D}_{\textup{std}}(\bm{\mathnormal{x}}_1, \bm{\mathnormal{x}}_2)}{\mathcal{D}_{\textup{adv}}(\bm{\mathnormal{x}}_1, \bm{\mathnormal{x}}_2, \epsilon)} 
&= \frac{\norm{(1-\sigma(\bm{w}^\top \bm{\mathnormal{x}}_1+ \jmlr{b}))\bm{\mathnormal{x}}_1} -
    \norm{(1-\sigma(\bm{w}^\top \bm{\mathnormal{x}}_2+ \jmlr{b}))\bm{\mathnormal{x}}_2}}{\norm{(1-\sigma(\bm{w}^\top (\bm{\mathnormal{x}}_1+\bm{\delta})+ \jmlr{b}))(\bm{\mathnormal{x}}_1+\bm{\delta})} - 
    \norm{(1-\sigma(\bm{w}^\top (\bm{\mathnormal{x}}_2+\bm{\delta})+ \jmlr{b}))(\bm{\mathnormal{x}}_2+\bm{\delta})}} \\
&= \frac{{(1-\sigma(\bm{w}^\top \bm{\mathnormal{x}}_1+ \jmlr{b}))}\norm{\bm{\mathnormal{x}}_1} -
    {(1-\sigma(\bm{w}^\top \bm{\mathnormal{x}}_2+ \jmlr{b}))}\norm{\bm{\mathnormal{x}}_2}}{{(1-\sigma(\bm{w}^\top (\bm{\mathnormal{x}}_1+\bm{\delta})+ \jmlr{b}))}\norm{(\bm{\mathnormal{x}}_1+\bm{\delta})} - 
    {(1-\sigma(\bm{w}^\top (\bm{\mathnormal{x}}_2+\bm{\delta})+ \jmlr{b}))}\norm{(\bm{\mathnormal{x}}_2+\bm{\delta})}} \\
&= \frac{{(1-\sigma(\bm{w}^\top \bm{\mathnormal{x}}_1+ \jmlr{b}))} - {(1-\sigma(\bm{w}^\top \bm{\mathnormal{x}}_2+ \jmlr{b}))}}{{(1-\sigma(\bm{w}^\top (\bm{\mathnormal{x}}_1+\bm{\delta})+ \jmlr{b}))} - 
    {(1-\sigma(\bm{w}^\top (\bm{\mathnormal{x}}_2+\bm{\delta})+ \jmlr{b}))}} \\
&=\frac{{\sigma(-\bm{w}^\top \bm{\mathnormal{x}}_1- \jmlr{b})} - {\sigma(-\bm{w}^\top \bm{\mathnormal{x}}_2- \jmlr{b})}}{{\sigma(-\bm{w}^\top (\bm{\mathnormal{x}}_1+\bm{\delta})- \jmlr{b})} - 
    {\sigma(-\bm{w}^\top (\bm{\mathnormal{x}}_2+\bm{\delta})- \jmlr{b})}} \\
&=\frac{{\sigma(-d_1)} - {\sigma(-d_2)}}{{\sigma(-(d_1-\epsilon))} - 
    {\sigma(-(d_2-\epsilon))}} \\
&=\frac{{\sigma(-d_1)} - {\sigma(-(d_1 + (d_2 - d_1)))}}{{\sigma(-(d_1-\epsilon))} - 
    {\sigma(-(d_1-\epsilon + (d_2-d_1)))}}.
\end{align}
Here, it satisfies Equation~\ref{eq:grad_difference_ineq} if the function $h(x)={\sigma(-x)} - {\sigma(-(x + (d_2 - d_1)))}$ is monotonically decreasing for input $x \geq d_1-\epsilon$, where $\epsilon > 0$. Let $a$ denote $(d_2-d_1)$, where $a=(d_2-d_1)$. Then,
\begin{align}
\frac{d}{dx}\big(\sigma(-x) - \sigma(-(x + a))\big) \label{eq:frac_ineq}
&= \frac{d}{dx}\bigg(\frac{1}{1+e^{x}} - \frac{1}{1+e^{x + a}}\bigg) \\
&= e^{x}\bigg(\frac{e^a}{(1+e^{x+a})^2} - \frac{1}{(1+e^{x})^2} \bigg) \\
&= e^{x}\bigg(\frac{(e^a-1)(1 - e^{2x+a})}{(e^x+1)^2(e^{x+a}+1)^2} \bigg) < 0.
\label{eq:decreasing_ineq}
\end{align}
Since $a > 0$, Equation~\ref{eq:decreasing_ineq} can be represented as $1 - e^{2x+a} < 0$. Then, we can rewrite Equation~\ref{eq:decreasing_ineq} as
\begin{equation}
x > -\frac{d_2 - d_1}{2}.
\end{equation}
Thus, the fuctnion $h(x)={\sigma(-x)} - {\sigma(-(x + (d_2 - d_1)))}$ is monotonically decreasing for $x > -\frac{d_2 - d_1}{2}$, and here, the minimum value of $x$ is $d_1-\epsilon$. Therefore, if $d_1-\epsilon > -\frac{d_2 - d_1}{2}$, which can be rewritten as $\epsilon < \frac{d_1+d_2}{2}$, it satisfies
\begin{equation}
\mathcal{D}_{\textup{std}}(\bm{\mathnormal{x}}_1, \bm{\mathnormal{x}}_2) < \mathcal{D}_{\textup{adv}}(\bm{\mathnormal{x}}_1, \bm{\mathnormal{x}}_2, \epsilon).
\end{equation}

\end{proof}

\begin{app_corollary}
The increase in gradient difference by an adversarial attack is proportional to the difficulty of the example and the size of the upper bound of adversarial perturbations. Thus, if $\norm{\bm{w}^\top \bm{\mathnormal{x}}_3}=d_3$ and $d_2 < d_3$, it satisfies
\begin{equation}
\mathcal{D}_{\textup{adv}}(\bm{\mathnormal{x}}_1, \bm{\mathnormal{x}}_2, \epsilon) <  \mathcal{D}_{\textup{adv}}(\bm{\mathnormal{x}}_1, \bm{\mathnormal{x}}_3, \epsilon),
\label{eq:method_corrllary_1}
\end{equation}
and if $\epsilon_1 < \epsilon_2$, it satisfies
\begin{equation}
\mathcal{D}_{\textup{adv}}(\bm{\mathnormal{x}}_1, \bm{\mathnormal{x}}_2, \epsilon_1) <  \mathcal{D}_{\textup{adv}}(\bm{\mathnormal{x}}_1, \bm{\mathnormal{x}}_2, \epsilon_2).    
\label{eq:method_corrllary_2}
\end{equation}
\end{app_corollary}
\label{eq:method_corollary}

\begin{proof}
In the equation~\ref{eq:frac_ineq} of the above proof, it satisfies Equation~\ref{eq:method_corrllary_1} if the function $h(x)={\sigma(-(d_1-\epsilon))} - {\sigma(-(d_1-\epsilon + (x - d_1)))}$ is monotonically increasing for input $x > d_1$. Then, 
\begin{align}
\frac{d}{dx}\big(\sigma(-(d_1-\epsilon)) - \sigma(-(d_1-\epsilon + (x-d_1)))\big)
&= \frac{d}{dx}\bigg(-\frac{1}{1+e^{x - d_1}}\bigg) \\
&= \bigg(\frac{e^{x-d_1}}{(1+e^{x-d_1})^2} \bigg) > 0.
\label{eq:increasing_ineq_1}
\end{align}
In the same way, it satisfies Equation~\ref{eq:method_corrllary_2} if the function $h(x)={\sigma(-(d_1-\epsilon))} - {\sigma(-(d_1-\epsilon + (d_2 - d_1)))}$ is monotonically increasing for input $\epsilon \geq 0$. Let $x$ denote $\epsilon - d_1$ and $a$ denote $d_2 - d_1$.
Then, if $\epsilon-d_1 < \frac{d_2 - d_1}{2}$, which can be rewritten as $\epsilon < \frac{d_1+d_2}{2}$, it satisfies
\begin{align}
\frac{d}{dx}\big(\sigma(x) - \sigma(x - a))\big)
&= \frac{d}{dx}\bigg(\frac{1}{1+e^{-x}} -\frac{1}{1+e^{-x+a}}\bigg) \\
&= e^{x-a}\bigg(\frac{(e^a-1)(1-e^{2x-a})}{(e^x+1)^2(e^{x-a}+1)^2} \bigg) >0.
\label{eq:increasing_ineq_2}
\end{align}

\end{proof}

\subsection{\jmlr{Section \ref{method:theoretical training of hard examples} proofs}}
\label{appendix:proof3.2}
\jmlr{
Next, we proceed with the proofs of Theorems \ref{eq:method_theorem_acc_comparison} and \ref{eq:method_theorem_acc_comparison_hard}. Before the proof, we will examine the derivation process of the adversarial training model weight $w_{adv}$ in Equation \ref{eq:method_std_adv_weight_hard}, which represents the weights of models trained with $m$ normal examples and $n$ hard examples. Following \citet{adv_odds_with_accuracy}, we utilize a soft-margin SVM classifier. The optimization problem for the model is as follows.
}
\jmlr{
\begin{equation}
\begin{gathered}
    \label{eq:tmp_eq}
    \min_{\bm{w}} \mathbb{E} \big[
    \max (0, 1-y\bm{w}^{\top}\bm{x})
    \big] \\
    \textnormal{subject to } \smalltab \norm{\bm{w}} = 1
\end{gathered}
\end{equation}
}
\jmlr{
Given that $\norm{\bm{x}} \leq 1$ and $\beta \eta < \epsilon < \eta$, the weight learning process for the adversarial training model is as follows.
}

\jmlr{
\begin{align}
    \mathbb{E} \big[
    \max &(0, 1-y\bm{w}^{\top}(\bm{x}+\bm{\delta}))
    \big] = \mathbb{E} \big[
    1-y\bm{w}^{\top}(\bm{x}+\bm{\delta})
    \big]
    \\
    &= \mathbb{E} \big[
    1- w_0 \mathcal{N}(\eta-\epsilon,1) - \sum_{i=1}^d w_i\mathcal{N}(\textnormal{sign}(w_i) \cdot (\beta \eta-\epsilon),1) - w_{d+1} \mathcal{N}(\eta-\epsilon,1)
    \big] \\     
    &=  1 - \frac{m}{m+n}w_0(\eta-\epsilon) - \sum_{i=1}^d w_i\textnormal{sign}(w_i) \cdot (\beta \eta - \epsilon) - \frac{n}{m+n}w_{d+1}(\eta - \epsilon)
\end{align}
}

\jmlr{
According to the above equation, in adversarial training, the optimal solution is to assign a value of 0 to the weights corresponding to the non-robust features, which is consistent with what was demonstrated by \citet{adv_odds_with_accuracy}. Therefore, when $w_{i \in 1,..., d}=0$, Equation \ref{eq:tmp_eq} can be simplified as follows.
}

\jmlr{
 \begin{equation}
 \begin{gathered}     
    \min_{\bm{w}}  - \frac{m}{m+n}w_0(\eta-\epsilon) - \frac{n}{m+n}w_{d+1}(\eta - \epsilon) \\
    \textnormal{subject to } \smalltab w_0^2 + w_{d+1}^2 = 1
\end{gathered}
 \end{equation}
 }

\jmlr{
Using the Lagrange multiplier method, we define the above equation, and the partial derivatives with respect to each term are as follows.
}

\jmlr{
 \begin{align}
    \mathcal{L}(w_0, w_{d+1}, \mu) &= - \frac{m}{m+n}w_0(\eta-\epsilon) - \frac{n}{m+n}w_{d+1}(\eta - \epsilon) + \mu(w_0^2 + w_{d+1}^2 - 1) \\
    \frac{\partial \mathcal{L}}{\partial w_0} &= - \frac{m}{m+n}(\eta-\epsilon) + 2\mu w_0 = 0  \\
    \frac{\partial \mathcal{L}}{\partial w_{d+1}} &= - \frac{n}{m+n}(\eta-\epsilon) + 2\mu w_{d+1} = 0 \\
    \frac{\partial \mathcal{L}}{\partial \mu} &= w_0^2 + w_{d+1}^2 - 1 = 0
 \end{align}
 }

\jmlr{
Given that $ -\frac{m}{m+n}(\eta - \epsilon) = a_0 $ and $ -\frac{n}{m+n}(\eta - \epsilon) = a_1 $, we can rewrite the equations as $ a_0 = -2\mu w_0 $ and $ a_1 = -2\mu w_{d+1} $. Dividing the two equations gives $\frac{a_0}{a_1} = \frac{w_0}{w_{d+1}}$, so $ w_{d+1} = \frac{a_1}{a_0}w_0 $. Substituting this into $ w_0^2 + w_{d+1}^2 - 1 = 0 $ gives $ w_0^2 = \frac{a_0^2}{a_0^2 + a_1^2} $ and $ w_{d+1}^2 = \frac{a_1^2}{a_0^2 + a_1^2} $. Therefore, to obtain the minimum values, $ w_0 = -\frac{a_0}{\sqrt{a_0^2 + a_1^2}} $ and $ w_{d+1} = -\frac{a_1}{\sqrt{a_0^2 + a_1^2}} $. Dividing $ a_0 $ and $ a_1 $ by $ -(\eta - \epsilon) $ gives $ w_0 = \frac{1}{Z_{adv}} \frac{m}{m+n} $ and $ w_{d+1} = \frac{1}{Z_{adv}} \frac{n}{m+n} $.
}

\jmlr{
Next, we proceed with the proof of Theorem \ref{eq:app_theorem_acc_comparison}.
}

\jmlr{
\begin{app_theorem}
\label{eq:app_theorem_acc_comparison}
For a model $f_{adv}$ adversarially trained on the combined dataset $\mathcal{X}_{normal}^{train} \cup \mathcal{X}_{hard}^{train}$, where $\mathcal{X}_{normal}^{train}$ consists of $m$ normal examples and $\mathcal{X}_{hard}^{train}$ consists of $n$ hard examples, both the standard accuracy $\text{Pr}\big[{f}_{adv}(\bm{x})=y\big]$ and the robust accuracy $\text{Pr}\big[{f}_{adv}(\bm{x}+\bm{\delta})=y\big]$ on the normal example test dataset $\mathcal{X}_{normal}^{test}$ are monotonically decreasing with respect to $n$.
\end{app_theorem}
}

\begin{proof}
\jmlr{
First, the standard accuracy $\text{Pr}\big[{f}_{adv}(\bm{x})=y\big]$ of ${f}_{adv}$ on $\mathcal{X}_{normal}^{test}$ is expressed as follows. Here, the bias $b$ is assumed to be $b \neq 0$, and it is also assumed that the proportions of the two classes in the test data are equal.
}

\jmlr{
\begin{align}
\label{eq_app_std_adv_acc_hard}
    \text{Pr}\big[{f}_{adv}(\bm{x})=y\big] &= \text{Pr}\big[y(\bm{w}^{\top}\bm{x})+b>0\big]\\
    & =\text{Pr}\bigg[\frac{1}{{Z}_{adv}}\bigg(\frac{m}{m+n} \cdot \mathcal{N}(\eta , 1) \bigg) + y \cdot b > 0\bigg] \\
    & =\text{Pr}\bigg[\frac{m}{\sqrt{m^2+n^2}} \cdot \mathcal{N}(\eta , 1) + y \cdot b > 0 \bigg] \\
    & = \text{Pr}\bigg[
    \mathcal{N}(0 , 1) > -y \cdot \frac{b{\sqrt{m^2+n^2}}}{m} - \eta 
    \bigg] \\
    & = \frac{1}{2}\text{Pr}\bigg[
    \mathcal{N}(0 , 1) > - \frac{b{\sqrt{m^2+n^2}}}{m} - \eta 
    \bigg] +
    \frac{1}{2}\text{Pr}\bigg[
    \mathcal{N}(0 , 1) >  \frac{b{\sqrt{m^2+n^2}}}{m} - \eta 
    \bigg]
\end{align}
}

\jmlr{
Let the CDF $\Phi(\cdot)$ represent the cumulative distribution function, and the PDF $\phi(\cdot)$ represent the probability density function. Additionally, for readability, we substitute $g(n) = \frac{b\sqrt{m^2+n^2}}{m}$. Then,
}
\jmlr{
\begin{align}
\label{eq_app_std_adv_acc_hard}
    \text{Pr}\big[{f}_{adv}(\bm{x})=y\big]
    &\stackrel{\Phi(\cdot) \textnormal{ is CDF}}{=} \frac{1}{2} \bigg( 2 - \Phi (-g(n)-\eta) -  \Phi (g(n)-\eta) \bigg) \\
    & = \frac{1}{2} \bigg( 1 +  \Phi (g(n)+\eta) -  \Phi (g(n)-\eta) \bigg) \\
    & \underset{\textnormal{derivative w.r.t. } n}{\overset{\phi(\cdot) \textnormal{ is PDF}}{\rightarrow}}
    \frac{1}{2} \bigg(
    \phi(g(n)+\eta)\cdot g'(n) - \phi(g(n)-\eta)\cdot g'(n)
    \bigg) \\
    & = \frac{g'(n)}{2\sqrt{2 \pi}} \bigg( \exp (-\frac{(g(n)+\eta)^2}{2}) - 
    \exp (-\frac{(g(n)-\eta)^2}{2})    
    \bigg) \leq 0
\end{align}
}
\jmlr{
  When $b < 0$, $g'(n) \leq 0$ and $-\big(g(n) + \eta\big)^2 > -\big(g(n) - \eta\big)^2$, so the above result is negative. Similarly, when $b > 0$, $g'(n) \geq 0$ and $-\big(g(n) + \eta\big)^2 < -\big(g(n) - \eta\big)^2$, so the above result is also negative. Consequently, with respect to the size $n$ of the hard example training dataset, the standard accuracy $\text{Pr}\big[{f}_{adv}(\bm{x})=y\big]$ of ${f}_{adv}$ on the normal example test dataset $\mathcal{X}_{normal}^{test}$ is monotonically decreasing as $n$ increases.
}

\jmlr{
For robust accuracy $\text{Pr}\big[{f}_{adv}(\bm{x}+\bm{\delta})=y\big]$, since $\eta - \epsilon > 0$, replacing $\eta$ with $\eta - \epsilon$ in the above proof regarding standard accuracy does not change the outcome of the proof. Therefore, similar to standard accuracy, the robust accuracy $\text{Pr}\big[{f}_{adv}(\bm{x}+\bm{\delta})=y\big]$ is also monotonically decreasing with respect to $n$.
}
\end{proof}
\jmlr{
\begin{app_theorem}
\label{eq:app_theorem_acc_comparison_hard}
For a model $f_{adv}$ adversarially trained on the combined dataset $\mathcal{X}_{normal}^{train} \cup \mathcal{X}_{hard}^{train}$, where $\mathcal{X}_{normal}^{train}$ consists of $m$ normal examples and $\mathcal{X}_{hard}^{train}$ consists of $n$ hard examples, the standard accuracy $\text{Pr}\big[{f}_{adv}(\bm{x})=y\big]$ on the hard example test dataset $\mathcal{X}_{hard}^{test}$ is monotonically increasing with respect to $n$, but the robust accuracy $\text{Pr}\big[{f}_{adv}(\bm{x}+\bm{\delta})=y\big]$ is monotonically decreasing with respect to $n$.
\end{app_theorem}
}
\begin{proof}
    
\jmlr{
First, the standard accuracy $\text{Pr}\big[{f}_{adv}(\bm{x})=y\big]$ of ${f}_{adv}$ on $\mathcal{X}_{hard}^{test}$ is expressed as follows. As before, it is assumed that the bias $b$ is $b \neq 0$, and that the two classes in the test data are equally represented.
}

\jmlr{
\begin{align}
\label{eq_app_std_adv_acc_hard}
    \text{Pr}\big[{f}_{adv}(\bm{x})=y\big] &= \text{Pr}\big[y(\bm{w}^{\top}\bm{x})+b>0\big]\\
    & =\text{Pr}\bigg[\frac{1}{{Z}_{adv}}\bigg(\frac{n}{m+n} \cdot \mathcal{N}(\beta\eta , 1) \bigg) + y \cdot b > 0\bigg] \\
    & =\text{Pr}\bigg[\frac{n}{\sqrt{m^2+n^2}} \cdot \mathcal{N}(\beta\eta , 1) + y \cdot b > 0 \bigg] \\
    & = \text{Pr}\bigg[
    \mathcal{N}(0 , 1) > -y \cdot \frac{b{\sqrt{m^2+n^2}}}{n} - \beta\eta 
    \bigg] \\
    & = \frac{1}{2}\text{Pr}\bigg[
    \mathcal{N}(0 , 1) > - \frac{b{\sqrt{m^2+n^2}}}{n} - \beta\eta 
    \bigg] +
    \frac{1}{2}\text{Pr}\bigg[
    \mathcal{N}(0 , 1) >  \frac{b{\sqrt{m^2+n^2}}}{n} - \beta\eta 
    \bigg]
\end{align}
}
\jmlr{
As with the above, let the CDF $\Phi(\cdot)$ represent the cumulative distribution function, and the PDF $\phi(\cdot)$ represent the probability density function. Additionally, we substitute $g(n) = \frac{b\sqrt{m^2+n^2}}{n}$. Then,
}
\jmlr{
\begin{align}
\label{eq_app_std_adv_acc_hard}
    \text{Pr}\big[{f}_{adv}(\bm{x})=y\big]
    &\stackrel{\Phi(\cdot) \textnormal{ is CDF}}{=} \frac{1}{2} \bigg( 2 - \Phi (-g(n)-\beta\eta) -  \Phi (g(n)-\beta\eta) \bigg) \\
    & = \frac{1}{2} \bigg( 1 +  \Phi (g(n)+\beta\eta) -  \Phi (g(n)-\beta\eta) \bigg) \\
    & \underset{\textnormal{derivative w.r.t. } n}{\overset{\phi(\cdot) \textnormal{ is PDF}}{\rightarrow}}
    \frac{1}{2} \bigg(
    \phi(g(n)+\beta\eta)\cdot g'(n) - \phi(g(n)-\beta\eta)\cdot g'(n)
    \bigg) \\
    & = \frac{g'(n)}{2\sqrt{2 \pi}} \bigg( \exp (-\frac{(g(n)+\beta\eta)^2}{2}) - 
    \exp (-\frac{(g(n)-\beta\eta)^2}{2})    
    \bigg) \geq 0 \\
\end{align}
}
\jmlr{
The CDF $\Phi(\cdot)$ represents the cumulative distribution function, and the PDF $\phi(\cdot)$ represents the probability density function. For readability, we substitute $g(n) = \frac{b\sqrt{m^2+n^2}}{n}$. Since the sign of $g'(n)$ in this result is opposite to that in the proof of Theorem \ref{eq:app_theorem_acc_comparison}, the outcome is also opposite. When $b < 0$, $g'(n) \geq 0$ and $-\big(g(n) + \eta\big)^2 > -\big(g(n) - \eta\big)^2$, so the result is positive. Similarly, when $b > 0$, $g'(n) \leq 0$ and $-\big(g(n) + \eta\big)^2 < -\big(g(n) - \eta\big)^2$, so the result is also positive. Consequently, with respect to the size $n$ of the hard example training dataset, the standard accuracy $\text{Pr}\big[{f}_{adv}(\bm{x})=y\big]$ of ${f}_{adv}$ on the hard example test dataset $\mathcal{X}_{hard}^{test}$ is monotonically increasing with $n$.
}

\jmlr{
For robust accuracy $\text{Pr}\big[{f}_{adv}(\bm{x}+\bm{\delta})=y\big]$, since $\eta - \epsilon < 0$, replacing $\eta$ with $\eta - \epsilon$ in the proof for standard accuracy results in $-\big(g(n) + \eta\big)^2$ and $\big(g(n) - \eta\big)^2$ being replaced by $-\big(g(n) + \eta - \epsilon\big)^2$ and $\big(g(n) - \eta + \epsilon\big)^2$, respectively, which reverses the order of the two terms. Consequently, the result is the opposite of that for standard accuracy, and the robust accuracy $\text{Pr}\big[{f}_{adv}(\bm{x}+\bm{\delta})=y\big]$ is monotonically decreasing with $n$.
}

\end{proof}

\subsection{\jmlr{Section \ref{method:empirical verification of hard examples} proofs}}
\label{appendix:proof3.3}

\jmlr{
Next, by comparing the relative increases and decreases in standard accuracy in Theorem \ref{eq:app_theorem_acc_comparison} and Theorem \ref{eq:app_theorem_acc_comparison_hard}, we can derive the following corollary.
}

\jmlr{
\begin{app_corollary}
\label{eq:app_corollary_acc_comparison_hard}
For a model $f_{adv}$ adversarially trained on the combined dataset $\mathcal{X}_{normal}^{train} \cup \mathcal{X}_{hard}^{train}$, where $\mathcal{X}_{normal}^{train}$ consists of $m$ normal examples and $\mathcal{X}_{hard}^{train}$ consists of $n$ hard examples, the standard accuracy $\text{Pr}\big[{f}_{adv}(\bm{x})=y\big]$ on the overall test dataset $\mathcal{X}_{normal}^{test} \cup \mathcal{X}_{hard}^{est}$ is monotonically increasing with respect to $n$, if $n^3 \leq \beta m^3$.
\end{app_corollary}
}

\begin{proof}
    
\jmlr{
To compare the expressions for the increase and decrease in standard accuracy derived from the proofs of the two theorems above, we approximate each value using a first-order Taylor expansion. The results are as follows.
}

\jmlr{
\begin{equation}
\begin{gathered}
\frac{g'(n)}{2\sqrt{2 \pi}} \bigg( \exp (-\frac{(g(n)+\eta)^2}{2}) - 
    \exp (-\frac{(g(n)-\eta)^2}{2})    
    \bigg) = \frac{g'(n)}{2\sqrt{2 \pi}} (-2g(n)\eta), \smalltab \smalltab \smalltab \textnormal{(normal)} \\
\frac{g'(n)}{2\sqrt{2 \pi}} \bigg( \exp (-\frac{(g(n)+\beta\eta)^2}{2}) - 
    \exp (-\frac{(g(n)-\beta\eta)^2}{2})    
    \bigg) = \frac{g'(n)}{2\sqrt{2 \pi}} (-2g(n)\beta\eta), \smalltab \textnormal{(hard)}
\end{gathered}
\end{equation}
}

\jmlr{
Here, we examine the sign of the sum of the two terms above. It is assumed that the ratio of normal examples to hard examples in the test phase is $m:n$. Let $g(n)$ for hard examples be denoted as $g_{hard}(n)$, which can be expressed as follows.
}

\jmlr{
\begin{align}
m \cdot \frac{g'(n)}{2\sqrt{2 \pi}} (-2g(n)\eta) + n \cdot \frac{g_{hard}'(n)}{2\sqrt{2 \pi}} (-2g_{hard}(n)\beta\eta) &= 
\frac{1}{2\sqrt{2 \pi}}\bigg(-\frac{2nb^2\eta}{m} + \frac{2m^2b^2\beta\eta}{n^2} \bigg) \\
&= \frac{b^2\eta}{\sqrt{2 \pi}mn^2}\big(-n^3 + \beta m^3 \big)
\end{align}
}

\jmlr{
Therefore, if $n^3 \leq \beta m^3$, increasing the number of hard examples $n$ in the training dataset will lead to an increase in standard accuracy, whereas in the opposite case, it will decrease. Since datasets collected under normal circumstances usually have a higher proportion of normal examples compared to hard examples, it can generally be assumed that the condition $n^3 \leq \beta m^3$ is satisfied. Consequently, increasing the number of hard examples $n$ in the training dataset will typically result in an increase in standard accuracy.
}

\end{proof}

\subsection{\jmlr{Section \ref{method:our method to mitigate the negative effect of hard examples} proofs}}
\label{appendix:proof4.2}

\jmlr{
We proceed with the proof of the following theory discussed in Section \ref{method:mitigation of the effect of hard examples}.
}

\jmlr{
\begin{app_theorem}
\label{eq:app_theorem_ls_factor}
Let the robust feature coefficients of each data point in the dataset $\mathcal{X}_{difficulty}^{train}$, consisting of $k$ data points $\bm{x}_1, \bm{x}_2, \dots, \bm{x}_k$, be denoted as $\alpha_1, \alpha_2, \dots, \alpha_k$. Then, when performing adversarial training using $\mathcal{X}_{difficulty}^{train}$, the optimal LS factors $\lambda_1, \lambda_2, \dots, \lambda_k$ for each training example are proportional to their respective robust feature coefficients $\alpha_1, \alpha_2, \dots, \alpha_k$.
\end{app_theorem}
}

\begin{proof}

\jmlr{
First, the weight of the adversarially trained model using the data given in Theorem \ref{eq:app_theorem_ls_factor} is derived through the same process as the weight $\bm{w}_adv$ of the above $f_{adv}$, and is given as follows.
}

\jmlr{  
\begin{equation}
    \bm{w} = \frac{1}{Z} \begin{cases}
        w_0 = \sum_{i=1}^k -(2\lambda_i-1)(\alpha_i\eta - \epsilon) \\
        w_{i \in {1,...,k}} = 0 \\
        w_{d+1} =\sum_{i=1}^k -(2\lambda_i-1)(\gamma_i\eta - \epsilon)
    \end{cases}
\end{equation}
}
\jmlr{
Here, $Z$ is a normalization term tuned to satisfy $\norm{\bm{w}} = 1$. To ensure that the weight parameter $\bm{w}$ is close to $\bm{w}^{*} = (1,0,...,0)$, the LS factor is assigned in such a way that $w_0$ increases and $w_{d+1}$ decreases. Therefore, the optimization equation to be solved can be expressed as follows.
}

\jmlr{
 \begin{equation}
 \begin{gathered}     
    \min_{\bm{\lambda}} w_{d+1}-w_0 \\
    \textnormal{subject to } \smalltab 0.5 \leq \lambda_{i} \leq 1, \smalltab i=1,...,k     
\end{gathered}
 \end{equation}
 }

 \jmlr{
Here, $\bm{\lambda} = (\lambda_0, ..., \lambda_k)$. The optimization objective can be expressed in the form of a vector operation with respect to $\bm{\lambda}$ as follows.
 }

 \jmlr{
\begin{align}
    w_{d+1}-w_0 = 2 \eta \bm{\lambda}^{\top}(\bm{\bm{\alpha}-\gamma}) - \eta(\bm{\alpha}-\bm{\gamma})^{\top}\bm{1}
\end{align}
}

\jmlr{
Therefore, the optimization objective is minimized when $\lambda$ is proportional to $(\alpha-\gamma)$. Here, let $c_1, c_2 > 0$ be constants. Since $\gamma$ is inversely proportional to $\alpha$, it can be expressed as $\gamma = -c_1\alpha$. Consequently, $\lambda$ satisfies $\lambda = \min(0.5, \max(1, c_2(1+c_1)\alpha))$. Therefore, the optimal LS factor can ultimately be said to be proportional to the robust feature coefficient.
}

\end{proof}

%% file: C_implementations.tex
\section{Additional implementation and experimental details}
\label{appendix:experiment_detail}

\subsection{Implementation details}
\label{appendix:implementation_details}

\paragraph{Datasets}

CIFAR-10~\citep{dataset_cifar} consists of 50,000 training images and 10,000 test images with 10 classes.
CIFAR-100~\citep{dataset_cifar} consists of 50,000 training images and 10,000 test images with 100 classes.
CIFAR-10 and CIFAR-100 images have sizes of $32\times32$ pixels.
The CIFAR datasets are subsets of the 80 million tiny images dataset~\citep{dataset_80mti}, and the 80 million tiny images dataset contains images downloaded from seven independent image search engines: Altavista, Ask, Flickr, Cydral, Google, Picsearch, and Webshots.
SVHN~\citep{dataset_svhn} consists of 73,257 training images and 26,032 test images with 10 classes.
SVHN images have sizes of $32\times32$ pixels. SVHN is obtained from a very large set of images from urban areas in various countries by using Google Street View.
\jmlr{
STL-10~\citep{dataset_stl10} consists of 5,000 training images and 8,000 test images with 10 classes. Additionally, it provides 100,000 unlabeled images to support unsupervised learning. STL-10 images have sizes of $96\times96$ pixels, which is larger than the typical $32\times32$ images found in datasets like CIFAR-10~\citep{dataset_cifar}. In our experiments, we resized the STL-10 dataset images to $64\times64$ pixels for model training.
}
Further details can be found in \url{https://paperswithcode.com/datasets}.

\paragraph{Implementation details}

We conducted experiments on CIFAR-10, CIFAR-100~\citep{dataset_cifar}, SVHN~\citep{dataset_svhn}, \jmlr{and STL-10} \citep{dataset_stl10}. We used PGD~\citep{attack_pgd} and TRADES~\citep{defense_trades} as the baseline adversarial training algorithms. We used WideResNet28-10~\citep{arch_wideresnet} as the architecture of our models. The learning rates for CIFAR-10, CIFAR-100, and \jmlr{STL-10} are set to 0.1 and 0.01 for SVHN, and the decay at 100 and 105 of the total training epoch 110 with decay factor 0.1 following~\citet{adv_bag_of_tricks}. We additionally conducted experiments with the learning rate decay at 50\% and 75\% of the total training epoch 200 with decay factor 0.1 following~\citet{attack_pgd}. We used stochastic gradient descent optimizer with the weight decay factor 5e-4 and the momentum 0.9. The upper bounds of adversarial perturbation were set to 0.031 ($\epsilon=8$), 0.0155 ($\epsilon=4$), and 0.00775 ($\epsilon=2$), and the step-size of training adversarial examples of each model were set to one fourth of the $\ell_\infty\text-$bound of each model with 10 steps.
To evaluate robustness of the models, we used the 10-step and 20-step PGD attacks, and adaptive auto attack ($A^3$)~\citep{attack_aaa} for reliable evaluation. We used a single RTX 8000 GPU with CUDA11.6 and CuDNN7.6.5 in our experiments.

\subsection{Experiment details and futher experiments on subset training}
\label{appendix:implementation_experiment_details_figure1}

\paragraph{Figure~\ref{fig:introduction_testacc_from_pgd}} For the experiments in Fig.~\ref{fig:introduction_testacc_from_pgd}, we executed 200 training epochs on the CIFAR-10 dataset in both standard and adversarial training. We constituted datasets for both standard and adversarial~\citep{attack_pgd} training by selecting 5k hard and 5k random examples from the training set, where the difficulty of data is measured by accumulating 0-1 loss along the training trajectory of the models trained with standard and adversarial training until epoch 90. Considering that the size of the dataset is small and to prevent overfitting before the learning rate decay in both standard and adversarial training, we used the ResNet-18~\citep{arch_resnet} architecture. The $\ell_\infty\text-$bound and step-size of adversarial examples were set to 0.0155, and 0.0035, respectively, with 5 steps.
The adversarial training time on the 5k+5k subset of the CIFAR-10 dataset is 2 hours for 200 epochs. The adversarial training time on the 5k subset (hard excluded) of the CIFAR-10 dataset is 0.5 hours for 100 epochs (after the learning rate decay epoch 100).

\input{fig/fig_method_acc_pgd_small}

For further experiments in adversarial training, we constituted a new dataset as in Fig.~\ref{fig:introduction_testacc_from_pgd}, with only using top 10k hard and 10k random examples (total 20k). We trained the PGD model until learning rate decay, then progress training for the models with the dataset: (1) entire examples, (2) 10k random examples (hard pruning), (3) top 10k hard examples (hard only). We used 10k instead of 5k to prevent overfitting before pruning.
In Fig.~\ref{fig:method_testacc_from_pgd_small}, the models trained with the dataset that includes hard examples show decrease in test robust accuracy; however, the model trained with the dataset that hard examples are pruned shows increases in test robust accuracy.
Thus, pruning hard examples improves robustness, indicating that hard examples degrades the performance in adversarial training.
The implementation details for Fig.~\ref{fig:method_testacc_from_pgd_small} are same with those of Fig.~\ref{fig:method_norm_ratio}. The adversarial training time on the 10k+10k subset of the CIFAR-10 dataset is 20 hours for 200 epochs. The adversarial training time on the 10k subset (hard excluded) of the CIFAR-10 dataset is 5 hours for 100 epochs (after learning rate decay epoch 100).

\subsection{Experiment details for comparing el2n score}
\label{appendix:implementation_experiment_details_el2n}

\paragraph{Figure~\ref{fig:method_norm_ratio}} \hgg{For the experiments in Fig.~\ref{fig:method_norm_ratio}, we executed 200 training epochs on the CIFAR-10 dataset in each training. The other experimental details are summarized in Appendix~\ref{appendix:implementation_details}.
The adversarial training time on the entire CIFAR-10 dataset is 47 hours for 200 epochs. The full results of Fig.~\ref{fig:method_norm_ratio} can be seen in Fig.~\ref{fig:appendix_norm_ratio}. For the results of the PGD model before the learning rate decay, because hard examples are difficult to learn also in the perspective of standard accuracy with high learning rate, the result shows large EL2N score for hard clean examples. After the learning rate decay, where the model starts to fit hard examples, the difficulty difference between clean and adversarial examples of hard examples starts to increase.
}

\input{fig/fig_supple_easy_hard_ratio_full}

\subsection{Experiment details and further experiments on other analyses}
\label{appendix:implementation_experiment_memorization_and_pruning}

\paragraph{Figure~\ref{fig:method_demo}} 
 We executed 200 training epochs on the CIFAR-10 dataset in each training. The other experimental details are summarized in Appendix~\ref{appendix:implementation_details}. 
 The hard and easy examples are selected according to the difficulty of each model. The training and test accuracy results and loss results for the PGD model with $\epsilon=4$ and $\epsilon=2$ can be seen from Fig.~\ref{fig:appendix_demo}. The robust overfitting phenomenon is observed to be severe as the adversarial budget of the training setting increases.

\input{fig/fig_supple_demo}

\paragraph{Figure~\ref{fig:method_memorization}} 
 We executed 200 training epochs on the CIFAR-10 dataset in each training. The other experimental details are summarized in Appendix~\ref{appendix:implementation_details}.
 We select top 5k easy and top 5k hard examples. We then trained the models with the dataset that prunes top 5k easy examples and that prunes top 5k hard examples, respectively.
 The memorization results including experiments on adversarial training with $\epsilon=4$, $\epsilon=2$, and $\epsilon=0$ (STD) are described in Fig.~\ref{fig:appendix_memorization}.
 
 \jmlr{
 These results illustrate the necessity of a memorization feature. To explain this, let us consider the possibility that the results in Fig.~\ref{fig:method_memorization} could be explained solely by the presence of robust and non-robust features, without the need for a memorization feature. First, if we assume that hard examples possess robust features, then it should be possible to predict the adversarial examples of hard examples through the training of the remaining 45k data points of CIFAR-10 in Fig.~\ref{fig:method_memorization}, even if these hard examples are not included in the adversarial training. However, since the model trained without the hard examples shows 0\% accuracy on them, this assumption contradicts the idea that hard examples contain robust features. Second, let us assume that hard examples are composed solely of non-robust features. In this case, the train accuracy for these examples should be 0\%. However, Fig.~\ref{fig:method_memorization} shows 100\% train accuracy, which contradicts this assumption. Therefore, the assumption that the results can be explained solely by the presence of robust and non-robust features without a memorization feature is difficult to support. To explain the results in Fig.~\ref{fig:method_memorization}, the existence of a feature that ``acts as a robust feature during training but as a non-robust feature during testing''—in other words, a memorization feature—is necessary. 
 }

\input{fig/fig_supple_memorization}

\paragraph{{Table~\ref{tab:method_subset}}} 
For the experiments in Table~\ref{tab:method_subset}, we executed 110 training epochs on the CIFAR-10 dataset in each model. The other experimental details are summarized in Appendix~\ref{appendix:implementation_details}. Table~\ref{tab:appendix_subset} shows the performance of the models trained with \jmlr{top 10k easy and top 10k hard examples as Table~\ref{tab:method_subset}} with varying the adversarial budget $\epsilon$. The robust accuracy is evaluated using $A^3$~\citep{attack_aaa}. As the adversarial budget decreases, the standard accuracy of the models that prunes hard examples increases, which suggests that the number of anti-correlated features which were slightly correlated features decreases.

\input{table/tab_supple_subset}

\paragraph{Figure~\ref{fig:method_corruption_plot}}
We assigned random labels to the hard examples for the corruption model from the beginning of the training, and we pruned the hard examples from the beginning of the training for the pruning model.
The robust accuracy is evaluated using $A^3$~\citep{attack_aaa}. Table~\ref{tab:appendix_noise} shows the final accuracy for the models trained with applying random labels to the top 5k hard examples. We compared the results with the performance of the normal models and the models that prunes top 5k hard examples for PGD and TRADES. Observably, assigning random labels to hard examples does not change the robustness performance.

\input{table/tab_supple_noise}

\paragraph{Distinguishing hard examples}

\jmlrr{
In Section \ref{method:theoretical training of hard examples}, we assumed properties of hard examples and conducted a theoretical analysis on the impact of learning hard examples on model performance. Based on this theoretical analysis, we can distinguish hard examples using the following metrics:
}
\jmlrr{
\begin{enumerate}
    \item ${{\text{Pr}_{f \sim A(D_{adv})}}} [\argmax f(x_i)=y_i] - {{\text{Pr}_{f \sim A(D_{adv} \backslash \{i\}}) }} [\argmax f(x_i)=y_i]$
    \item $\mathbb{E}_{(x,y) \sim D_{std}} \left[ Pr\left( f_{std}(x;A(D_{std})) = y \right) - Pr\left( f_{std}(x; A(D_{std} \setminus \{i\})) = y \right) \right] \\ - \mathbb{E}_{(x,y) \sim D_{adv}} \left[ Pr\left( f_{adv}(x; A(D_{adv})) = y \right) - Pr\left( f_{adv}(x; A(D_{adv} \setminus \{i\})) = y \right) \right]$
    \item $\mathbb{E}_{(x,y) \sim D_{std}} \left[ Pr\left( f_{adv}(x;A(D_{adv})) = y \right) - Pr\left( f_{adv}(x; A(D_{adv} \setminus \{i\})) = y \right) \right] \\ - \mathbb{E}_{(x,y) \sim D_{adv}} \left[ Pr\left( f_{adv}(x; A(D_{adv})) = y \right) - Pr\left( f_{adv}(x; A(D_{adv} \setminus \{i\})) = y \right) \right]$
\end{enumerate}
}

\jmlrr{
In all the above metrics, a higher value indicates a stronger representation of the properties presented in Assumption 1. The first metric is the same as the memorization score \citep{memorization}, which reflects the difference in performance for a data point $\left(x_i,y_i\right)$ when it is included in training versus when it is not. The second metric compares the difference for the data point $\left(x_i,y_i\right)$ between standard training (with and without inclusion in training) and adversarial training (with and without inclusion). The third metric measures the trade-off $\left(x_i,y_i\right)$ by comparing the standard accuracy and adversarial accuracy in adversarial training when the data point is included versus when it is excluded from training. 
}

\jmlrr{
While the metrics above can identify data with high values as hard examples, it is challenging to train every model with and without each data point for comparison. Therefore, as in Section \ref{method:effect of hard examples}, we conducted analyses in Section \ref{method:empirical verification of hard examples} using hard examples identified through 0-1 loss to check whether data that is difficult to learn actually exhibits the properties assumed for hard examples in Assumption \ref{eq:method_assumption_hard_example}.
}

\paragraph{Figure~\ref{fig:method_excluding}} 
\hgg{For the experiments in Fig.~\ref{fig:method_excluding}, We executed 110 training epochs on the CIFAR-10 dataset in each model. The other experimental details are summarized in Appendix~\ref{appendix:implementation_details}. 
We trained the baseline models until epoch 90, which is the 10 epoch before the learning rate decay for the stability of the training, then we continued by the training using the dataset with the subset pruned as the experiments in Fig.~\ref{fig:introduction_testacc_from_pgd}. We evaluated the performance of the best checkpoint in each model using $A^3$~\citep{attack_aaa}. 
}

\input{table/tab_supple_implementations_factor}

\subsection{Experiment details and ablation studies of DPLS}
\label{appendix:implementation_experiment_dpls}

\paragraph{Table~\ref{tab:experiment_all110_main_2}} We executed 110 training epochs on the CIFAR-10 dataset in adversarial training.
The other experimental details are summarized in Appendix~\ref{appendix:implementation_details}.
\hgg{We mainly use 0-1 loss as difficulty measurement loss for DPLS.
We set the label smoothing factor of the most difficult example in DPLS considering the average difficulty score of each dataset and algorithm. We tuned the factor such that the average label smoothing factor is between 0.8 and 0.9 in 10-class datasets (CIFAR-10, SVHN, \jmlr{and STL-10}) and is between 0.7 and 0.9 in a 100-class dataset (CIFAR-100). We selected difficulty calculation epoch as $T=90$, which is 10 epochs before the first learning rate decay.
The label smoothing factor hyperparameter $\lambda$ of the most difficult example for the DPLS models and the label smoothing factor for LS models in Table~\ref{tab:experiment_all110_main_2} of the main paper is summarized in Table~\ref{tab:supple_implementations_factor}. The PGD model in SVHN with label smoothing methods shows a result similar to gradient masking effect~\citep{adv_obfuscated}. We applied DPLS only to the labels of outer maximization loss to reduce the effect. However, it is considered as the properties of the SVHN dataset with PGD and label smoothing because the experiments on other datasets and algorithms do not show gradient masking. Furthermore, the results of our method still show improvement also against $A^3$ for SVHN by 6.81\%p, which could ignore gradient masking and evaluate true robustness. It means that our method increases the true robustness of the model also for SVHN.
}

\paragraph{Figure~\ref{fig:experiment_acc_from_pgd_all} and Table~\ref{tab:experiment_overfitting_methods}} For the experiments of previous robust overfitting mitigation methods, we used the code uploaded in the official Github by the author of each method.
The initial learning rate was set to 0.1, and the learning rate decay was applied at the learning rate decay scheduling of each method with total training epochs 200 and with a decay factor of 0.1. The learning rate decay was applied at 50 and 150 epochs for the method of knowledge distillation with stochastic weight averaging (KD+SWA)~\citep{overfitting_adv_kd} and the learning rate decay was applied at 100 and 150 epochs for the other methods (the baseline model, our method (DPLS), self-adpative training (SAT)~\citep{overfitting_sat}, and temporal ensemble (TE)~\citep{overfitting_adv_memorization}).
We executed 200 training epochs on the CIFAR-10 dataset in adversarial training. The other experimental details are summarized in Appendix~\ref{appendix:implementation_details}.

\input{table/tab_supple_difficulty}

\paragraph{Table~\ref{tab:experiment_difficulty_loss}}

For the experiments in Table~\ref{tab:experiment_difficulty_loss}, we executed 110 training epochs on the subsets of CIFAR-10
dataset in each model. The other experimental details are summarized in Appendix~\ref{appendix:implementation_details}. 
We employed the pre-calculated C-score~\citep{importance_cscore} from a standard model, and the difficulty from the other losses was calculated during training according to the procedure described in Algorithm~\ref{algorithm:DPLS}. The detailed results of the performance of DPLS according to various difficulty losses are listed in Table~\ref{tab:appendix_difficulty_loss}.

\input{table/tab_supple_algorithms_main}

\paragraph{Table~\ref{tab:experiment_training_algorithms}} For the experiments of applying DPLS to other adversarial training algorithms, we used the code uploaded in the official Github by the author of each method.
We executed 110 training epochs on the CIFAR-10 dataset in adversarial training.
The other experimental details are summarized in Appendix~\ref{appendix:implementation_details}.
Table~\ref{tab:mart} and Table~\ref{tab:rst} show the experimental details of applying DPLS to misclassification aware adversarial training (MART)~\citep{defense_mart} and robust self training (RST)~\citep{defense_rst}. We use the 1,000k extra data generated by a denoising diffusion probabilistic model (DDPM)~\citep{dataset_ddpm} as in~\citet{adv_ddpm}.
For the MART models, we used the difficulty score calculated from the TRADES model because MART uses weighted loss considering difficulty (ground-truth class confidence) of each sample, in which the trained model by MART can be biased and miscalculate the difficulty of each sample.
The MART model combined with DPLS where the difficulty is calculated from TRADES shows higher performance than the other models. It can be inferred that calculating accurate difficulty score is important, but applying DPLS with approximate difficulty is also effective at improving robustness.
For the RST models, we applied DPLS to both labeled and unlabeled dataset (Sup+Unsup), only the labeled dataset (Sup), and only the unlabeled dataset (Unsup). We calculated difficulty of the unlabeled dataset as the ratio of the number of training to the number of right predictions. For the DPLS (Sup) model, it is inferred that the model is trained to depend on the unlabeled dataset because of regularization only on the labeled data. For the DPLS (Unsup) model, although the difficulty of the unlabeled dataset is not calculated from the full trajectory, the DPLS successfully mitigates the negative effect of hard examples for the unlabeled dataset.

\input{fig/fig_supple_examples_ratio}

\input{fig/fig_supple_data_hist}

\input{fig/fig_supple_examples}

\subsection{Other experiment details}
\label{appendix:implementation_experiment_other}

\paragraph{Additional details of easy and hard examples}

Fig.~\ref{fig:supple_examples_distribution} shows the class distributions of 10k easy and 10k hard examples in the training dataset of CIFAR-10 selected by using accumulated 0-1 loss difficulty along the training trajectory of TRADES~\cite{defense_trades}. Fig.~\ref{fig:supple_dataset_hist} shows the 0-1 loss distributions of the training datasets for CIFAR-10, CIFAR-100, and SVHN, which are calculated with the training trajectory of TRADES. The value of 0-1 loss is normalized by using min-max normalization. Fig.~\ref{fig:supple_easy_hard_examples} shows easy and hard examples in the training dataset of CIFAR-10.

%% file: fig/fig_method_acc_pgd_small.tex
\begin{wrapfigure}[13]{R}{0.42\linewidth} %
\begin{center}
\vskip -2mm
\raisebox{0pt}[\dimexpr\height-1.5\baselineskip\relax]{\includegraphics[width=1.0\linewidth]{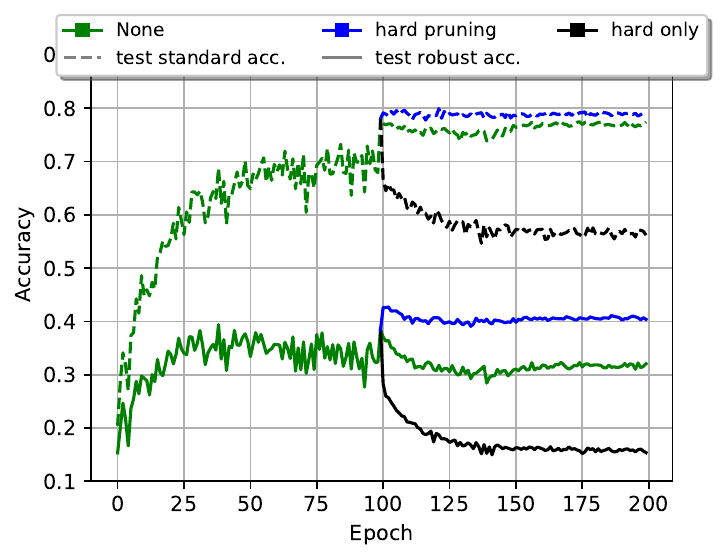}}%
\end{center}
\vskip -6mm
\caption{Test accuracy of PGD models trained with subsets of CIFAR-10.
}
\label{fig:method_testacc_from_pgd_small}
\end{wrapfigure}

%% file: fig/fig_supple_easy_hard_ratio_full.tex
\begin{figure}[h]
\centering
{
\includegraphics[width=0.8\linewidth]{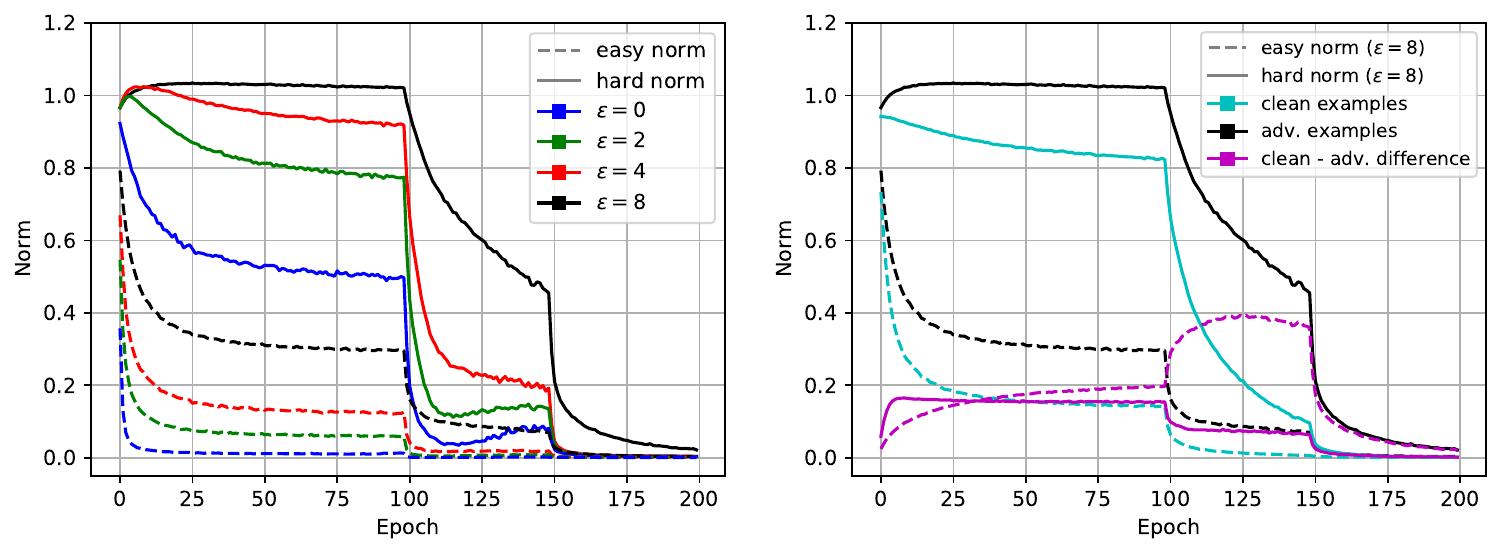}}
\caption{Average error vector norm (EL2N score) of hard and easy examples.}
\vskip -3mm
\label{fig:appendix_norm_ratio}
\end{figure}

%% file: fig/fig_supple_demo.tex
\begin{figure}[h]
  \centering
  \begin{subfigure}[b]{0.32\linewidth}
    \includegraphics[width=\linewidth]{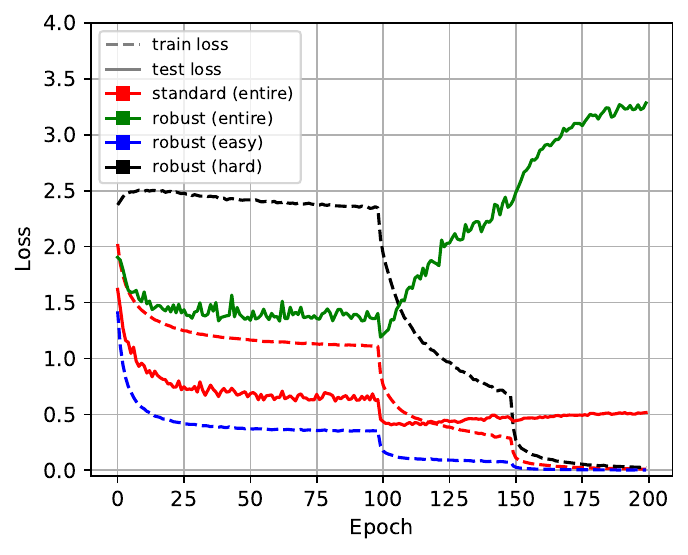}
  \caption{}
  \end{subfigure}
  \hfill
  \begin{subfigure}[b]{0.32\linewidth}
    \includegraphics[width=\linewidth]{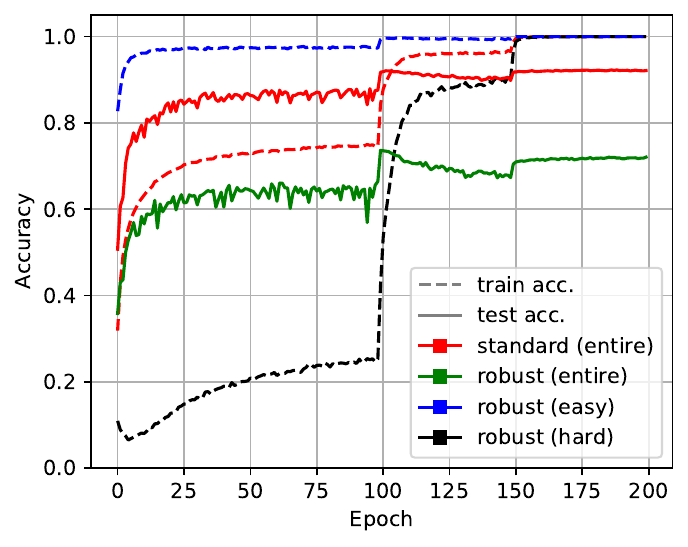}
  \caption{}
  \end{subfigure}
  \hfill
  \begin{subfigure}[b]{0.32\linewidth}
    \includegraphics[width=\linewidth]{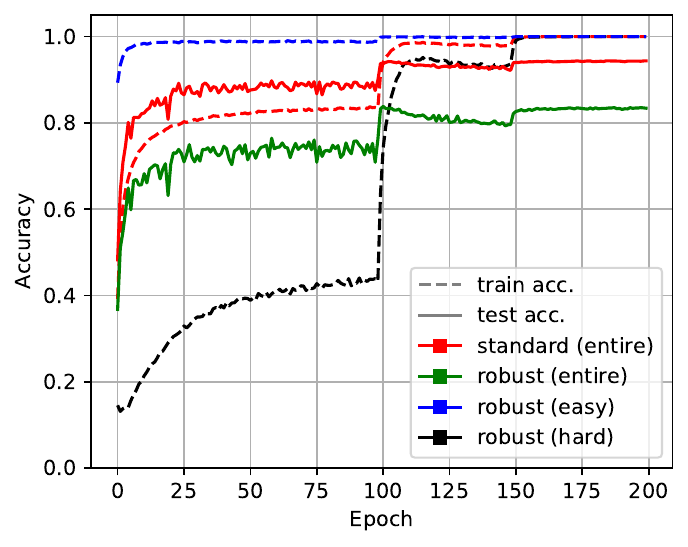}
  \caption{}
  \end{subfigure}
  \caption{(a) presents the training and test loss curves in the PGD model. (b) shows the training and test accuracy in the PGD models with $\epsilon=4$. (c) shows the training and test accuracy in the PGD models with $\epsilon=2$.}
  \label{fig:appendix_demo}
\end{figure}

%% file: fig/fig_supple_memorization.tex
\begin{figure}[h]
  \centering
  \begin{subfigure}[b]{0.24\linewidth}
    \includegraphics[width=\linewidth]{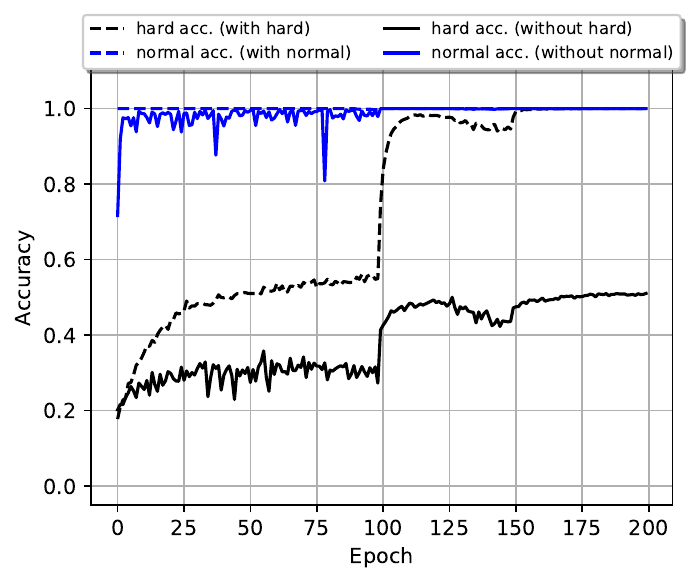}
    \caption{$\epsilon=0$}
    \label{fig:supple_memorization_std}
  \end{subfigure}
  \hfill
  \begin{subfigure}[b]{0.24\linewidth}
    \includegraphics[width=\linewidth]{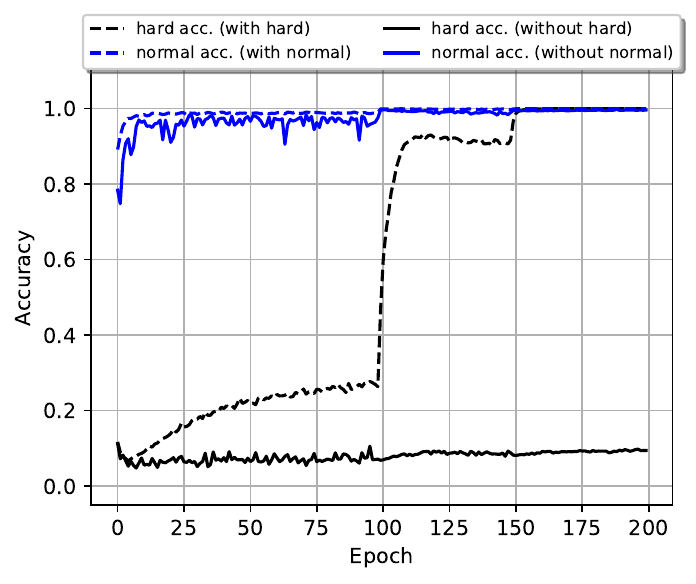}
    \caption{$\epsilon=2$}
    \label{fig:supple_memorization_pgd_0.25x}
  \end{subfigure}
  \hfill
  \begin{subfigure}[b]{0.24\linewidth}
    \includegraphics[width=\linewidth]{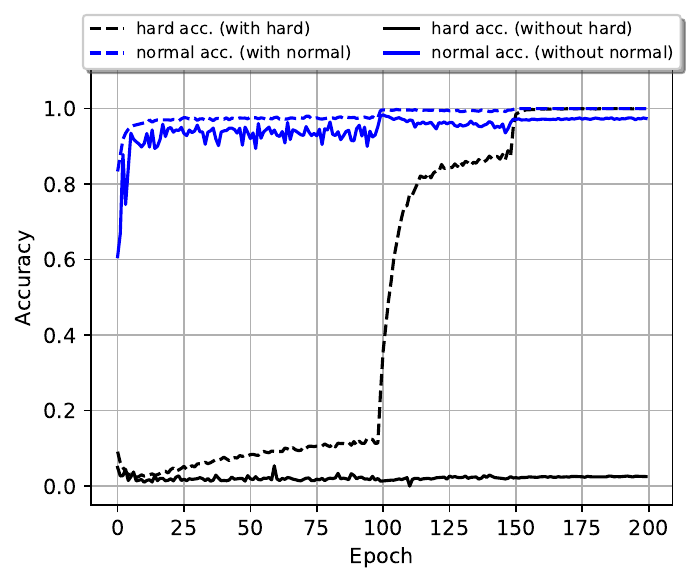}
    \caption{$\epsilon=4$}
    \label{fig:supple_memorization_pgd_0.5x}
  \end{subfigure}
  \hfill
  \begin{subfigure}[b]{0.24\linewidth}
    \includegraphics[width=\linewidth]{fig/_fig_method_memorization_v3.pdf}
    \caption{$\epsilon=8$}
    \label{fig:supple_memorization_pgd}
  \end{subfigure}
  \caption{Memorization of easy and hard examples in each setting. Clean example results for a standard setting and adversarial example results for adversarial settings.}
  \label{fig:appendix_memorization}
\end{figure}

%% file: table/tab_supple_subset.tex
\begin{table}[h]
\centering
\caption{\jmlr{Performance of models trained with 10k examples.}}
\begin{tabular}{c|cc|cc|cc|cc}
    \toprule
     \multicolumn{1}{c|}{\multirow{2}{*}{Method}} &
     \multicolumn{2}{c|}{\multirow{1}{*}{STD}} &
     \multicolumn{2}{c|}{\multirow{1}{*}{PGD ($\epsilon=2$)}} &
     \multicolumn{2}{c|}{\multirow{1}{*}{PGD ($\epsilon=4$)}} & 
     \multicolumn{2}{c}{\multirow{1}{*}{PGD ($\epsilon=8$)}} \\
     & Std. & \multicolumn{1}{c|}{\multirow{1}{*}{Rob.}} 
     & Std. & \multicolumn{1}{c|}{\multirow{1}{*}{Rob.}} 
     & Std. & \multicolumn{1}{c|}{\multirow{1}{*}{Rob.}}
     & Std. & \multicolumn{1}{c}{\multirow{1}{*}{Rob.}} \\
    \midrule[.05em]     
     \jmlr{Easy} & \jmlr{78.46} & - & \jmlr{75.66} & \jmlr{62.00} &  \jmlr{69.65} & \jmlr{51.51} & \jmlr{61.63} & \jmlr{39.56} \\
     \jmlr{Hard} & \jmlr{77.60} & - & \jmlr{45.47} & \jmlr{22.54} & \jmlr{32.17} & \jmlr{9.04} & \jmlr{19.92} & \jmlr{8.68} \\
    
    \midrule[.1em]
\end{tabular}
\label{tab:appendix_subset}
\end{table}

%% file: table/tab_supple_noise.tex
\begin{table}[h]
\centering
\caption{{Performance comparison
of the final accuracy for models with label corruption.}}
\vspace{0pt}
\begin{tabular}{c|ccc|ccc}
    \toprule
     \multicolumn{1}{c|}{\multirow{3}{*}{Dataset}} & \multicolumn{3}{c|}{PGD} &
     \multicolumn{3}{c}{TRADES} \\
    & \multicolumn{1}{c}{\multirow{2}{*}{Std.}} & Robust & \multicolumn{1}{c|}{Robust}
    & \multicolumn{1}{c}{\multirow{2}{*}{Std.}} & Robust & Robust
    \\
    & & (PGD) & \multicolumn{1}{c|}{($A^3$)} & & (PGD) & ($A^3$) \\
    
    \midrule[.05em]
    Normal & \textbf{86.82} & 50.25 & 46.07
      & \textbf{84.83} & 51.43 & 46.69 
    \\
    Corrupted & 82.31 & 49.32 & 45.67
            & 82.11 & 50.64 & 46.15    
    \\
    Pruned & 85.99 & \textbf{52.36} & \textbf{48.81}
    & 84.71 & \textbf{53.44} & \textbf{48.78}
    \\
    
    \midrule[.1em]
\end{tabular}
\label{tab:appendix_noise}
\end{table}

%% file: table/tab_supple_implementations_factor.tex
\begin{table*}[h]
\centering
\footnotesize
\addtolength{\tabcolsep}{-1pt}
\caption{The smoothing factor $\lambda$ and the average of label smoothing factor for each model in Table~\ref{tab:experiment_all110_main_2} in the main paper. }
\begin{tabular}{ccc|cc}
\toprule
\multicolumn{1}{c}{\multirow{1}{*}{Dataset}} &
\multicolumn{1}{c}{\multirow{1}{*}{Model}} & \multicolumn{1}{c|}{\multirow{1}{*}{Method}}
& \multirow{1}{*}{$\lambda$ (smoothing factor)} & Avg. smoothing
\\
\midrule[.05em]
\multirow{4}{*}{CIFAR-10}
    & \multirow{2}{*}{PGD}
    & LS & 0.900 & 0.900 \\
    & & DPLS & 0.500 & 0.897 \\
    
    \cmidrule{2-5}
    & \multirow{2}{*}{TRADES}
    & LS & 0.900 & 0.900 \\
    & & DPLS & 0.500 & 0.893 \\
    \midrule[.1em]
\multirow{4}{*}{CIFAR-100}
    & \multirow{2}{*}{PGD}
    & LS & 0.900 & 0.900 \\
    & & DPLS & 0.700 & 0.862 \\
        \cmidrule{2-5}
    & \multirow{2}{*}{TRADES}
    & LS & 0.700 & 0.700 \\
    & & DPLS & 0.300 & 0.703\\
        \midrule[.1em]
        
\multirow{4}{*}{SVHN}
    & \multirow{2}{*}{PGD}
    & LS & 0.900 & 0.900 \\
    & & DPLS & 0.100 & 0.940 \\
        \cmidrule{2-5}
    & \multirow{2}{*}{TRADES}
    & LS & 0.900 & 0.900 \\
    & & DPLS & 0.100 & 0.927 \\
    \midrule[.1em]
\multirow{4}{*}{\jmlr{STL-10}}
    & \multirow{2}{*}{PGD}
    & LS & \jmlr{0.800} & \jmlr{0.800} \\
    & & DPLS & \jmlr{0.500} & \jmlr{0.788} \\
        \cmidrule{2-5}
    & \multirow{2}{*}{TRADES}
    & LS & \jmlr{0.800} & \jmlr{0.800} \\
    & & DPLS & \jmlr{0.500} & \jmlr{0.797} \\

\midrule[.1em]
\end{tabular}
\label{tab:supple_implementations_factor}
\end{table*}

%% file: table/tab_supple_difficulty.tex
\begin{table}[h]
\centering
\caption{Performance of DPLS according to various difficulty measurement losses.}
\vspace{0pt}
\begin{tabular}{c|ccc|ccc}
    \toprule
     \multicolumn{1}{c|}{\multirow{3}{*}{Method}} & \multicolumn{3}{c|}{PGD} &
     \multicolumn{3}{c}{TRADES} \\
    & \multicolumn{1}{c}{\multirow{2}{*}{Std.}} & Robust & \multicolumn{1}{c|}{Robust}
    & \multicolumn{1}{c}{\multirow{2}{*}{Std.}} & Robust & Robust
    \\
    & & (PGD) & \multicolumn{1}{c|}{($A^3$)} & & (PGD) & ($A^3$) \\
    
    \midrule[.05em]
    - & 87.19 & 56.44 & 51.8
      & 85.66 & 58.46 & 54.08 
    \\
    C-score & \textbf{88.05} & 57.58 & 52.9
            & 86.6 & 59.0 & 54.71 
    \\
    CE loss & 87.51 & 57.47 & 52.73
            & 86.02 & 58.64 & 54.45
    \\
    EL2N & 87.27 & 58.89 & 53.38
    & \textbf{86.94} & 58.94 & 54.32 
    \\
    0-1 loss & 87.13 & \textbf{58.43} & \textbf{54.03}
    & 85.36 & \textbf{59.33} & \textbf{55.19}
    \\
    
    \midrule[.1em]
\end{tabular}
\label{tab:appendix_difficulty_loss}
\end{table}

%% file: table/tab_supple_algorithms_main.tex
\begin{table}[h]
\begin{minipage}{.48\linewidth}
\caption{Applying DPLS to MART~\citep{defense_mart}. DPLS model denotes the model that calculate the difficulty of training examples for DPLS.}
\centering
\footnotesize
\addtolength{\tabcolsep}{-0pt}
\begin{tabular}{cc|cc}
    \toprule
     \multicolumn{1}{c}{\multirow{1}{*}{Method}} &
     \multicolumn{1}{c|}{\begin{tabular}[c]{@{}c@{}}DPLS\\ model\end{tabular}}
     &\multicolumn{1}{c}{Standard} &
    \multicolumn{1}{c}{\begin{tabular}[c]{@{}c@{}}Robust\\ ($A^3$)\end{tabular}} \\
    
    \midrule[.05em]
    \multirow{3}{*}{MART} & -  & \textbf{83.9} & 52.19 \\
     & MART  & 83.18 & 52.84 \\
     & TRADES & 83.81 & \textbf{53.2} \\
    \midrule[.1em]
\end{tabular}
\label{tab:mart}
\end{minipage}%
\hfill
\begin{minipage}{.48\linewidth}
\centering
\footnotesize
\addtolength{\tabcolsep}{-0pt}
\caption{Applying DPLS to RST~\citep{defense_rst}. DPLS dataset denotes the dataset to which DPLS is applied.}
\begin{tabular}{cc|cc}
    \toprule
     \multicolumn{1}{c}{\multirow{1}{*}{Method}} &
     \multicolumn{1}{c|}{\begin{tabular}[c]{@{}c@{}}DPLS\\ dataset\end{tabular}}
     &\multicolumn{1}{c}{Standard} &
    \multicolumn{1}{c}{\begin{tabular}[c]{@{}c@{}}Robust\\ ($A^3$)\end{tabular}} \\
    
    \midrule[.05em]
    \multirow{4}{*}{RST} & -  & 84.82 & 57.26 \\
     & Sup  & 84.36 & 57.35 \\
     & Unsup    & \textbf{85.2} & 57.62 \\
     & Sup+Unsup    & 84.43 & \textbf{57.81} \\
    \midrule[.1em]
\end{tabular}
\label{tab:rst}
\end{minipage} 
\end{table}

%% file: fig/fig_supple_examples_ratio.tex
\begin{figure}[h]
  \centering
  \begin{subfigure}[b]{0.3\linewidth}
    \includegraphics[width=\linewidth]{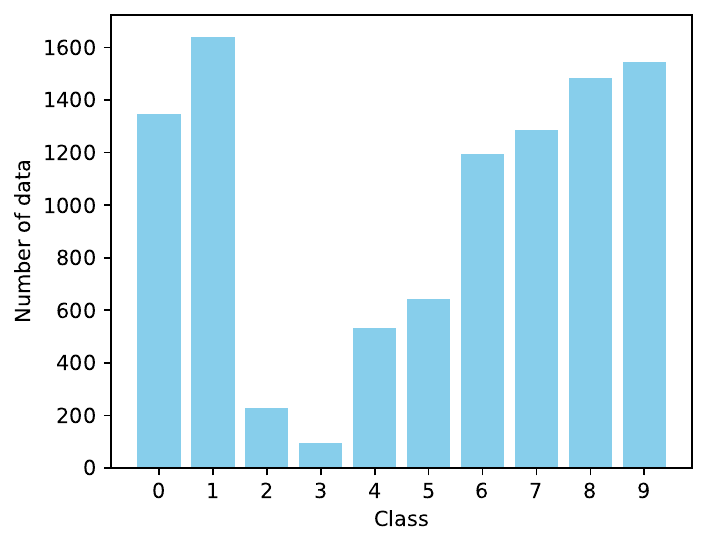}
    \caption{Easy examples}
    \label{fig:supple_easy_examples_distribution}
  \end{subfigure}
  \hfill
  \begin{subfigure}[b]{0.3\linewidth}
    \includegraphics[width=\linewidth]{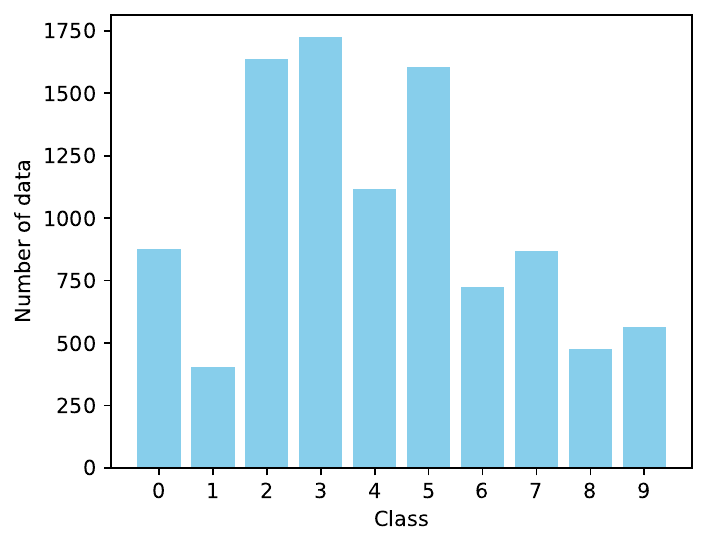}
    \caption{Hard examples}
    \label{fig:supple_hard_examples_distribution}
  \end{subfigure}
  \caption{The class distributions of 10k easy and 10k hard examples in CIFAR-10 selected by using accumulated 0-1 loss difficulty.}
  \label{fig:supple_examples_distribution}
\end{figure}

%% file: fig/fig_supple_data_hist.tex
\begin{figure}[h]
  \centering
  \begin{subfigure}[b]{0.3\linewidth}
    \includegraphics[width=\linewidth]{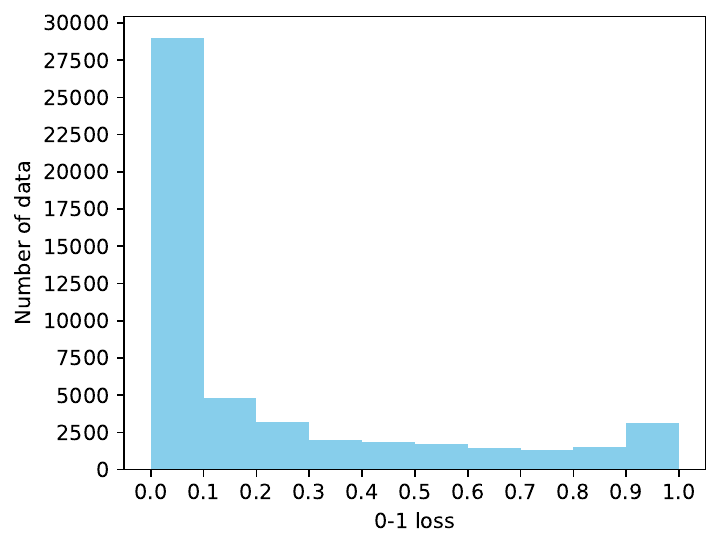}
    \caption{CIFAR-10}
    \label{fig:supple_cifar10_hist}
  \end{subfigure}
  \hfill
  \begin{subfigure}[b]{0.3\linewidth}
    \includegraphics[width=\linewidth]{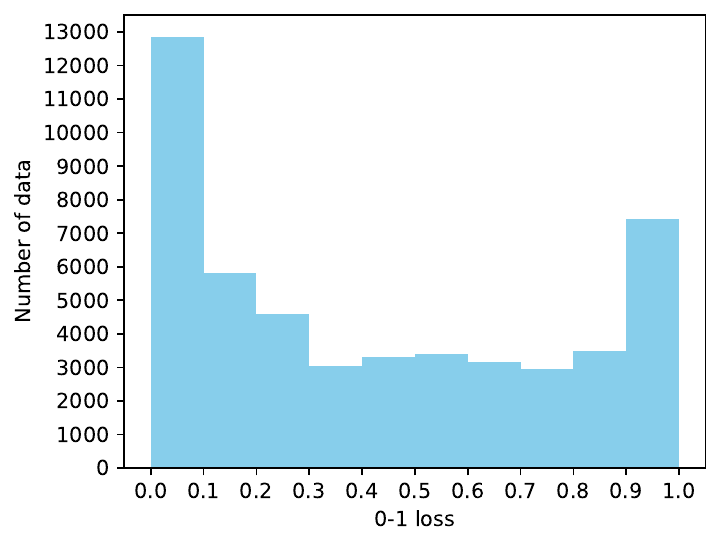}
    \caption{CIFAR-100}
    \label{fig:supple_cifar100_hist}
  \end{subfigure}
  \hfill
  \begin{subfigure}[b]{0.3\linewidth}
    \includegraphics[width=\linewidth]{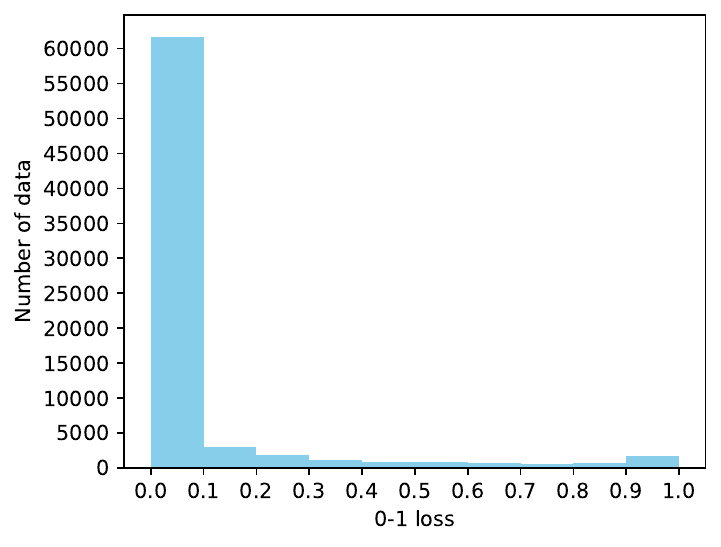}
    \caption{SVHN}
    \label{fig:supple_svhn_hist}
  \end{subfigure}
  \caption{The 0-1 loss distributions of the each training dataset: CIFAR-10, CIFAR-100, and SVHN.}
  \label{fig:supple_dataset_hist}
\end{figure}

%% file: fig/fig_supple_examples.tex
\begin{figure}[h]
  \centering
  \begin{subfigure}[b]{0.95\linewidth}
    \includegraphics[width=\linewidth]{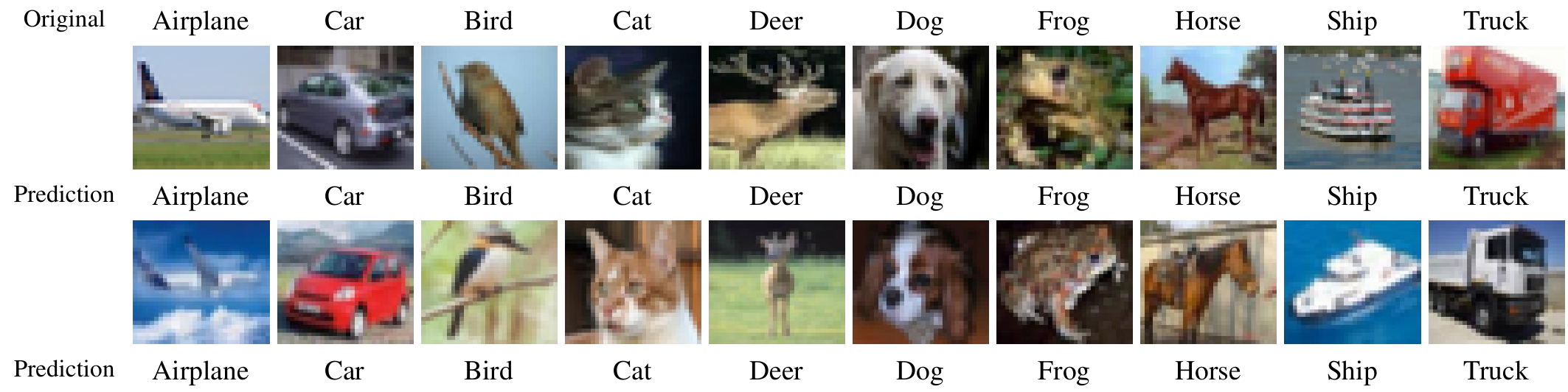}
    \caption{Easy examples}
    \label{fig:supple_easy_examples}
  \end{subfigure}
  \hfill
  \begin{subfigure}[b]{0.95\linewidth}
    \includegraphics[width=\linewidth]{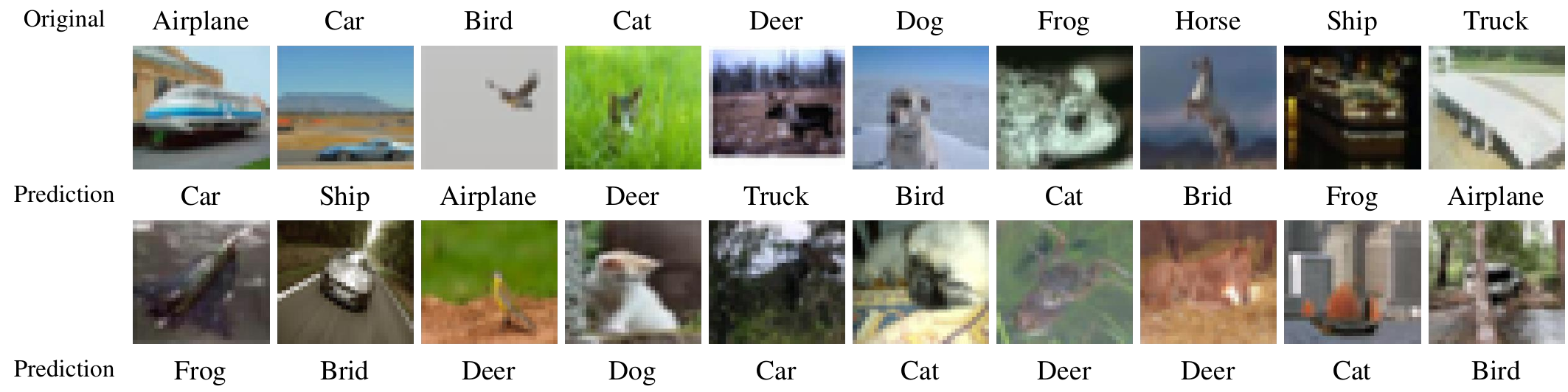}
    \caption{Hard examples}
    \label{fig:supple_hard_examples}
  \end{subfigure}
  \caption{Easy and hard examples in CIFAR-10~\cite{dataset_cifar} selected by using accumulated 0-1 loss difficulty. Original denotes the ground-truth class of images and prediction denotes the most frequently predicted class of images at each epoch of training until the calculation epoch.}
  \label{fig:supple_easy_hard_examples}
\end{figure}

%% file: D_RelatedWork.tex
\section{Related work}

\citet{overfitting_adv} conducted analysis on overfitting in adversarial training, which is termed as robust overfitting. They applied various data augmentation and regularziation methods to reduce the generalization gap between training and test robust accuracy, concluding that the effect of early stopping can be matched by other methods. \citet{overfitting_adv_kd} utilized smoothing of both logits and weights in adversarial training. They applied knowledge distillation~\citep{smoothing_kd} to logits instead of label smoothing~\citep{arch_inception} with robust-trained and standard-trained models as teacher models and used stochastic weight average~\citep{smoothing_swa} as a weight smoothing method by averaging weights in another model along the training trajectory. They demonstrated that the smoothing methods are effective at mitigating robust overfitting. \citet{overfitting_aug} combined data augmentation methods and weight averaging and significantly improved robust accuracy. They noted that weight averaging is effective when robust accuracy between training iterations is maintained. They experimented on several heuristics-driven augmentation such as cutout~\citep{aug_cutout}, mixup~\citep{aug_mixup}, and cutmix~\citep{aug_cutmix}.
\citet{overfitting_adv_memorization} demonstrated that DNNs are sufficient to memorize training adversarial examples with random labels. They showed that the memorization of hard training examples leads to decrease in training loss but increase in test error, noting that the causes of robust overfitting may be the memorization of one-hot labels. Specifically, they showed that training adversarial examples with completely random labels are possible when the training algorithm is TRADES~\citep{defense_trades}. For the PGD~\citep{attack_pgd} models, when the label noise ratio exceeds a certain threshold, the generalization gap between train and test accuracy becomes zero and the model exhibits a performance similar to that of the random prediction.

\input{table/tab_supple_compare_reweight_2}

In~\citet{importance_difficulty}, they utilize average training loss as difficulty measurement and analyze overfitting mitigation methods in adversarial training. However, instead of using difficulty score to improve the performance of adversarial training, they proposed Adpative Target and Reweight by using moving average predictions as in~\citet{overfitting_sat,overfitting_adv_memorization} and ground truth prediction as in~\citet{defense_mart}. Because the performance for standard adversarial training is not provided in the paper, but the performance only for fast adversarial training and fine-tuning with extra data is provided, we trained and evaluated their models for the comparison. 
The results of Table~\ref{tab:supple_compare_reweight} shows that Reweight is not applicable to the standard adversarial training and Adaptive Target shows the same performance as SAT~\citet{overfitting_sat} because it is simply a linear combination of SAT and one-hot label.
In~\citet{importance_difficulty}, they constituted a subset dataset in a balanced form between classes. However, the difficulty of classes can be imbalanced, and selecting the most difficult data considering the balance between classes can result in a deviation from correct observations. As noted in~\citet{adv_fairness}, adversarial training differently increases the robustness of each class in a dataset, so the performance imbalance between classes becomes more significant in adversarial training. Thus, it can be negligible in standard training to balance the number of each class in training example subsets, but it can be non-trivial in adversarial training. The distributions of hard and easy example subsets in our paper are plotted in Fig~\ref{fig:supple_examples_distribution}. The distributions of easy and hard example subsets are highly class-imbalanced. The comparison study on the pruning balanced or imbalanced training subsets with different difficulty losses can be seen in Table~\ref{tab:ablation_balancing}, where the models that prune imbalanced subsets consistently show higher performance than the models that prune balanced subsets. Furthermore, we showed that our method does not only exploit the performance increase of the easy classes, but increases the both performance of easy and hard classes as shown in Table~\ref{tab:ablation_fairness}.

%% file: table/tab_supple_compare_reweight_2.tex
\begin{table*}[h]
\centering
\caption{Performance comparison with the methods in~\citet{overfitting_reweight}. The models trained using the code in~\citet{overfitting_reweight} are denoted as (*). }
\begin{tabular}{l|ccc}
    \toprule
     \multicolumn{1}{c|}{\multirow{1}{*}{Method}} & \multicolumn{1}{c}{Std.} & 
    \multicolumn{1}{c}{\begin{tabular}[c]{@{}c@{}}Robust\\ (PGD)\end{tabular}}& 
    \multicolumn{1}{c}{\begin{tabular}[c]{@{}c@{}}Robust\\ ($A^3$)\end{tabular}} \\
    \midrule[.05em]
     PGD & \textbf{87.19} & 56.44 & 51.8 \\
     +DPLS & 86.17 & \textbf{58.14} & \textbf{53.98} \\
     \midrule[.05em]
     PGD (*)  & 86.75 & 56.24 & 51.81 \\
     +Reweight (*) & 25.91 & 22.16 & 21.93 \\
     +Adaptive Target (*) & 86.04 & 58.21 & 52.87 \\
     +Reweight+Adaptive Target (*) & 30.49 & 25.81 & 25.6 \\
    \midrule[.1em]
\end{tabular}
\label{tab:supple_compare_reweight}
\end{table*}

%% file: E_ablation.tex
\input{table/tab_supple_ablation_sup}

\section{Additional ablation study}
\label{appendix:ablation}

\paragraph{Ablation on the DPLS calculation epoch}
In Table~\ref{tab:ablation_1}, we conducted experiments varying the calculation epoch of DPLS.
The results show that applying DPLS at any epoch is effective at increasing robustness. 
However, the results show that the improvement is slightly higher when using the difficulty calculated until higher epoch. Additionally, when DPLS is applied after the learning rate decay, the result shows that the increase in robustness is diminished, but it still shows better performance than the baseline model.
Thus, the application of DPLS before memorization of hard examples can maximize the mitigation effect of the method, which indicates the overfitting prevention effect of DPLS.

\paragraph{Ablation on the re-initialization of training by using DPLS}

In Table~\ref{tab:ablation_2}, we conducted ablation study varying the training re-initialization epoch with pre-calculated DPLS by using the difficulty of that calculated until the training epoch 90.
There are small differences between the results.
Comparing the result in Table~\ref{tab:ablation_1} with the result in Table~\ref{tab:ablation_2}, it indicates that the difficulty score is important for DPLS and the applying checkpoint is less important after fitting training examples and before memorization of hard examples. However, all models still show the improvement in the robustness performance of the model as in the above DPLS calculation epoch ablation study.

\input{table/tab_supple_ablation_sup2}
\input{table/tab_supple_ablation_sup3}

\paragraph{Ablation study on the factor of DPLS}

Table~\ref{tab:ablation_dpls_param_pgd} and Table~\ref{tab:ablation_dpls_param_trades} show the performance of DPLS applying to the PGD and TRADES models on CIFAR-10, respectively. We calculated the average value for the smoothing factor of DPLS.
In both results, it is observed that applying DPLS with the factor of high value further increases the robustness performance.

\paragraph{Ablation study on the regularization of DPLS}

We used an LS approach for mitigating the effect of hard examples, instead of simply reweighting the loss of hard examples. Pruning hard examples corresponds to assigning zero values to the loss of hard examples, so we compare the performance of DPLS with the reweighting method. We applied the difficulty of each example to the loss to reweight the loss of each example, which is the same difficulty with DPLS.
Because TRADES utilized the robustness loss without labels, we applied DPLS to cross-entropy loss for clean examples and reweighted the robust loss with the same difficulty; thus, the difference between DPLS and the reweighting method in TRADES is the difference of application of the method to the loss for clean examples.
Table\ref{tab:appendix_reweight} lists the result of applying difficulty proportional reweight (DPR) and DPLS. In the results of both PGD and TRADES, DPLS shows higher standard accuracy. It is inferred that because the DPR method still ensures that the predictions are identical to one-hot labels and that the method focuses on the memorization of examples, the method improves robustness and deteriorates accuracy depending on robustness-accuracy trade-off.
In contrast, instead of memorizing hard examples with one-hot label, our method takes advantage of hard examples by allowing soft labels and learns distributions that smoothly include hard examples, which results in improving robustness with decreasing the reduction of standard performance.

\input{table/tab_supple_reweight}

\input{table/tab_supple_ablation_bal}

\paragraph{Study on balancing the number of data in each class for pruning}

In our experiments, we selected easy and hard examples without consideration of class imbalance. The class distributions of 10k easy and 10k hard examples in our experiment are shown in Figure~\ref{fig:supple_examples_distribution}. We conducted the ablation study on the effect of balancing the class distribution of the hard examples subset. Table~\ref{tab:ablation_balancing} shows the results of the pruning models of TRADES, which pruned top 10\% examples with high difficulty from the training dataset of CIFAR-10. We experimented with varying the difficulty measurement loss and the balance of the number of pruned data for each class (the number of pruned data for each class in balanced subset is 500). We extracted the balanced pruning subset by selecting top 10\% examples with high difficulty from the training subset of each class.
It is observed that while the pruning models without consideration of class imbalance improve robustness, the pruning models with the balanced pruning subset show the similar robustness performance with that of the baseline model. It indicates that when selecting hard examples which have the negative effect to training, it is less effective for the improvement of the robustness performance to consider the balance of the number of data for each class.

\paragraph{Study on the robust fairness}

In~\citet{adv_fairness}, they indicated that adversarial training differently increases robustness of each class in a dataset, which effectively increases the robustness of easy classes but not effectively increases the robustness of hard classes. They noted that the performance imbalance between classes, which is termed as the robustness fairness problem, becomes more significant in adversarial training. Because DPLS regularizes hard examples, it can be assumed that the increase in the robustness performance is attributed to the increase in the performance of easy classes. Thus, we compared the robustness performance of easy and hard classes in Table~\ref{tab:ablation_fairness}. We divide the class of CIFAR-10 into 5 easy and 5 hard class subsets. We referred to the class distributions of easy and hard examples from Figure~\ref{fig:supple_examples_distribution} and divide them into easy (class 0, 1, 7, 8, 9) and hard (class 2, 3, 4, 5, 6). From Table~\ref{tab:ablation_fairness}, it is observed that the performance increase is attributed to the increase in the robustness improvement of both the easy and hard class subsets in the DPLS models. It indicates that our method does not only exploit the performance increase of the easy classes, but increases the both performance of easy and hard classes.

\input{table/tab_supple_ablation_fairness}

%% file: table/tab_supple_ablation_sup.tex
\begin{table}[h]
\begin{minipage}{.48\linewidth}
\caption{Performance of DPLS according to
the DPLS calculation epoch $T$.}
\vspace{8pt}
\centering
\footnotesize
\addtolength{\tabcolsep}{-0pt}
\begin{tabular}{c|cc|cc}
    \toprule
     \multicolumn{1}{c|}{\multirow{2}{*}{$T$}} &
    \multicolumn{2}{c|}{PGD} &
    \multicolumn{2}{c}{TRADES} 
    \\
      \multicolumn{1}{c|}{} &  \multicolumn{1}{c}{Std.} & 
      \multicolumn{1}{c|}{Rob.} &
      \multicolumn{1}{c}{Std.} & 
      \multicolumn{1}{c}{Rob.}
    \\
    \midrule[.05em]
     \multicolumn{1}{c|}{10} & 84.13 & 53.34 & 84.88 & 54.69 \\
     \multicolumn{1}{c|}{50} & 86.33 & \textbf{54.05} & \textbf{85.56} &  55.0
      \\
     \multicolumn{1}{c|}{90} & \textbf{87.13} & 54.03 & 85.36 &  \textbf{55.19}
      \\
     \multicolumn{1}{c|}{100} & 86.94 & 53.8 & 85.53 & 54.77
      \\
    \midrule[.1em]
\end{tabular}
\label{tab:ablation_1}
\end{minipage}%
\hfill
\begin{minipage}{.48\linewidth}
\centering
\footnotesize
\addtolength{\tabcolsep}{-0pt}
\caption{Performance of DPLS according to the re-initialized epoch, where DPLS is pre-calculated by using DPLS of the calculation epoch 90.}
\vspace{8pt}
\begin{tabular}{cc|cc|cc}
    \toprule
     \multicolumn{1}{c}{\multirow{2}{*}{$T$}} &
     \multicolumn{1}{c|}{\multirow{2}{*}{Reinit.}} &
    \multicolumn{2}{c|}{PGD} &
    \multicolumn{2}{c}{TRADES} 
    \\
      & \multicolumn{1}{c|}{} &  \multicolumn{1}{c}{Std.} & 
      \multicolumn{1}{c|}{Rob.} &
      \multicolumn{1}{c}{Std.} & 
      \multicolumn{1}{c}{Rob.}
    \\
    \midrule[.05em]
     \multirow{4}{*}{90} & 0 & 86.62 & 53.69 & 84.99 & 55.05 \\
     &\multicolumn{1}{c|}{50} & 86.68 & \textbf{54.09} & 84.89 & 55.11
      \\
     &\multicolumn{1}{c|}{90} & \textbf{87.13} & 54.03 & \textbf{85.36} & \textbf{55.19}
      \\
     &\multicolumn{1}{c|}{100} & 86.76 & 53.76 & 85.26 & 54.93
      \\

    \midrule[.1em]
\end{tabular}
\label{tab:ablation_2}
\end{minipage} 
\end{table}

%% file: table/tab_supple_ablation_sup2.tex
\begin{table*}[h]
\centering
\caption{Performance of DPLS applying to the PGD model on CIFAR-10 according to
the smoothing factor $\lambda$.}
\begin{tabular}{ccc|ccc}
    \toprule
     \multicolumn{1}{c}{\multirow{1}{*}{Model}} &
     \multicolumn{1}{c}{\begin{tabular}[c]{@{}c@{}}$\lambda$\\ (smoothing factor)\end{tabular}}
     &
     \multicolumn{1}{c|}{\multirow{1}{*}{Avg. smoothing}} &
     \multicolumn{1}{c}{\multirow{1}{*}{Standard}} &
     
    \multicolumn{1}{c}{\begin{tabular}[c]{@{}c@{}}Robust\\ (PGD)\end{tabular}} &
    \multicolumn{1}{c}{\begin{tabular}[c]{@{}c@{}}Robust\\ ($A^3$)\end{tabular}} \\
    
    \midrule[.05em]
    \multirow{6}{*}{PGD} & -  & 1.0 & 87.19 & 56.44 & 51.8\\
     & 0.9  & 0.98 & \textbf{87.73} & 56.8 & 52.37 \\
     & 0.7  & 0.939 & 87.25 & 57.58 & 53.17 \\
     & 0.5  & 0.898 & 87.13 & 58.43 & 54.03 \\
     & 0.3  & 0.857 & 86.09 & \textbf{58.51} & \textbf{54.29} \\
     & 0.1  & 0.816 & 85.2 & 58.36 & 53.92 \\
    \midrule[.1em]
\end{tabular}
\label{tab:ablation_dpls_param_pgd}
\end{table*}

%% file: table/tab_supple_ablation_sup3.tex
\begin{table*}[h]
\centering
\caption{Performance of DPLS applying to the TRADES model on CIFAR-10 according to
the smoothing factor $\lambda$.}
\begin{tabular}{ccc|cccc}
    \toprule
     \multicolumn{1}{c}{\multirow{1}{*}{Model}} &
     \multicolumn{1}{c}{\begin{tabular}[c]{@{}c@{}}$\lambda$\\ (smoothing factor)\end{tabular}}
     &
     \multicolumn{1}{c|}{\multirow{1}{*}{Avg. smoothing}} &
     \multicolumn{1}{c}{\multirow{1}{*}{Standard}} &
     
    \multicolumn{1}{c}{\begin{tabular}[c]{@{}c@{}}Robust\\ (PGD)\end{tabular}} &
    \multicolumn{1}{c}{\begin{tabular}[c]{@{}c@{}}Robust\\ ($A^3$)\end{tabular}} \\
    
    \midrule[.05em]
    \multirow{6}{*}{TRADES} & -  & 1.0 & \textbf{85.66} & 58.46 & 54.08 \\
     & 0.9  & 0.979 & 85.49 & 58.74 & 54.39 \\
     & 0.7  & 0.935 & 85.41 & 59.08 & 54.8 \\
     & 0.5  & 0.893 & 85.36 & \textbf{59.33} & 55.19 \\
     & 0.3  & 0.85 & 84.95 & 59.29 & 55.22 \\
     & 0.1  & 0.807 & 84.06 & 59.25 & \textbf{55.38} \\
    \midrule[.1em]
\end{tabular}
\label{tab:ablation_dpls_param_trades}
\end{table*}

%% file: table/tab_supple_reweight.tex
\begin{table}[h]
\centering
\caption{Performance comparison of DPLS with reweighting methods.}
\vspace{0pt}
\begin{tabular}{c|ccc|ccc}
    \toprule
     \multicolumn{1}{c|}{\multirow{3}{*}{Method}} & \multicolumn{3}{c|}{PGD} &
     \multicolumn{3}{c}{TRADES} \\
    & \multicolumn{1}{c}{\multirow{2}{*}{Std.}} & Robust & \multicolumn{1}{c|}{Robust}
    & \multicolumn{1}{c}{\multirow{2}{*}{Std.}} & Robust & Robust
    \\
    & & (PGD) & \multicolumn{1}{c|}{($A^3$)} & & (PGD) & ($A^3$) \\
    
    \midrule[.05em]
    - & \textbf{87.19} & 56.44 & 51.8
      & \textbf{85.66} & 58.46 & 54.08 
    \\
    DPR & 86.1 & 57.32 & 53.2
             & 84.63 & 58.54 & 54.88
    \\
    DPLS & 87.13 & \textbf{58.43} & \textbf{54.03}
            & 85.36 & \textbf{59.33} & \textbf{55.19}
    \\
    
    \midrule[.1em]
\end{tabular}
\label{tab:appendix_reweight}
\end{table}

%% file: table/tab_supple_ablation_bal.tex
\begin{table*}[h]
\centering
\caption{Performance of pruning models according to various difficulty losses with and without balancing the excluded data number of each class.}
\begin{tabular}{ccc|cccc}
    \toprule
     \multicolumn{1}{c}{\multirow{1}{*}{Model}} &
     \multicolumn{1}{c}{Method} &
     \multicolumn{1}{c|}{Balancing} &
     \multicolumn{1}{c}{\multirow{1}{*}{Standard}} &
     
    \multicolumn{1}{c}{\begin{tabular}[c]{@{}c@{}}Robust\\ (PGD)\end{tabular}} &
    \multicolumn{1}{c}{\begin{tabular}[c]{@{}c@{}}Robust\\ (CW)\end{tabular}} &
    \multicolumn{1}{c}{\begin{tabular}[c]{@{}c@{}}Robust\\ ($A^3$)\end{tabular}} \\
    
    \midrule[.05em]
    \multirow{9}{*}{TRADES} & -  & - & \textbf{85.66} & 58.46 & 56.53 & 54.08 \\
     & C-score  & $\mcirc$ & 85.2 & 57.87 & 56.72 & 53.97 \\
     & C-score  & $\times$ & 85.35 & \textbf{58.52} & 56.82 & 54.25 \\
     & CE loss  & $\mcirc$ & 83.7 & 57.49 & 56.76 & 54.01\\
     & CE loss  & $\times$ & 83.82 & 58.37 & \textbf{57.31} & 54.45 \\
     & EL2N  & $\mcirc$ & 83.8 & 57.76 & 56.89 & 54.07 \\
     & EL2N  & $\times$ & 83.25 & 57.86 & 56.9 & 54.28 \\
     & 0-1 loss  & $\mcirc$ & 83.81 & 57.84 & 56.76 & 54.22 \\
     & 0-1 loss  & $\times$ & 84.07 & 58.04 & 57.4 & \textbf{54.64} \\
    \midrule[.1em]
\end{tabular}
\label{tab:ablation_balancing}
\end{table*}

%% file: table/tab_supple_ablation_fairness.tex
\begin{table*}[h]
\centering
\caption{Robustness performance against adversarial attacks for easy class and hard class subsets of CIFAR-10.}
\begin{tabular}{cc|cccccc}
    \toprule
     \multicolumn{1}{c}{\multirow{1}{*}{Model}} &
     \multicolumn{1}{c|}{Method} &
     \multicolumn{1}{c}{\multirow{1}{*}{PGD}} &
    \multicolumn{1}{c}{\begin{tabular}[c]{@{}c@{}}PGD\\ (easy)\end{tabular}} &
    \multicolumn{1}{c}{\begin{tabular}[c]{@{}c@{}}PGD\\ (hard)\end{tabular}} &
    \multicolumn{1}{c}{\multirow{1}{*}{$A^3$}} &
    \multicolumn{1}{c}{\begin{tabular}[c]{@{}c@{}}$A^3$\\ (easy)\end{tabular}} &
    \multicolumn{1}{c}{\begin{tabular}[c]{@{}c@{}}$A^3$\\ (hard)\end{tabular}}  \\
    \midrule[.05em]
    \multirow{2}{*}{PGD} & -  
    & 56.54 &  70.5 & 42.58
    & 51.8 & 67.62 & 35.98 \\
     & DPLS  
     & \textbf{58.49} & \textbf{71.82} & \textbf{45.16}
     & \textbf{54.05} & \textbf{68.66} & \textbf{39.44} \\
     \cmidrule{1-8}
     \multirow{2}{*}{TRADES} & -  
     & 58.52 & 71.42 & 45.62
     & 54.08 & 68.64 & 39.52 \\
     & DPLS  
     & \textbf{59.32} & \textbf{72.9} & 
     \textbf{45.74}
     & \textbf{55.19} & \textbf{70.22} & 
     \textbf{40.16} \\
    \midrule[.1em]
\end{tabular}
\label{tab:ablation_fairness}
\end{table*}

%% file: F_comparison.tex
\section{Additional comparison study}
\label{appendix:comparison}

\input{table/tab_supple_md}

\paragraph{Additional comparative study using existing evaluation attack methods}

\jmlrr{
We primarily evaluated our models against AutoAttack \citep{attack_autoattack} because it provides an accurate assessment without giving a false sense of security due to issues like gradient obfuscation. Since the MD ensemble attack \citep{attack_md}, which includes margin decomposition (MD), appears to be a variant of AutoAttack, we also measured the CIFAR-10 model's performance against this attack. The results are presented in Table \ref{tab:appendix_md}. As shown in the table, the MD ensemble attack yields performance results and patterns very similar to those of AutoAttack, demonstrating that our method, DPLS, is also robust against the MD attack.
}

\jmlrr{
Furthermore, we evaluated our models against RayS \citep{attack_rays}, an attack method that can reveal vulnerabilities leading to a false sense of security. The results, also presented in Table \ref{tab:appendix_md}, indicate that our methodology, DPLS, is robust against RayS and shows improved performance compared to the baseline model.
}

\input{table/tab_supple_diffpure}

\paragraph{Additional comparative study combining existing defense methods}

\jmlrr{
DiffPure \citep{defense_diffpure} is an image purification method that leverages pre-trained diffusion-based models. Since the classifier model operates independently of the image purification method, it can be combined with any adversarial training-based approach. We applied the DiffPure method to two models: one trained using baseline adversarial training methods without DPLS, and another trained using adversarial training methods with DPLS. In the CIFAR-10 setting of DiffPure, we measured the performance by replacing the classifier model and evaluated its robustness using RobustBench \citep{robustbench}.
}

\jmlrr{
As shown in Table \ref{tab:appendix_diffpure}, we compared the performance of adding only DiffPure to the PGD and TRADES baseline models with the performance when both DiffPure and DPLS are added together. We observed that the models with DPLS added achieved relatively higher performance. Since DiffPure operates independently of the classifier model to enhance adversarial defense performance, utilizing a model trained with DPLS—which is comparatively more robust—appears to result in better defense performance.
}

%% file: table/tab_supple_md.tex
\begin{table}[h]
\centering
\caption{\jmlrr{Performance evaluation of DPLS using MD \citep{attack_md} and RayS \citep{attack_rays} attacks.}}
\vspace{0pt}
\begin{tabular}{c|cccc|cccc}
    \toprule
     \multicolumn{1}{c|}{\multirow{3}{*}{Method}} & \multicolumn{4}{c|}{PGD} &
     \multicolumn{4}{c}{TRADES} \\
    & \multicolumn{1}{c}{\multirow{2}{*}{Std.}} & Robust & Robust & \multicolumn{1}{c|}{Robust}
    & \multicolumn{1}{c}{\multirow{2}{*}{Std.}} & Robust & Robust & Robust
    \\
    & & (MD) & (RayS) & \multicolumn{1}{c|}{($A^3$)} & & (MD) & (RayS) & ($A^3$) \\
    
    \midrule[.05em]
    - & \textbf{87.19} & 51.93 & 61.13 & 51.80
      & \textbf{85.66} & 54.17 & 60.84 & 54.08 
    \\
    DPLS & 87.13 & \textbf{54.07} & \textbf{62.14} & \textbf{54.03}
            & 85.36 & \textbf{55.25} & \textbf{62.11} & \textbf{55.19}
    \\
    
    \midrule[.1em]
\end{tabular}
\label{tab:appendix_md}
\end{table}

%% file: table/tab_supple_diffpure.tex
\begin{table}[h]
\centering
\caption{\jmlrr{Performance evaluation of DPLS combined with Diffpure \citep{defense_diffpure} on RobustBench \citep{robustbench}}}
\vspace{0pt}
\begin{tabular}{l|cc|cc}
    \toprule
     \multicolumn{1}{c|}{\multirow{3}{*}{Dataset}} & \multicolumn{2}{c|}{PGD} &
     \multicolumn{2}{c}{TRADES} \\
    & \multicolumn{1}{c}{\multirow{2}{*}{Std.}}  & \multicolumn{1}{c|}{Robust}
    & \multicolumn{1}{c}{\multirow{2}{*}{Std.}}  & Robust
    \\
    & & \multicolumn{1}{c|}{($A^3$)} & & ($A^3$) \\
    
    \midrule[.05em]
    Baseline & \textbf{88.28} & 53.32
      & 86.13 & 56.05 
    \\
    Baseline + DPLS & 86.72 & 56.25
            & 85.15 & 56.84    
    \\
    Baseline + DiffPure & \textbf{88.28} &  66.87
    & \textbf{87.11}  & 66.80
    \\
    Baseline + DPLS + DiffPure & 87.50 &  \textbf{68.61}
    & 86.72  & \textbf{71.09}
    \\
    
    \midrule[.1em]
\end{tabular}
\label{tab:appendix_diffpure}
\end{table}